%% file: main.tex
\documentclass{article}

\usepackage[main,final]{neurips_2026}

\usepackage[utf8]{inputenc}
\usepackage[T1]{fontenc}
\usepackage{url}
\usepackage{booktabs}
\usepackage{amsfonts}
\usepackage{nicefrac}
\usepackage{microtype}
\usepackage{xcolor}

\usepackage{amsmath}
\usepackage{amssymb}
\usepackage{amsthm}

\usepackage{graphicx}
\graphicspath{{figures/}}

\usepackage{enumitem}   
\usepackage{colortbl}   

\usepackage{mathtools}   

\usepackage{algorithm}
\usepackage{algpseudocode}
\usepackage{placeins}
\usepackage{needspace}
\usepackage{hyperref}

\newtheorem{theorem}{Theorem}
\newtheorem{lemma}[theorem]{Lemma}
\newtheorem{proposition}[theorem]{Proposition}
\newtheorem{corollary}[theorem]{Corollary}

\newtheorem{remark}[theorem]{Remark}
\newtheorem{assumption}{Assumption}   

\title{Preconditioned Implicit Midpoint Langevin Sampling \\ for Non-Smooth Bayesian Imaging}

\author{Sichen Wang$^{1}$ \qquad Zhipeng Lu$^{1,2}$\\
$^{1}$Department of Engineering, Shenzhen MSU-BIT University\\
$^{2}$Guangdong Laboratory of Machine Perception and Intelligent Computing\\
Shenzhen, China\\
Correspondence: \texttt{zhipeng.lu@smbu.edu.cn}}

\begin{document}

\maketitle

\begin{abstract}
We introduce P-IMLA, a preconditioned implicit-midpoint Langevin sampler for high-dimensional log-concave targets with non-smooth regularization. The obstruction is that the large steps enabled by matrix preconditioning are precisely where preconditioned Euler--Maruyama inflates stationary variance. Implicit midpoint removes this surplus via a Cayley-transform identity, making P-IMLA Gaussian-exact at any step size. Beyond Gaussians, whitening transfers the sharp IMLA contraction rate to $\kappa_{\mathrm{eff}}$, and backward-error analysis gives drift-only rather than Euler-type diffusion bias under explicit smoothness assumptions. An $M$-norm Moreau envelope yields a proximal non-smooth variant with a three-term error budget. Controlled Gaussian and TV-regularized imaging experiments confirm the predicted separation: the unpreconditioned midpoint scheme retains long transients as $\kappa_f$ grows, whereas P-IMLA remains $\kappa_{\mathrm{eff}}$-governed and near variance-exact.
\end{abstract}


\input{sections/introduction}

\input{sections/notation}

\input{sections/background}

\input{sections/algorithm}

\input{sections/theory}

\input{sections/jp_guarantees}

\input{sections/experiments}

\input{sections/conclusion}

\clearpage
\section*{Acknowledgments}
We thank the anonymous reviewers for their careful reading and constructive feedback, which substantially improved the paper. This work is supported by Guangdong Province (No.\ 2023QN10X215), Shenzhen Science and Technology Program (No.\ JCYJ20241202130548062), the Natural Science Foundation of Shenzhen (No.\ JCYJ20230807142703006), and the Key Research Platforms and Projects of the Guangdong Provincial Department of Education (No.\ 2023ZDZX1034).
The authors declare no competing interests.

\bibliographystyle{plainnat}
\bibliography{references}

\clearpage
\appendix
\input{sections/appendix}
\end{document}

%% file: sections/introduction.tex
\section{Introduction}
\label{sec:intro}

Large-scale Bayesian inverse problems---image deblurring, computed tomography, magnetic resonance imaging \citep{stuart2010inverse}---give rise to posterior distributions $\pi(x) \propto \exp(-f(x) - g(x))$ in which $f$ is a smooth data-fit term and $g$ is a non-smooth regulariser such as total variation or wavelet-$\ell_1$ \citep{pereyra2016proximal,durmus2018efficient,habring2024subgradient}. At image resolutions of $256 \times 256$ and above, the Hessian of $f$ routinely exhibits condition numbers $\kappa_f$ on the order of $10^5$--$10^7$, inflating the mixing time of the unadjusted Langevin algorithm \citep[ULA;][]{roberts1996exponential} for smooth strongly log-concave targets to $O(\kappa_f / \varepsilon^2)$ iterations \citep{dalalyan2017theoretical, durmus2019high, cheng2018kl, vempala2019rapid, chewi2026logconcave}---in practice, computationally out of reach.

On a smooth potential $U$, ULA takes $X_{n+1}=X_n-\delta\nabla U(X_n)+\sqrt{2\delta}\,\xi_n$, with $\xi_n\sim\mathcal N(0,I)$. Preconditioning is the classical remedy. Replacing the Langevin drift by $-M^{-1}\nabla f$ with $M \approx \nabla^2 f$ equalises curvatures across coordinates and, in principle, lowers the effective condition number $\kappa_{\mathrm{eff}}=L_M/m_M$ toward $O(1)$, where $m_M,L_M$ bound the eigenvalues of $M^{-1/2}\nabla^2fM^{-1/2}$, echoing long-established accelerations in iterative linear solvers and optimisation \citep{girolami2011riemann, titsias2024optimal, hird2025preconditioning}. Yet the preconditioned unadjusted Langevin algorithm (P-ULA) faces an obstruction: at large dimensionless steps $\delta\ell$, its explicit-Euler discretisation incurs variance inflation $1/(1-\delta\ell/2)$ on a whitened mode of precision $\ell$, offsetting the intended conditioning gain.

\textbf{An algebraic resolution.} On a scalar Gaussian target with precision $\ell$, every $\theta$-scheme reduces the Langevin diffusion to an AR(1) recursion $z_{n+1} = r\,z_n + \sigma\,\xi_n$ with stationary variance $\sigma^2 / (1 - r^2)$. Matching this to the target variance $v = \ell^{-1}$ is equivalent to the \emph{variance-budget identity} $r^2 + \sigma^2/v = 1$. Direct computation, with $a \coloneqq \delta\ell/2$, gives two qualitatively different outcomes:
\begin{align}
\text{explicit Euler:} \quad & r^2 + \sigma^2/v \;=\; (1 - 2a)^2 + 4a \;=\; 1 + 4a^2, \label{eq:budget-euler}\\[2pt]
\text{implicit midpoint ($\theta = \tfrac{1}{2}$):} \quad & r^2 + \sigma^2/v \;=\; \frac{(1-a)^2 + 4a}{(1+a)^2} \;\equiv\; 1. \label{eq:budget-midpoint}
\end{align}
The $4a^2$ surplus in~\eqref{eq:budget-euler} is precisely what preconditioning's large steps amplify; its exact cancellation in~\eqref{eq:budget-midpoint}---unique to $\theta = 1/2$ among $\theta$-schemes, traceable to the Cayley transform $r(a) = (1-a)/(1+a)$---is the algebraic cancellation that this paper builds on, making implicit-midpoint integration a natural ingredient for realising large-step preconditioning. Figure~\ref{fig:2x2} maps the resulting $2 \times 2$ landscape.

\begin{figure}[t]
\centering
\small
\renewcommand{\arraystretch}{1.35}
\setlength{\tabcolsep}{8pt}
\begin{tabular}{r|c|c|}
\cline{2-3}
& \multicolumn{1}{c|}{\textbf{No preconditioning} ($M = I$)}
& \multicolumn{1}{c|}{\textbf{Preconditioned} ($M \approx \nabla^2 f$)} \\
\cline{2-3}
\textbf{Explicit Euler}
& \cellcolor{red!8}\begin{tabular}{@{}c@{}}slow-mixing\\[-1pt]\scriptsize ULA: VR $= 0.24$\end{tabular}
& \cellcolor{red!8}\begin{tabular}{@{}c@{}}unstable\\[-1pt]\scriptsize P-ULA: VR $\to \infty$\end{tabular} \\
\cline{2-3}
\textbf{Implicit midpoint}
& \cellcolor{orange!10}\begin{tabular}{@{}c@{}}long transient\\[-1pt]\scriptsize IMLA: VR $= 48.6$\end{tabular}
& \cellcolor{green!12}\begin{tabular}{@{}c@{}}\textbf{rapid relaxation}\\[-1pt]\scriptsize \textbf{P-IMLA: VR $= 1.0246 \pm 0.0001$}\end{tabular} \\
\cline{2-3}
\end{tabular}
\caption{\textbf{The $2\times2$ landscape at a fixed simulation budget.} Gaussian target with mean $\mu$, precision $H$ and $\kappa_f=4\times10^7$ (App.~\ref{app:exp-2x2}). VR denotes the time-averaged normalised mean-square displacement from the target mean, $d^{-1}\overline{(X-\mu)^\top H(X-\mu)}$, including the cold-start transient. It is not a stationary variance estimate. Both midpoint schemes are Gaussian-exact with exact solves; $98.7\%$ of IMLA's displayed $48.6$ is transient mean error (App.~\ref{app:r5-vr}). Preconditioning makes that transient short at the reported budget.}
\label{fig:2x2}
\end{figure}

\textbf{Contributions.} We introduce the \emph{Preconditioned Implicit Midpoint Langevin Algorithm} (P-IMLA) and its non-smooth variant P-IMLA$_\lambda$ based on an $M$-norm Moreau envelope; to our knowledge this is the first matrix-preconditioned implicit-midpoint proximal Langevin sampler. Our contributions are:
\begin{enumerate}[leftmargin=1.5em, topsep=2pt, itemsep=2pt]
\item \textbf{Algorithm (§\ref{sec:algorithm}).} An inner solver computes $\mathrm{prox}^M_{(\delta/2)U_\lambda}$ for $U_\lambda = f + g^{\lambda, M}$ at $O(K_{\mathrm P}K_{\mathrm{in}}d\log d)$ per outer iteration on convolutional models with TV priors, matching MYULA's per-step cost; an engineering variant JP-IMLA (App.~\ref{app:jp-equivalence}) gives $\sim\!60\!\times$ per-iteration speedup.
\item \textbf{Gaussian exactness and contraction (Theorems~\ref{thm:gaussian-exact}--\ref{thm:contraction}).} On Gaussian targets, P-IMLA's stationary distribution equals $\pi$ exactly at every step size---via the identity in~\eqref{eq:budget-midpoint}. On general $m$-strongly log-concave targets, whitening transfers the reflected-proximal bound of \citet{klatzer2024accelerated} to an $M$-norm Wasserstein contraction rate $\rho^\star = (\sqrt{\kappa_{\mathrm{eff}}} - 1)/(\sqrt{\kappa_{\mathrm{eff}}} + 1)$, with quadratic targets attaining the bound.
\item \textbf{Sharp drift-only bias (Theorem~\ref{thm:bea-bias}, primary novelty).} A backward-error analysis under the regularity and Poisson assumptions of Theorem~\ref{thm:bea-bias} gives the asymptotic bias formula
\begin{equation}
\pi_\delta(\varphi) - \pi(\varphi)
\;=\;
\tfrac{\delta}{4}\,\mathrm{Cov}_\pi\!\big(\operatorname{tr}(M^{-1}\nabla^2 f),\,\varphi\big) + o(\delta),
\label{eq:bea-bias-intro}
\end{equation}
whose bias operator is first-order (drift-only), in structural contrast to P-ULA's second-order drift-\emph{and}-diffusion bias. The drift-only form forces the bias to vanish on Gaussians (where $\operatorname{tr}(M^{-1}\nabla^2 f)$ is constant) while retaining an $O(\delta)$ stationary weak bias in general. Here ``first-order'' describes the differential operator, not an improvement in the order in $\delta$.
\item \textbf{End-to-end error budget (Theorem~\ref{thm:end-to-end}).} A three-term $W_2$ decomposition---contraction $+$ discretisation $+$ Moreau smoothing---with an explicit $C^{1,1}$ discretisation remainder and coupled $(\delta,\lambda)$ dependence; see Rem.~\ref{rem:precision-protocol} for the fixed-$\lambda$ (P1) vs full-$\varepsilon$ (P2) regimes. We further verify (App.~\ref{app:moreau-bias}) the smoothing term's $\sqrt{\lambda}$ exponent is non-tight on polyhedral $g$ (convex functions with polyhedral epigraphs, including $\ell_1$ penalties \citep{bauschke2011convex}; empirical slopes $0.97$ asymmetric / $1.95$ symmetric, both strictly above the worst-case exponent $0.5$).
\item \textbf{Imaging-scale experiments at fixed iteration budgets (§\ref{sec:experiments}).} On $256\!\times\!256$ TV deconvolution at a $100\!\times$ iteration-budget ratio, P-IMLA$_\lambda$ reaches cold-start per-pixel $\hat{R}<1.05$ at iteration $10$ (contraction timescale $N\approx22$, not a theoretical $\hat R$ threshold at $\kappa_{\mathrm{eff}}^\lambda=201$) and $\hat{R}\leq 1.001$ at $N=2{,}000$ with SSE $= 855.22 \pm 0.66$, while SP-MYULA reaches near-unity $\hat{R}$ at $\sim\!4\!\times$ larger SSE; a MALA comparison (App.~\ref{app:exp-mala-ref}) gives covariance-trace ratios $7.68$ vs $0.95$, complemented by overdispersed-start diagnostics and target-exact small-image Gibbs checks (App.~\ref{app:rebuttal-experiments}), and MYULA / SK-ROCK fail to mix. The $3.5\!\times\!\!-\!\!4.0\!\times$ advantage is stable across geometric, Shepp--Logan, and cameraman phantoms.
\end{enumerate}

\textbf{Related work.} Prior work splits into two regimes. The \emph{preconditioned-Langevin} thread \citep{girolami2011riemann,titsias2024optimal,hird2025preconditioning} reshapes sampling geometry, while explicit Euler retains a stability and bias trade-off. SK-ROCK \citep{pereyra2020accelerating} stabilises the explicit step via Chebyshev stages, and \citet{corbineau2019preconditioned} combines preconditioning with non-smooth imaging priors. The \emph{implicit-integrator} thread \citep{hodgkinson2021implicit,klatzer2024accelerated} studies implicit Langevin discretisations, alongside the geometric-integrator-for-sampling tradition of \citet{bouRabee2018geometric}; the semi-implicit pCNL \citep{cotter2013mcmc} treats a Gaussian-prior covariance implicitly with Metropolis correction. Underdamped Langevin \citep{cheng2018underdamped,leimkuhler2024contraction}, HMC/NUTS \citep{neal2011mcmc,hoffman2014nuts}, and Wasserstein-gradient-flow samplers \citep{wibisono2018sampling} provide complementary acceleration mechanisms outside our controlled integrator comparison. IMLA \citep{klatzer2024accelerated} ($M=I$, $\tilde O(\sqrt{\kappa_f})$) is the closest predecessor. At large $\kappa_f$, its low-curvature modes retain long initial transients; the VR $=48.6$ in Figure~\ref{fig:2x2} is a finite-budget statistic, not a failure of Gaussian exactness. For fixed $M$, P-IMLA is exactly IMLA in whitened coordinates: Theorem~\ref{thm:contraction} transports the factor of \citet{klatzer2024accelerated}, Thm.~3.3, while relaxing $C^2$ to $C^{1,1}$ regularity. We differ in the $M$-norm Moreau construction, stationary-bias analysis, and treatment of the joint-proximal implementation. Coupling methods \citep{eberle2016reflection,klatzer2024accelerated} control convergence distances; building on weak-error analysis \citep{talay1990expansion,abdulle2014highorder}, we identify the leading invariant coefficient and its drift-only defect, under explicit regularity and Poisson assumptions. The generalized Hessian therefore determines both the rate and the leading bias, which vanishes on Gaussians while remaining first order in the step size for general smooth targets. The non-smooth construction extends proximal Langevin methods \citep{pereyra2016proximal,durmus2018efficient,durmus2022proximal,habring2026diffusion} using an $M$-norm envelope whose smoothness constant adds to $L_M$ in the same metric. Recent Bayesian imaging work \citep{laumont2022bayesian,holden2022bayesian,ehrhardt2024proximal,renaud2025stability} provides a broader modelling and algorithmic context for this construction. Our JP bounds separately quantify the change of proximal model and the finite inner-solver error.

%% file: sections/notation.tex
\begin{table}[h]
\centering
\caption{Notation. The preconditioner $M$ is fixed throughout.}
\label{tab:notation}
\small
\begin{tabular}{ll}
\toprule
Symbol & Meaning \\
\midrule
$\pi,\;\pi_\lambda$ & Original and Moreau-smoothed targets \\
$\delta,\;\lambda,\;\tau=\delta/2$ & Langevin step, Moreau parameter, proximal step \\
$\ell,\;m_M,\;L_M$ & Spectral value and curvature bounds in the $M$-metric \\
$L_\lambda=L_M+\lambda^{-1}$ & Smoothness bound for $U_\lambda=f+g^{\lambda,M}$ \\
$P,\;Q,\;\widehat Q$ & Exact smoothed, exact joint-proximal, implemented JP updates \\
$K_{\mathrm P},\;K_{\mathrm{in}}$ & Picard and inner ADMM iteration budgets \\
\bottomrule
\end{tabular}
\end{table}

%% file: sections/background.tex
\section{Background}
\label{sec:background}

\subsection{Problem setup and preconditioned Langevin}
\label{sec:setup}

We sample from $\pi(x) \propto \exp(-f(x) - g(x))$ on $\mathbb{R}^d$ as introduced in §\ref{sec:intro}, under the following structural hypothesis throughout:

\begin{assumption}[Target regularity]
\label{ass:target}
$f \in C^2(\mathbb{R}^d)$ is $m$-strongly convex and $L$-smooth, i.e.\ $m\,I \preceq \nabla^2\!f(x) \preceq L\,I$ for every $x$; and $g\colon\mathbb{R}^d \to \mathbb{R}\cup\{+\infty\}$ is proper, convex, and lower semicontinuous, but possibly non-smooth.
\end{assumption}

A fixed, position-independent symmetric positive-definite \emph{preconditioner} $M \succ 0$ reshapes the Langevin diffusion:
\begin{equation}
    \mathrm{d}X_t \;=\; -M^{-1}\nabla\!f(X_t)\,\mathrm{d}t \;+\; \sqrt{2}\,M^{-1/2}\,\mathrm{d}W_t,
    \label{eq:precond-sde}
\end{equation}
whose invariant distribution is still $\pi \propto \exp(-f)$ (restricting momentarily to $g \equiv 0$): the preconditioner accelerates mixing without distorting the target. The effective strong-convexity and smoothness constants of $f$ in the induced $M$-norm $\|v\|_M^2 \coloneqq \langle v, Mv\rangle$ are $m_M$ and $L_M$, defined by
\begin{equation}
    m_M\,I \;\preceq\; M^{-1/2}\,\nabla^2\!f(x)\,M^{-1/2} \;\preceq\; L_M\,I \quad \forall\, x,
    \qquad \kappa_{\mathrm{eff}} \;\coloneqq\; \frac{L_M}{m_M}.
    \label{eq:kappa-eff}
\end{equation}
When $M \approx \nabla^2\!f$ (the optimization-motivated ``Newton-like'' choice), $\kappa_{\mathrm{eff}} = O(1)$ and the preconditioner has, in principle, fully absorbed the conditioning of the target. The $M$-induced Wasserstein distance $W_2^M$ on $\mathcal{P}_2(\mathbb{R}^d)$---defined by replacing Euclidean cost with $\|\cdot\|_M^2$ in the standard optimal-transport formulation---is the natural metric for quantifying convergence of preconditioned chains.

\paragraph{Whitening.} Let $T=M^{1/2}$ and $\widetilde f(z)=f(T^{-1}z)$, $\widetilde g(z)=g(T^{-1}z)$. Then $T\operatorname{prox}_{\tau f}^{M}(w)=\operatorname{prox}_{\tau\widetilde f}(Tw)$, $g^{\lambda,M}(T^{-1}z)=\widetilde g^\lambda(z)$, and $W_2^M(\mu,\nu)=W_2(T_\#\mu,T_\#\nu)$. Thus fixed-$M$ P-IMLA is canonically conjugate to IMLA; the following calculation isolates Euler's discretisation bias in these coordinates.

\begin{proposition}[P-ULA variance inflation at preconditioned step sizes]
\label{prop:p-ula-failure}
Let $f(x) = \tfrac{1}{2}x^\top H x$ with $H \succ 0$, $g \equiv 0$, preconditioner $M \succ 0$, and write $\ell_1,\ldots,\ell_d$ for the eigenvalues of $M^{-1/2} H M^{-1/2}$. For any $\delta > 0$ with $\delta\ell_i < 2$ for all $i$ (the forward-Euler stability boundary), the preconditioned Euler--Maruyama iteration
\[
X_{n+1} \;=\; X_n \;-\; \delta\, M^{-1}\nabla f(X_n) \;+\; \sqrt{2\delta}\,M^{-1/2}\xi_n, \qquad \xi_n \sim \mathcal{N}(0, I),
\]
admits a unique stationary distribution $\pi_\delta^{\mathrm{Euler}}$ whose variance in generalized-Hessian eigencomponent $z_i=q_i^\top M^{1/2}x$ (with $q_i$ a unit eigenvector of $M^{-1/2}HM^{-1/2}$) is
\begin{equation}
\operatorname{Var}_{\pi_\delta^{\mathrm{Euler}}}\!(z_i) \;=\; \ell_i^{-1}\,\cdot\,\frac{1}{1 \;-\; \delta\ell_i/2},
\label{eq:ula-inflation}
\end{equation}
against the target's $\operatorname{Var}_\pi(z_i) = \ell_i^{-1}$. The inflation factor $(1 - \delta\ell_i/2)^{-1}$ diverges as $\delta\ell_i \to 2^{-}$, Euler therefore trades mixing speed against stationary accuracy even after preconditioning; instability occurs at $\delta\ell_i\geq2$.
\end{proposition}

\begin{proof}
In whitened coordinates $z = Q^\top M^{1/2}x$, where $Q$ diagonalises $M^{-1/2} H M^{-1/2} = Q\operatorname{diag}(\ell)Q^\top$, the iteration decouples into scalar AR(1) processes $z_{n+1,i} = (1 - \delta\ell_i)\,z_{n,i} + \sqrt{2\delta}\,\eta_{n,i}$ with $\eta_{n,i} \sim \mathcal{N}(0, 1)$ independent. The stationary variance of an AR(1) $Z_{n+1} = rZ_n + \sigma\eta_n$ is $\sigma^2/(1-r^2)$; substituting $r = 1 - \delta\ell_i$ and $\sigma^2 = 2\delta$ and simplifying via $1 - r^2 = \delta\ell_i(2 - \delta\ell_i)$ yields~\eqref{eq:ula-inflation}.
\end{proof}

\subsection{The \texorpdfstring{$M$}{M}-norm Moreau envelope}
\label{sec:moreau}

To handle a non-smooth $g$, we do not smooth it in the Euclidean metric---as MYULA~\citep{durmus2018efficient} does---but in the \emph{preconditioned} $M$-metric. For a parameter $\lambda > 0$, define
\begin{equation}
    g^{\lambda, M}(x) \;\coloneqq\; \min_{z \in \mathbb{R}^d}\,\Big\{\, g(z) + \tfrac{1}{2\lambda}\|z - x\|_M^2 \,\Big\},
    \label{eq:Mnorm-moreau}
\end{equation}
the \emph{$M$-norm Moreau envelope} of $g$ with parameter $\lambda$. This envelope is a $C^{1,1}$ convex surrogate for $g$ with the key gradient-Lipschitz property
\begin{equation}
    \big\|\nabla g^{\lambda,M}(x) - \nabla g^{\lambda,M}(y)\big\|_{M^{-1}} \;\leq\; \tfrac{1}{\lambda}\,\|x - y\|_M,
    \label{eq:Mmoreau-lip}
\end{equation}
i.e.\ $\nabla g^{\lambda, M}$ is $(1/\lambda)$-Lipschitz \emph{in the $M$-metric} (\citealt{bauschke2011convex}, Prop.~12.27, applied to the $M$-Hilbert space $(\mathbb{R}^d, \langle\cdot,\cdot\rangle_M)$). Thus the smoothed potential $U_\lambda \coloneqq f + g^{\lambda, M}$ is $m_M$-strongly convex and $(L_M + 1/\lambda)$-smooth in the $M$-norm, with effective condition number $\kappa_{\mathrm{eff}}^\lambda = (L_M + 1/\lambda)/m_M$. The surrogate target is $\pi_\lambda \propto \exp(-U_\lambda)$; the Moreau bias $W_2^M(\pi_\lambda, \pi)$ is controlled as Term~III of Theorem~\ref{thm:end-to-end}.

The explicit preconditioned comparator SP-MYULA uses the same surrogate:
\[
X_{n+1}=X_n-\delta M^{-1}\nabla U_\lambda(X_n)+\sqrt{2\delta}\,M^{-1/2}\xi_n.
\]
Its smoothing, discretisation and inner-solve errors are distinct.

\subsection{Where P-IMLA sits in the landscape}
\label{sec:landscape}

Table~\ref{tab:landscape} plots the proximal-Langevin landscape along the two axes of Figure~\ref{fig:2x2}: discretisation of the underlying Langevin diffusion, and geometry imposed on it by a preconditioner. On the explicit-Euler side, MYULA \citep{durmus2018efficient} and SKROCK \citep{pereyra2020accelerating} achieve $\kappa_f$-dependent rates with no preconditioning, and the sole prior method at the ``non-trivial $M$ + non-smooth $g$'' intersection, Corbineau P-ULA \citep{corbineau2019preconditioned}, retains explicit Euler---so inherits~\eqref{eq:budget-euler}'s variance inflation---and comes with no mixing-rate theory. On the implicit side, \citet{klatzer2024accelerated} obtain the sharp $\tilde{O}(\sqrt{\kappa_f})$ rate but uniformly fix $M = I$; the semi-implicit variant pCNL \citep{cotter2013mcmc} treats only the prior covariance implicitly, is bound to Gaussian-prior function-space formulations, and is always Metropolis-corrected. P-IMLA, to our knowledge, is the first to combine a matrix preconditioner, implicit-midpoint discretisation, and $M$-metric Moreau treatment of nonsmooth priors; its rate $\tilde{O}(\sqrt{\kappa_{\mathrm{eff}}})$ (Theorem~\ref{thm:contraction}) improves on \citet{klatzer2024accelerated}'s $\tilde{O}(\sqrt{\kappa_f})$ whenever $\kappa_{\mathrm{eff}} < \kappa_f$.

\begin{table}[h]
\centering
\caption{\textbf{The proximal-Langevin landscape.} Rates reproduce the cited settings and are not uniform comparisons of total work. For IMLA and P-IMLA, the displayed $\widetilde O(\sqrt\kappa)$ is contraction to the discretised invariant law; accuracy relative to the target also requires bias control (Rem.~\ref{rem:precision-protocol}). ``None'' means no corresponding rate stated here; $I$-Moreau denotes Euclidean smoothing.}
\label{tab:landscape}
\small
\setlength{\tabcolsep}{6pt}
\renewcommand{\arraystretch}{1.2}
\begin{tabular}{lcccc}
\toprule
Method        & Discretisation            & $M$                         & Non-smooth $g$         & Rate / setting \\
\midrule
ULA           & Expl.\ Euler              & $I$                         & ---                    & $\tilde{O}(\kappa_f / \varepsilon^2)$ \\
MYULA         & Expl.\ Euler              & $I$                         & $I$-Moreau             & $\tilde{O}(\kappa_\lambda / \varepsilon^2)$ \\
SKROCK        & Expl.\ stabilised         & $I$                         & $I$-Moreau             & $\tilde{O}(\sqrt{\kappa_\lambda})$, Gaussian \\
P-ULA         & Expl.\ Euler              & $f$ only                    & $I$-Moreau             & \emph{none} \\
RMMALA        & Euler + MH                & position-dep.               & ---                    & \emph{none} \\
Fisher MALA   & Euler + MH                & adaptive                    & ---                    & \emph{none} \\
pCNL          & Semi-impl.\ + MH          & prior cov.                  & ---                    & dim-free, Gaussian prior \\
IMLA  & \textbf{Impl.\ midpoint}  & $I$                         & $I$-Moreau             & $\tilde{O}(\sqrt{\kappa_f})$ \\
\midrule
\rowcolor{green!8}\textbf{P-IMLA (ours)} & \textbf{Impl.\ midpoint} & \textbf{general $\succ 0$} & \textbf{$M$-Moreau} & $\boldsymbol{\tilde{O}(\sqrt{\kappa_{\mathrm{eff}}})}$ \\
\bottomrule
\end{tabular}
\end{table}

%% file: sections/algorithm.tex

\section{The P-IMLA Algorithm}
\label{sec:algorithm}

\textbf{Smooth case.} Fix $\delta > 0$ and a constant SPD preconditioner $M \succ 0$. For smooth targets ($g \equiv 0$), the \emph{Preconditioned Implicit Midpoint Langevin Algorithm} is the one-step kernel
\begin{equation}
X_{n+1} \;=\; -X_n \;+\; 2\,\mathrm{prox}^{M}_{(\delta/2) f}\!\left( X_n + \tfrac{\sqrt{2\delta}}{2}\,M^{-1/2}\xi_n \right),
\qquad \xi_n \stackrel{\mathrm{iid}}{\sim} \mathcal{N}(0, I),
\label{eq:pimla-smooth}
\end{equation}
with $\mathrm{prox}^M_{\tau f}(y) \coloneqq \arg\min_x \{ f(x) + \tfrac{1}{2\tau}\|x - y\|_M^2 \}$. At $M = I$ the update recovers IMLA \citep{klatzer2024accelerated}; for quadratic $f$, at $M=H=\nabla^2f$ (or a fixed spectrally equivalent surrogate) it realises the matrix-preconditioned variant introduced here. We use a single fixed $M$, as in \citet{titsias2024optimal,hird2025preconditioning}, rather than the position-dependent metric of \citet{girolami2011riemann}---a simplification essential for the $M$-norm Moreau construction (§\ref{sec:moreau}) and the spectral analysis of §\ref{sec:theory}.

\textbf{Why this is implicit midpoint.} Let $p \coloneqq \mathrm{prox}^M_{(\delta/2)f}(w)$ with $w \coloneqq X_n + \tfrac{\sqrt{2\delta}}{2}M^{-1/2}\xi_n$. The proximal first-order condition
\begin{equation}
p \;=\; w - \tfrac{\delta}{2}M^{-1}\nabla f(p)
\label{eq:prox-foc}
\end{equation}
combined with $X_{n+1} = 2p - X_n$ gives
\begin{equation}
X_{n+1} \;=\; X_n \;-\; \delta\, M^{-1}\nabla f\!\bigg(\tfrac{X_n + X_{n+1}}{2}\bigg) \;+\; \sqrt{2\delta}\,M^{-1/2}\xi_n,
\label{eq:pimla-midpoint}
\end{equation}
since $p = (X_n + X_{n+1})/2$: the proximal operator \emph{is} the solution of the midpoint equation \eqref{eq:pimla-midpoint}, repackaged as a single off-the-shelf call. The proximal is strongly convex in $x$ for every $\delta > 0$, so no step-size restriction $\delta < 2/L_M$ arises. When $f$ is quadratic, $f(x) = \tfrac{1}{2}x^\top H x - h_{\mathrm{data}}^\top x$ with $h_{\mathrm{data}} = A^\top y/\sigma_{\mathrm{noise}}^2$ (Gaussian likelihood, forward operator $A$); when additionally $H, M$ are Fourier-diagonal, the proximal admits a closed form costing $O(d\log d)$ per iteration.

\textbf{Non-smooth case: P-IMLA$_\lambda$.} For $\pi \propto \exp(-(f + g))$ with $g$ proper convex lsc, run P-IMLA on the Moreau-smoothed $U_\lambda \coloneqq f + g^{\lambda, M}$ from \eqref{eq:Mnorm-moreau}, in the spirit of \citet{pereyra2016proximal,durmus2018efficient}'s MYULA but with $M$-metric Moreau and implicit-midpoint discretisation, yielding
\begin{equation}
X_{n+1} \;=\; -X_n \;+\; 2\,\mathrm{prox}^{M}_{(\delta/2) U_\lambda}\!\left( X_n + \tfrac{\sqrt{2\delta}}{2}\,M^{-1/2}\xi_n \right).
\label{eq:pimla-lambda}
\end{equation}
$U_\lambda$ is $C^{1,1}$ with $M$-metric Lipschitz constant $L_M + 1/\lambda$ \eqref{eq:Mmoreau-lip}, so Theorem~\ref{thm:contraction} applies; Theorem~\ref{thm:bea-bias} requires additional smoothness and Poisson regularity; the gap $W_2^M(\pi_\lambda, \pi)$ is controlled by Theorem~\ref{thm:end-to-end} Term~III. The inner problem is the proximal of the sum $U_\lambda$. With $\tau \coloneqq \delta/2$, differentiating \eqref{eq:Mnorm-moreau} at its unique minimizer gives $\nabla g^{\lambda,M}(p) = (M/\lambda)(p-u)$, where $u = \mathrm{prox}^M_{\lambda g}(p)$. The first-order condition $0 = \nabla f(p) + (M/\lambda)(p-u) + (M/\tau)(p-w)$ for $p = \mathrm{prox}^{M}_{\tau U_\lambda}(w)$ gives
\begin{equation}
\bigl(\tau\hat{H} + (1 + \tau/\lambda)\hat{M}\bigr)\,\hat{p}
\;=\; \tau\,\hat{h}_{\mathrm{data}} + \hat{M}\hat{w} + (\tau/\lambda)\,\hat{M}\,\widehat{\mathrm{prox}^M_{\lambda g}(p)}.
\label{eq:picard-fixed-point}
\end{equation}
The LHS is a closed-form solve when $H, M$ are circulant (so $\hat{H}, \hat{M}$ are diagonal); for anisotropic TV, $g(x)=\rho_{\mathrm{TV}}\|Dx\|_1$, the inner $M$-norm proximal is computed by a 2-block ADMM with the constraint $z=Dx$ and soft-thresholding in $z$ \citep{boyd2011distributed}. We solve \eqref{eq:picard-fixed-point} by Picard iteration on $p$, giving Algorithm~\ref{alg:picard}.

\begin{algorithm}[t]
\caption{Picard inner solver for $p = \mathrm{prox}^M_{(\delta/2) U_\lambda}(w)$ with $U_\lambda = f + g^{\lambda,M}$.}
\label{alg:picard}
\begin{algorithmic}[1]
\Require $w \in \mathbb{R}^d$, parameters $\delta, \lambda$ (set $\tau \coloneqq \delta/2$), Fourier-diagonal $\hat{M}, \hat{H}, \hat{h}_{\mathrm{data}}$, periodic forward-difference operator $D$, ADMM penalty $\rho_a$, Picard steps $K_{\mathrm{P}}$, ADMM steps $K_{\mathrm{in}}$
\State Cache $\hat{D}_{\mathrm{denom}} \gets \tau\hat{H} + (1+\tau/\lambda)\hat{M}$, $\hat{w} \gets \mathcal{F}(w)$;\quad warm-start $x$
\For{$k = 1, \ldots, K_{\mathrm{P}}$}
    \State $u \gets \mathrm{prox}^{M}_{\lambda g}(x)$ via 2-block ADMM (warm-started; $K_{\mathrm{in}}$ FFTs + soft-threshold)
    \State $\hat{u} \gets \mathcal{F}(u)$;\quad $x \gets \mathcal{F}^{-1}\!\Bigl(\hat{D}_{\mathrm{denom}}^{-1}\!\bigl(\tau\,\hat{h}_{\mathrm{data}} + \hat{M}\hat{w} + (\tau/\lambda)\hat{M}\hat{u}\bigr)\Bigr)$
\EndFor
\State \Return $p \gets x$
\end{algorithmic}
\end{algorithm}

With exact inner proximals, the Picard map is a contraction with factor $(\tau/\lambda)/(1+\tau m_M+\tau/\lambda)=0.929$ at the imaging parameters ($m_M=1$, $\delta=0.141$, $\lambda=5\times10^{-3}$). Warm starts reduce the initial residual; a fixed sweep count alone is not an accuracy certificate. Per-outer cost is $O(K_{\mathrm P}K_{\mathrm{in}}d\log d)$, with the TV proximal computed inexactly in practice.

\textbf{Engineering optimization (JP-IMLA).} A practical variant \emph{JP-IMLA} replaces the nested Picard solve with a single 2-block ADMM evaluating $\mathrm{prox}^M_{\tau(f+g)}$ (App.~\ref{app:jp-equivalence}); its reported finite-budget SSE differs from the Picard result by at most $0.7\%$ across imaging phantoms (Table~\ref{tab:jp-vs-true}) at $\sim\!60\!\times$ per-iter speedup. The main experiments use Picard; the ideal theorems concern exact proximal updates. Appendix~\ref{app:jp-equivalence} separates exact-kernel and finite-solver guarantees; Table~\ref{tab:jpcr-budget} reports separate budget timings and pointwise certificates, which do not certify every production update.

%% file: sections/theory.tex

\section{The Theory of P-IMLA}
\label{sec:theory}

The theoretical analysis chains four results, all proved in Appendix~\ref{app:proofs}. Theorem~\ref{thm:gaussian-exact} establishes Gaussian exactness at any step size. Theorem~\ref{thm:contraction} gives the corresponding reflected-proximal contraction for strongly convex $C^{1,1}$ potentials; for fixed $M$, this is the sharp IMLA factor in whitened coordinates \citep{klatzer2024accelerated}. Theorem~\ref{thm:bea-bias} characterises the leading stationary bias via backward error analysis (BEA): its defect is a \emph{first-order differential operator}, whereas P-ULA's also contains a second-order term. Theorem~\ref{thm:end-to-end} gives an error budget for non-smooth targets. Write $r(\ell)=(1-\delta\ell/2)/(1+\delta\ell/2)$ for a spectral value $\ell$; $\lambda$ denotes the Moreau parameter.

\subsection{Gaussian exactness and Cayley contraction}
\label{sec:exact-gaussian}

\begin{theorem}[Gaussian exactness]
\label{thm:gaussian-exact}
Let $f(x) = \tfrac{1}{2}x^\top H x - h^\top x$ with $H \succ 0$, $g \equiv 0$, $\pi = \mathcal{N}(H^{-1}h,\, H^{-1})$, and preconditioner $M \succ 0$ with $\kappa_{\mathrm{eff}} = L_M/m_M$ as in \eqref{eq:kappa-eff}. For every $\delta > 0$, the P-IMLA iteration \eqref{eq:pimla-smooth} has $\pi$ as its unique stationary distribution, and contracts as $W_2^{M}(\mathrm{Law}(X_N),\,\pi) \leq \rho(\delta)^{N}\, W_2^{M}(\mu_0,\pi)$ with $\rho(\delta) = \max\{|r(m_M)|,\,|r(L_M)|\}$. The minimax-optimal step size $\delta^\star = 2/\sqrt{m_M L_M}$ achieves $\rho^\star = (\sqrt{\kappa_{\mathrm{eff}}}-1)/(\sqrt{\kappa_{\mathrm{eff}}}+1)$, giving mixing complexity $N = O(\sqrt{\kappa_{\mathrm{eff}}}\,\log(1/\varepsilon))$.
\end{theorem}

\subsection{Contraction of the reflected proximal map}
\label{sec:contraction}

\begin{theorem}[$M$-norm Wasserstein contraction]
\label{thm:contraction}
Let $f \in C^{1,1}(\mathbb{R}^d)$ be $m_M$-strongly convex and $L_M$-smooth in the $M$-norm, $g \equiv 0$. For every $\delta > 0$, two P-IMLA chains $X_n, Y_n$ driven by shared Gaussian noise satisfy $\|X_{n+1} - Y_{n+1}\|_M \le \rho(\delta)\,\|X_n - Y_n\|_M$ almost surely, with $\rho(\delta)$ as in Theorem~\ref{thm:gaussian-exact}; the bound is attained by quadratic $f$ (Gaussian worst case). Consequently P-IMLA admits a unique stationary distribution $\pi_\delta$ and $W_2^M(\mathrm{Law}(X_N), \pi_\delta) \leq \rho(\delta)^N\, W_2^M(\mu_0, \pi_\delta)$.
\end{theorem}

\begin{remark}[Contraction and empirical convergence]
\label{rem:thm2-sharpness}
The bound controls distance to the discrete invariant law, not a threshold for $\widehat R$ or reconstruction error. The diagnostics in §\ref{sec:experiments} measure empirical convergence separately. Gaussian exactness also holds at $M=I$; preconditioning improves the finite-budget contraction factor through $\kappa_{\mathrm{eff}}$.
\end{remark}

\subsection{Sharp drift-only bias via backward error analysis}
\label{sec:bea}

\begin{theorem}[Drift-only BEA bias]
\label{thm:bea-bias}
Let $g\equiv0$, and let $f\in C^6(\mathbb R^d)$ be globally $m_M$-strongly convex and $L_M$-smooth in the fixed $M$-norm, with derivatives of orders $3$--$6$ of at most polynomial growth. For an observable $\varphi$, assume that $\mathcal L u_\varphi=-(\varphi-\pi(\varphi))$ has a solution $u_\varphi\in C^6_{\mathrm{pol}}$, where $\mathcal L u=-M^{-1}\nabla f\cdot\nabla u+M^{-1}:\nabla^2u$ and $C^6_{\mathrm{pol}}$ means polynomial growth of the function and its derivatives through order six. Define $h(x)\coloneqq\operatorname{tr}(M^{-1}\nabla^2f(x))$. Then, as $\delta\downarrow0$,
\begin{equation}
    \pi_\delta(\varphi) - \pi(\varphi) \;=\; \tfrac{\delta}{4}\,\operatorname{Cov}_\pi\!\bigl(h,\, \varphi\bigr) \;+\; o(\delta).
    \label{eq:bea-weak}
\end{equation}
The required $\delta$-uniform polynomial moments are proved in Lemma~\ref{lem:uniform-moments}. On Gaussian targets, $h$ is constant and the coefficient vanishes, consistently with Theorem~\ref{thm:gaussian-exact}.
\end{theorem}

\emph{Why drift-only.} The weighted expansion gives $P_\delta u=u+\delta\mathcal Lu+\delta^2(\mathcal L^2u/2+B_2u)+o(\delta^2)$. The proximal Jacobian identity $SM^{-1}S^\top=M^{-1}-\delta\Gamma+O(\delta^2)$, with $S=(I+(\delta/2)M^{-1}\nabla^2f)^{-1}$ and $\Gamma=M^{-1}\nabla^2fM^{-1}$, cancels the second-order modification, leaving $B_2u=\tfrac14(M^{-1}\nabla h)\cdot\nabla u$. P-ULA retains a diffusion modification (Appendix~\ref{app:pula-bea}); both schemes generally have $O(\delta)$ weak invariant bias. Theorem~\ref{thm:bea-bias} does not assume a density expansion: the leading $W_2^M$ coefficient is conditional on the additional $L^3(\pi)$ condition in Proposition~\ref{prop:conditional-density}. Under stronger $C^8$ and two-level Poisson regularity, Corollary~\ref{cor:richardson} gives second-order Richardson cancellation. A TV Moreau envelope is generally only $C^{1,1}$, so its guarantee below does not invoke this BEA theorem.

\subsection{End-to-end error budget for non-smooth targets}
\label{sec:end-to-end}

\begin{theorem}[End-to-end $W_2^M$ bound]
\label{thm:end-to-end}
Let $\pi\propto\exp(-f-g)$ with $f$ as in Assumption~\ref{ass:target} and $g$ convex and globally $L_g^M$-Lipschitz in the $M$-norm. Set $U_\lambda=f+g^{\lambda,M}$ and $L_\lambda=L_M+1/\lambda$. For $\lambda>0$, $0<\delta\le2/L_\lambda$ and $N\ge1$, the exact P-IMLA$_\lambda$ kernel satisfies
\begin{equation}
    W_2^{M}\!\bigl(\mathrm{Law}(X_N),\, \pi\bigr) \;\leq\; \underbrace{\rho(\delta,\lambda)^{N}\, W_0}_{\textnormal{\rm(I) mixing}} \;+\; \underbrace{\frac{C_1(\lambda)\sqrt{d}\,\delta^{3/2}+\bar C_2(\lambda)\delta^2}{1 - \rho(\delta,\lambda)}}_{\textnormal{\rm(II) discretisation}} \;+\; \underbrace{L_g^M\,\sqrt{\lambda / m_M}}_{\textnormal{\rm(III) Moreau}},
    \label{eq:end-to-end}
\end{equation}
where $W_0=W_2^M(\mu_0,\pi_{\delta,\lambda})$, $\rho(\delta,\lambda)=\max\{|r(m_M)|,|r(L_\lambda)|\}$, $C_1(\lambda)=L_\lambda/\sqrt2$, and one may take
\[
\bar C_2(\lambda)=L_\lambda^2\left(\tfrac32\sqrt{10d/m_M}+\tfrac12\sqrt{d/L_\lambda}\right).
\]
The terms are coupled through $\rho$ and their $\lambda$-dependent constants.
\end{theorem}

\begin{remark}[Precision protocol]
\label{rem:precision-protocol}
At fixed $\lambda$, the all-step contraction theorem gives $N=O(\sqrt{\kappa_{\mathrm{eff}}^\lambda}\log(W_0/\varepsilon))$ to the \emph{discrete invariant law} at $\delta_\lambda^\star=2/\sqrt{m_ML_\lambda}$, where $\kappa_{\mathrm{eff}}^\lambda=L_\lambda/m_M$. This optimum generally exceeds the step range of Theorem~\ref{thm:end-to-end}. Full $\varepsilon$-accuracy to $\pi$ requires controlling the discretisation and Moreau terms as well; the displayed contraction complexity alone does not provide that guarantee. At fixed $\lambda$, the two accumulated discretisation terms are $O(\sqrt\delta)$ and $O(\delta)$ as $\delta\downarrow0$.
\end{remark}

\begin{remark}[Gaussian duality and polyhedral tightness]
\label{rem:duality}
Quadratic targets attain the contraction bound and have zero stationary bias; at $M=H$ and $\delta=2$ the update produces an independent exact Gaussian sample. For Term~III, the generic $\sqrt\lambda$ bound can be conservative: 1-D quantile integration on $f(x)=x^2/2$ with the polyhedral penalties $|x|$ and $\max(2x,-x)$ gives empirical $W_2^M$-vs-$\lambda$ slopes $1.95$ and $0.97$ (Appendix~\ref{app:moreau-bias}). These examples do not establish a uniform improved rate for all polyhedral penalties.
\end{remark}

%% file: sections/jp_guarantees.tex
\subsection{Guarantees for JP-IMLA}
\label{sec:jp-guarantees}
Let $P$ denote the canonical P-IMLA$_\lambda$ kernel and $Q$ the exact
JP-IMLA kernel, obtained by replacing $\operatorname{prox}_{\tau U_\lambda}^M$
with $\operatorname{prox}_{\tau(f+g)}^M$, $\tau=\delta/2$.

\begin{proposition}[JP model and solver perturbations]
\label{prop:jp-main}
Under the hypotheses on $f,g$ of Theorem~\ref{thm:end-to-end}, fix
$\delta,\lambda>0$ and write $\rho=\rho(\delta,\lambda)$.
For joint and smoothed proximal points $p,q$ at a common input,
and $\mu,\nu\in\mathcal P_2$,
\begin{equation}
\begin{gathered}
 \varepsilon_{\rm model}=2\lambda L_g^M
 \frac{\tau^{-1}+L_M}{\tau^{-1}+m_M},
 \qquad \|p-q\|_M\le\varepsilon_{\rm model}/2,\\
 W_2^M(Q\mu,P\nu)
 \le\rho W_2^M(\mu,\nu)+\varepsilon_{\rm model}.
\end{gathered}
\label{eq:jp-main-gap}
\end{equation}
For a finite-ADMM chain $\widehat X_j$ satisfying
Proposition~\ref{prop:jpcr-certificate}, with fresh noise independent of the
past and possibly history-dependent warm starts, let $U_{K,j}$ denote
its certificate at the actual step input.
Set $e_{K,j}=(8\mathbb E U_{K,j}/(m_M+\tau^{-1}))^{1/2}$ and
$D_j=W_2^M(\mathcal L(\widehat X_j),\pi_{\delta,\lambda})$,
assuming these quantities are finite. Then
\begin{equation}
\begin{aligned}
 D_{j+1}&\le\rho D_j+\varepsilon_{\rm model}+e_{K,j},\\
 D_N&\le\rho^N D_0+
 \sum_{j=0}^{N-1}\rho^{N-1-j}
       (\varepsilon_{\rm model}+e_{K,j}).
\end{aligned}
\label{eq:jp-main-recursion}
\end{equation}
Every invariant law $\nu_Q\in\mathcal P_2$ of $Q$ satisfies
$W_2^M(\nu_Q,\pi_{\delta,\lambda})
\le\varepsilon_{\rm model}/(1-\rho)$.
\end{proposition}

\emph{Proof sketch.}
Monotonicity of $\partial g$ gives the proximal gap; reflection and
shared-noise coupling combine it with canonical contraction.
The certificate and Minkowski's inequality give the solver recursion;
complete proofs are in Appendix~\ref{app:jp-theory}.

Adding the discretisation and Moreau terms of
Theorem~\ref{thm:end-to-end} gives the original-target bound
\eqref{eq:jpcr-end-to-end} when $0<\delta\le2/L_\lambda$;
the perturbation and solver bounds above hold for every $\delta>0$.
The exact $Q$ has an invariant law, unique in $\mathcal P_2$ for quadratic
$f$ and finite $g=\sum_e\rho_e|d_e^\top x|$, $\rho_e>0$, including our imaging model
(Theorem~\ref{thm:jpcr-unique}).
This supplies no mixing rate or uniqueness claim for finite ADMM.
Appendix~\ref{app:jpcr-audit} reports the corresponding numerical audits.

%% file: sections/experiments.tex

\section{Experiments}
\label{sec:experiments}

We compare samplers on $256\!\times\!256$ TV deconvolution using geometric, Shepp--Logan and cameraman phantoms. Appendix~\ref{app:experiments} reports Gaussian variance and contraction checks, nonquadratic bias stress tests, Moreau-bias measurements and step-size sweeps. A separate $64\!\times\!64$ random fan-beam experiment tests a non-Fourier forward operator (Appendix~\ref{app:exp-radial}). Appendix~\ref{app:rebuttal-experiments} adds a $d=256$ bias-coefficient experiment satisfying the potential assumptions of Theorem~\ref{thm:bea-bias}, and separates convergence diagnostics from reconstruction quality and posterior texture. For TV deconvolution we use $M = \hat{H}$, giving $\kappa_{\mathrm{eff}}^\lambda = 1 + 1/\lambda$ for the Moreau-smoothed composite~\eqref{eq:pimla-lambda}.

\textbf{Setup.} A $256\!\times\!256$ phantom under Gaussian blur ($\sigma_{\mathrm{blur}} = 2.0$, $\sigma_{\mathrm{noise}} = 0.05$); $f$ quadratic with $\kappa_f = 4\!\times\!10^{5}$, $g = \rho_{\mathrm{TV}}\|D\!\cdot\|_1$ \citep[anisotropic TV in the spirit of][]{rudin1992nonlinear} with $\rho_{\mathrm{TV}} = 0.3$, $\lambda = 5\!\times\!10^{-3}$, giving $\kappa_{\mathrm{eff}}^\lambda = 201$. We compare four samplers under a fixed $100\!\times$ iteration allocation: ours $N_{\mathrm{ours}} = 2{,}000$ via Algorithm~\ref{alg:picard} ($K_{\mathrm{P}}=30$, $K_{\mathrm{in}}=50$, $\rho_a=200$, $\delta=0.141$); SP-MYULA, MYULA, SK-ROCK \citep[$s=10$,][]{pereyra2020accelerating} each $N_{\mathrm{base}}=2\!\times\!10^{5}$ at largest stable step (App.~\ref{app:exp-baseline-tuning}). Eight chains per sampler start from $X_0 = 0$, with averages accumulated without burn-in, on an NVIDIA RTX~PRO~6000 Blackwell ($96$\,GB) with $4$-process parallelisation. Actual per-chain wall-clocks are reported in Table~\ref{tab:imaging}; the iteration allocation does not make all wall-clocks equal.

\begin{figure}[!t]
\centering
\includegraphics[width=0.7\linewidth]{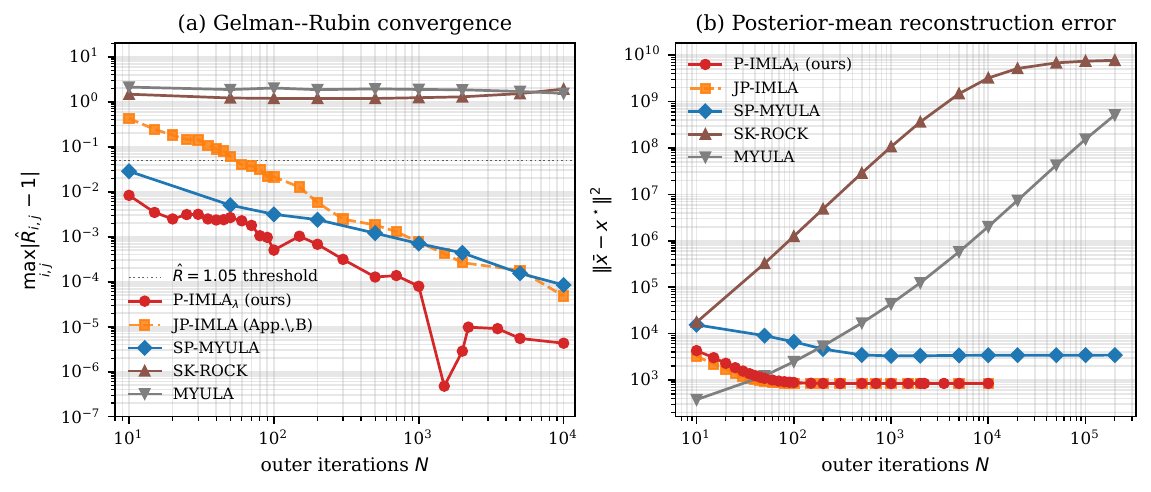}
\caption{\textbf{Classical per-pixel Gelman--Rubin trajectory at $256\!\times\!256$.} Eight chains from the same zero start; (a) $\max_{i,j}|\hat R_{i,j}-1|$ on log scale to $N=10{,}000$; (b) extends to $N=2\!\times\!10^5$. P-IMLA$_\lambda$ reaches $\hat R<1.05$ at $N=10$ and its per-chain SSE changes from $855$ to $846$ by $N=10{,}000$. SP-MYULA also reaches $\hat R\leq1.001$, despite larger reconstruction error; MYULA and SK-ROCK end at $2.41$ and $11.07$. These same-start diagnostics \citep{gelman1992inference} are complemented by overdispersed, rank-normalized split diagnostics in Appendix~\ref{app:overdispersed-diagnostics}.}
\label{fig:mixing-trajectory}
\end{figure}

\textbf{Diagnostics and matched-budget comparison.} The classical per-pixel $\hat R$ in Figure~\ref{fig:mixing-trajectory} reaches $1.008$ at $N=10$ and $\leq1.001$ at $N=2{,}000$ for P-IMLA$_\lambda$; an $N=10{,}000$ extension (ours and JP-IMLA, on the same hardware) shows stable reconstruction SSE. Because these chains share $X_0=0$, this diagnostic alone does not establish convergence; Theorem~\ref{thm:contraction} bounds Wasserstein contraction, not a threshold-crossing time for $\hat R$. Table~\ref{tab:imaging} separately reports reconstruction SSE $=\|\bar{x}-x^\star\|^2$ ($\pm$\,SE): P-IMLA$_\lambda$ $855.22\pm0.66$ (per-pixel RMSE $\approx0.11$), versus $3{,}432$ for SP-MYULA. This approximately fourfold difference is a cold-start, finite-budget result, not a measurement of stationary bias. MYULA / SK-ROCK have SSE $\sim10^9,10^{10}$ and $\hat R=2.41,11.07$.

\begin{table}[h]
\centering
\small
\caption{\textbf{$256\!\times\!256$ TV deconvolution, geometric phantom, fixed iteration budgets.} $\rho_{\mathrm{TV}}\!=\!0.3$, $\sigma_{\mathrm{noise}}\!=\!0.05$, $\lambda\!=\!5\!\times\!10^{-3}$; RTX PRO 6000 Blackwell, 4-way parallel. SSE is per-chain $\|\bar{x}_c - x^\star\|^2$ averaged across $K$ chains $\pm$ SE, including the initial transient. $\hat R$ is classical per-pixel, from common zero starts. These are reconstruction and implementation-budget comparisons; joint-kernel guarantees are stated separately in the appendix. Cross-phantom results appear in Table~\ref{tab:cross-phantom}.}
\label{tab:imaging}
\setlength{\tabcolsep}{6pt}
\renewcommand{\arraystretch}{0.95}
\begin{tabular}{@{}lccccc@{}}
\toprule
Method & SSE at final $N$ & $K$ & $N_{\mathrm{final}}$ & Wall-clock/chain & $\hat{R}$ at $N_{\mathrm{final}}$ \\
\midrule
MYULA                           & $\sim\!10^{9}$ (no mix)      & 8 & $2\!\times\!10^{5}$ & $43$\,min  & $2.41$ \\
SK-ROCK ($s\!=\!10$)            & $\sim\!10^{10}$ (no mix)     & 8 & $2\!\times\!10^{5}$ & $425$\,min & $11.07$ \\
SP-MYULA                        & $3{,}432.0 \pm 1.0$           & 8 & $2\!\times\!10^{5}$ & $74$\,min  & $\leq\!1.001$ \\
\textbf{P-IMLA$_\lambda$ (ours)} & $\mathbf{855.22 \!\pm\! 0.66}$ & 8 & $\mathbf{2{,}000}$  & $\mathbf{70}$\,\textbf{min} & $\mathbf{\leq\!1.001}$ \\
\textbf{JP-IMLA (variant, §\ref{sec:algorithm})}  & $\mathbf{852.72 \pm 0.65}$ & 8 & $\mathbf{2{,}000}$    & $\mathbf{1.2}$\,\textbf{min}  & $\mathbf{\leq\!1.001}$ \\
\bottomrule
\end{tabular}
\end{table}

\begin{figure}[!t]
\centering
\includegraphics[width=0.85\linewidth]{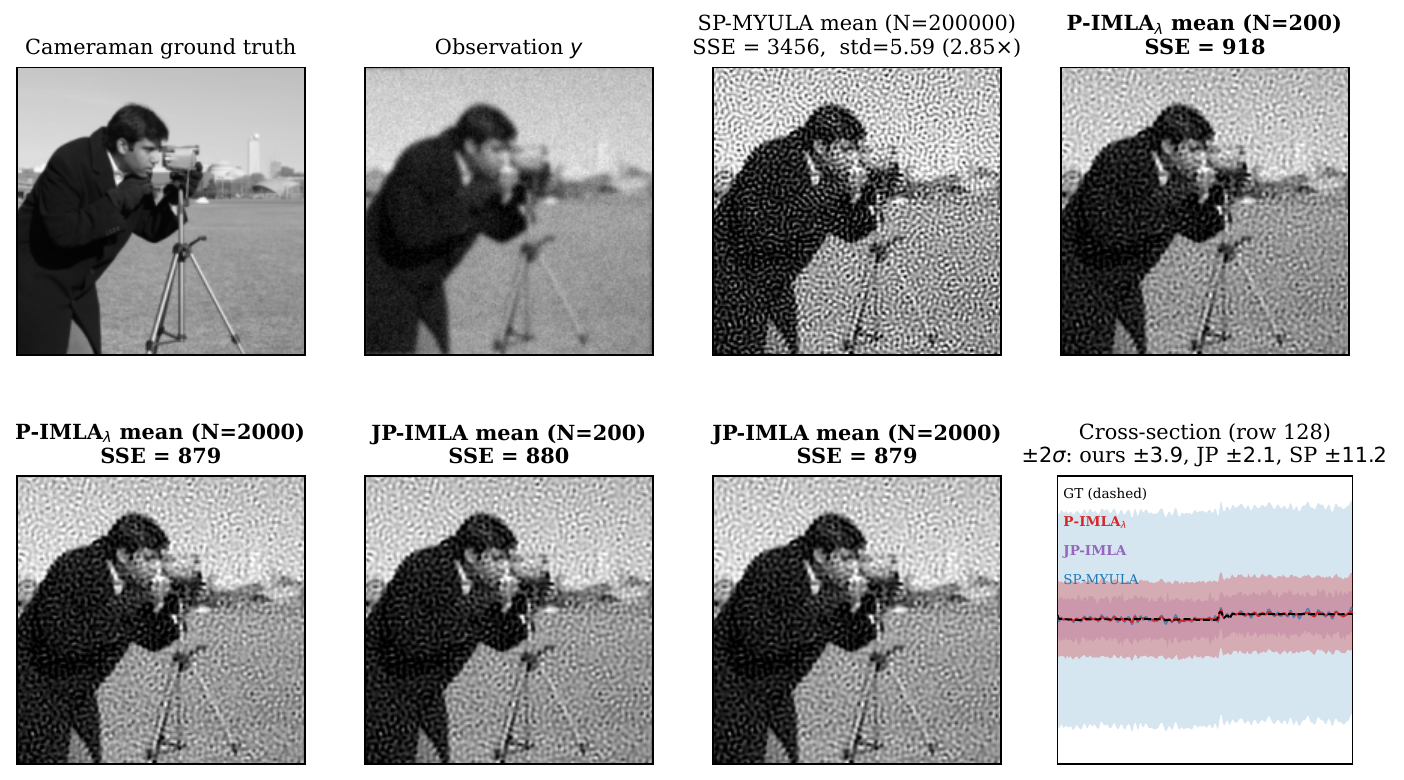}
\caption{\textbf{$256\!\times\!256$ cameraman posterior-mean reconstructions at fixed budgets.} Images are averages across chains, not individual posterior draws. R1: GT, observation, SP-MYULA at $N\!=\!2\!\times\!10^{5}$ (R1C3; pooled-mean SSE $3{,}456$, mean pixel std $\bar s=5.59$), and P-IMLA$_\lambda$ at $N=200$ (R1C4; SSE $918$). R2C1: P-IMLA$_\lambda$ at $N=2{,}000$ (SSE $879$, $\bar s=1.96$). R2C2--C3: JP-IMLA (§\ref{sec:algorithm}) at $N=200$ (SSE $880$) and $N=2{,}000$ (SSE $879$), with the historical $\sim60\!\times$ per-iteration implementation speedup. R2C4: row-$128$ cross-section with $\pm2$ posterior-standard-deviation bands. Pooled-mean SSE differs from the per-chain average in Table~\ref{tab:cross-phantom}. Appendix~\ref{app:roughness-decomposition} shows that stronger TV regularization reduces the means' visible texture.}
\label{fig:imaging-panel}
\end{figure}

\textbf{Cross-phantom robustness (Shepp--Logan, cameraman).} We extend the same finite-budget protocol to \emph{Shepp--Logan} and \emph{cameraman}, four chains each. Across all three phantoms ours achieves a $3.5\!\times\!\!-\!\!4.0\!\times$ reconstruction-SSE reduction over SP-MYULA with the $100\!\times$ iteration allocation (Table~\ref{tab:cross-phantom}). Changing burn-in, initialization or solver accuracy defines a different comparison; these ratios do not isolate invariant-measure error.

\textbf{Additional bias and roughness checks.} On a $16\!\times\!16$ log-cosh lattice satisfying Theorem~\ref{thm:bea-bias}'s potential assumptions, all nine coefficient estimates across three preconditioners and three observables agree with the prediction within their $95\%$ intervals. Separate overdispersed imaging runs distinguish high-frequency draw variability controlled by $\lambda$ from mid-frequency posterior-mean texture reduced by $\rho_{\mathrm{TV}}$; doubling $\rho_{\mathrm{TV}}$ improves PSNR by $3.5$\,dB (geometric) and $2.7$\,dB (cameraman). Independent exact-target Gibbs experiments at $32^2$ and $48^2$ retain the same texture without Moreau smoothing or time discretization. Full protocols, observable-specific diagnostics and their remaining slow modes are in Appendix~\ref{app:rebuttal-experiments}.

\begin{table}[H]
\centering
\small
\setlength{\tabcolsep}{6pt}
\renewcommand{\arraystretch}{0.95}
\caption{\textbf{Cross-phantom matched-budget SSE} (geometric: 8 chains; Shepp--Logan, cameraman: 4 chains; $\pm$\,SE; ours $N\!=\!2{,}000$, baselines $N\!=\!2\!\times\!10^{5}$). MYULA SSE $\sim\!10^{9}$ on all three.}
\label{tab:cross-phantom}
\begin{tabular}{@{}lccc@{}}
\toprule
Phantom & ours SSE & SP-MYULA SSE & SP-MYULA / ours \\
\midrule
geometric    & $855.22 \pm 0.66$     & $3{,}432.0 \pm 1.0$ & $4.01\!\times$ \\
Shepp--Logan & $1{,}013.99 \pm 0.81$ & $3{,}525.6 \pm 1.0$ & $3.48\!\times$ \\
cameraman    & $890.24 \pm 1.19$     & $3{,}460.4 \pm 2.0$ & $3.89\!\times$ \\
\bottomrule
\end{tabular}
\end{table}

%% file: sections/conclusion.tex
\section{Conclusion}
\label{sec:discussion}

Implicit midpoint with fixed matrix preconditioning combines Gaussian exactness with the sharp IMLA contraction factor in the effective condition number. The variance-budget identity~\eqref{eq:budget-midpoint} explains the Gaussian cancellation; the drift-only BEA coefficient identifies the leading weak bias beyond Gaussians under explicit smoothness assumptions. At imaging scale, P-IMLA$_\lambda$ improves finite-budget reconstruction and relaxation at the reported settings, while the diagnostic sweeps separate smoothing, solver and prior effects. The joint-proximal variant retains a substantial per-iteration speed advantage, with model and solver errors controlled separately by the kernel comparison and primal--dual certificate.

\textbf{Limitations.} The BEA expansion does not apply directly to a $C^{1,1}$ TV envelope, and the end-to-end accuracy bound uses a smaller step range than the all-step contraction theorem. Exact-JP uniqueness is proved for the imaging penalty class, not for arbitrary finite ADMM implementations. Small-image exact-posterior checks and overdispersed full-image diagnostics complement the original experiments but do not certify all full-image posterior moments. Position-dependent preconditioning and image-scale non-Fourier validation remain open; Appendix~\ref{app:limitations} gives the full scope.

%% file: sections/appendix.tex

\section{Full Proofs from Section~\ref{sec:theory}}
\label{app:proofs}

This appendix provides the full proofs of Theorems~\ref{thm:gaussian-exact}--\ref{thm:end-to-end} and the supporting technical lemmas referenced in Section~\ref{sec:theory}. The proofs are self-contained given the definitions and labeled equations of Sections~\ref{sec:background}--\ref{sec:theory}; no additional notation is introduced beyond the specifications below.

\paragraph{Standing conventions for the appendix.} We write $H = \nabla^2 f$, $A = M^{-1/2}HM^{-1/2}$, $\Gamma = M^{-1}HM^{-1}$, and $b = -M^{-1}\nabla f$; these are functions of the current point unless $f$ is quadratic. We use the Einstein summation convention on repeated indices and write $M^{-1}_{ij} = (M^{-1})_{ij}$. The spectral variable is $\ell$; $\lambda$ denotes Moreau smoothing.

\subsection{Proof of Theorem~\ref{thm:gaussian-exact} (Gaussian exactness)}
\label{app:thm1}

Recall the setting: $f(x) = \tfrac{1}{2}x^\top Hx - h^\top x$ with $H \succ 0$, $g \equiv 0$, $\pi = \mathcal{N}(\mu, H^{-1})$ with $\mu = H^{-1}h$, and $M \succ 0$. P-IMLA reads
\begin{equation}
X_{n+1} = -X_n + 2\,\mathrm{prox}^{M}_{(\delta/2)f}\!\Bigl(X_n + \tfrac{\sqrt{2\delta}}{2}\,M^{-1/2}\xi_n\Bigr),\qquad \xi_n \overset{\textrm{iid}}{\sim} \mathcal{N}(0,I).
\label{eq:app-pimla}
\end{equation}

\paragraph{Step 1 (Joint diagonalization and whitening).}
Let $A = M^{-1/2}HM^{-1/2} = Q\Lambda Q^\top$ be the spectral decomposition with $\Lambda = \mathrm{diag}(\ell_1,\dots,\ell_d)$ and $Q$ orthogonal. Introduce whitened coordinates
\[
Z_n \;\coloneqq\; Q^\top M^{1/2}(X_n - \mu).
\]
Under this change of variables: the $M$-norm becomes Euclidean, $\|X-\mu\|_M = \|Z\|$; the target becomes $\pi_Z = \bigotimes_{i=1}^d \mathcal{N}(0, \ell_i^{-1})$; and $W_2^M$ in $X$-coordinates equals Euclidean $W_2$ in $Z$-coordinates.

We claim that \eqref{eq:app-pimla} maps to $d$ independent scalar recurrences
\begin{equation}
Z_{n+1,i} \;=\; r_i\,Z_{n,i} \;+\; \frac{\sqrt{2\delta}}{1+a_i}\,\eta_{n,i},\qquad r_i \coloneqq \tfrac{1-a_i}{1+a_i},\ \ a_i \coloneqq \tfrac{\delta}{2}\ell_i,\ \ \eta_{n,i} \sim \mathcal{N}(0,1).
\label{eq:app-ar1}
\end{equation}
Indeed, the proximal of the quadratic $f$ at $y$ satisfies the first-order condition $H p - h + \tfrac{2}{\delta}M(p - y) = 0$, which gives
\[
p \;=\; \bigl(\tfrac{\delta}{2}H + M\bigr)^{-1}\bigl(\tfrac{\delta}{2}(H\mu + (h - H\mu)) + My\bigr) \;=\; \bigl(\tfrac{\delta}{2}H + M\bigr)^{-1}\bigl(\tfrac{\delta}{2}H\mu + My\bigr),
\]
using $h = H\mu$. Subtracting $\mu = (\tfrac{\delta}{2}H + M)^{-1}(\tfrac{\delta}{2}H + M)\mu$ gives $p - \mu = (\tfrac{\delta}{2}H + M)^{-1}M(y - \mu)$. Applying $X_{n+1} - \mu = -(X_n - \mu) + 2(p - \mu)$ with $y = X_n + \tfrac{\sqrt{2\delta}}{2}M^{-1/2}\xi_n$ yields
\[
X_{n+1} - \mu \;=\; \bigl(-I + 2(\tfrac{\delta}{2}H + M)^{-1}M\bigr)\,(X_n - \mu) \;+\; \sqrt{2\delta}\,(\tfrac{\delta}{2}H + M)^{-1}M^{1/2}\xi_n.
\]
Whitening gives $Z_{n+1} = R\, Z_n + \sqrt{2\delta}\,S\,\eta_n$ where $\eta_n \coloneqq Q^\top \xi_n \sim \mathcal{N}(0, I)$, $R = (I + \tfrac{\delta}{2}\Lambda)^{-1}(I - \tfrac{\delta}{2}\Lambda)$, and $S = (I + \tfrac{\delta}{2}\Lambda)^{-1}$. Since $R, S$ are diagonal with entries $r_i, 1/(1+a_i)$, the recurrence \eqref{eq:app-ar1} follows coordinate-wise.

\paragraph{Step 2 (Zero bias via variance budget).}
The stationary variance of the Cayley AR(1) \eqref{eq:app-ar1} is
\[
\mathrm{Var}(Z_{\infty,i}) \;=\; \frac{2\delta/(1+a_i)^2}{1 - r_i^2}.
\]
Since $1 - r_i^2 = 1 - \bigl(\tfrac{1-a_i}{1+a_i}\bigr)^2 = \tfrac{(1+a_i)^2 - (1-a_i)^2}{(1+a_i)^2} = \tfrac{4a_i}{(1+a_i)^2}$, the Cayley denominators cancel:
\[
\mathrm{Var}(Z_{\infty,i}) \;=\; \frac{2\delta/(1+a_i)^2}{4a_i/(1+a_i)^2} \;=\; \frac{2\delta}{4\cdot(\delta\ell_i/2)} \;=\; \frac{1}{\ell_i} \quad \text{for every } \delta > 0.
\]
This is the variance-budget identity \eqref{eq:budget-midpoint}. Combined with $\mathbb{E}[Z_{\infty,i}] = 0$ (from $|r_i| < 1$) and the independence of components, $Z_\infty \sim \pi_Z$, hence $X_\infty \sim \pi$. The synchronous contraction below gives uniqueness, as in Appendix~\ref{app:thm2}. This establishes Gaussian exactness.

\paragraph{Step 3 (Wasserstein contraction via synchronous coupling).}
For two chains driven by the same noise $\{\xi_n\}$, whitening preserves the shared-noise structure and \eqref{eq:app-ar1} gives
\[
Z_{n+1,i} - Z'_{n+1,i} \;=\; r_i\,(Z_{n,i} - Z'_{n,i}) \qquad \textrm{(deterministically, for each } i \textrm{)}.
\]
Iterating, $\|Z_N - Z'_N\| \leq \bigl(\max_i |r_i|\bigr)^N\,\|Z_0 - Z'_0\|$, i.e.,
\[
\|X_N - X'_N\|_M \;\leq\; \rho(\delta)^N\,\|X_0 - X'_0\|_M, \qquad \rho(\delta) = \max_{i} |r_i| = \max\bigl(|r(m_M)|, |r(L_M)|\bigr),
\]
since $r'(\ell)=-\delta/(1+\delta\ell/2)^2<0$, so $|r|$ attains its maximum on $[m_M,L_M]$ at an endpoint. Taking an optimal $W_2^M$-coupling at time $0$ and using synchronous noise proves the Wasserstein contraction bound.

\paragraph{Step 4 (Minimax optimum and mixing complexity).}
For $\delta\in[2/L_M,2/m_M]$, the endpoint magnitudes are $r(m_M)=(1-\delta m_M/2)/(1+\delta m_M/2)$ and $|r(L_M)|=(\delta L_M/2-1)/(\delta L_M/2+1)$, respectively decreasing and increasing in $\delta$. Outside this interval, moving $\delta$ toward it decreases both magnitudes. Equating $r(m_M)=|r(L_M)|$ yields
\[
\tfrac{1 - \delta m_M/2}{1 + \delta m_M/2} \;=\; \tfrac{\delta L_M/2 - 1}{\delta L_M/2 + 1} \iff \delta^\star \;=\; \tfrac{2}{\sqrt{m_M L_M}},
\]
and substituting $\delta^\star$ gives the common value
\[
\rho^\star \;=\; \tfrac{1 - \sqrt{m_M/L_M}}{1 + \sqrt{m_M/L_M}} \;=\; \tfrac{\sqrt{\kappa_{\mathrm{eff}}} - 1}{\sqrt{\kappa_{\mathrm{eff}}} + 1},
\]
establishing the optimal step and factor. For mixing: solving $\rho^{\star N} \leq \varepsilon$ gives $N \geq \log(1/\varepsilon) / \log(1/\rho^\star) = O(\sqrt{\kappa_{\mathrm{eff}}}\log(1/\varepsilon))$ using $\log(1/\rho^\star) = 2/\sqrt{\kappa_{\mathrm{eff}}} + O(\kappa_{\mathrm{eff}}^{-3/2})$. This proves the stated mixing complexity. \hfill$\square$

\input{sections/appendix_theory_rebuttal}

\section{JP-IMLA: kernel guarantees and empirical comparisons}
\label{app:jp-equivalence}

\input{sections/appendix_jp_rebuttal}

\subsection{Finite-budget comparison at early-$N$}
\label{app:jp-empirical-earlyN}

We run JP-IMLA on the geometric phantom under the same protocol as the body's TRUE-Picard headline benchmark---same $\sigma_{\mathrm{blur}}$, $\sigma_{\mathrm{noise}}$, $\rho_{\mathrm{TV}}$, $\delta = 0.141$, $\lambda = 5\!\times\!10^{-3}$, $K = 8$ chains from $X_0 = 0$, and \emph{matched seed base}---so that JP and TRUE trajectories are directly comparable per chain. These are reconstruction and mixing diagnostics; Proposition~\ref{prop:jp-equivalence} bounds reflected proximal distance, not relative SSE.

\begin{figure}[h]
\centering
\includegraphics[width=0.95\linewidth]{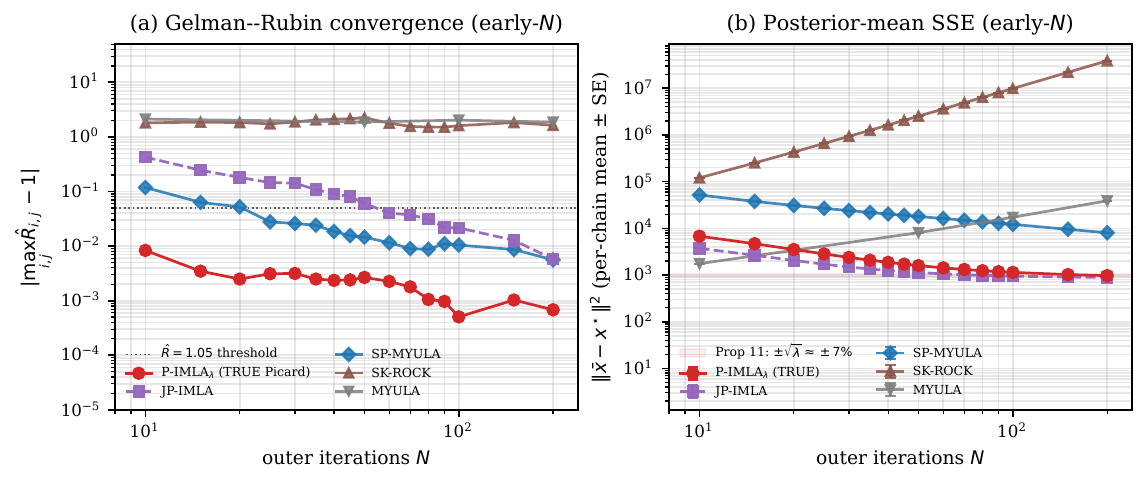}
\caption{\textbf{Finite-budget JP/Picard comparison at early-$N$.} (a) Classical per-pixel $\hat R$ from the common zero start. (b) Per-chain reconstruction SSE. At $N=200$, JP gives $888.28\pm2.63$ versus Picard's $972.09\pm2.62$, an $8.6\%$ difference. The shaded percentage band is a visual guide from the original plot, not a consequence of a proximal or Wasserstein bound; it does not certify relative SSE.}
\label{fig:app-b-jp-earlyN}
\end{figure}

\paragraph{Wall-clock per outer iteration.}
At $256\times256$ on an NVIDIA RTX PRO 6000 Blackwell, the recorded JP implementation averages $35.0$ ms per iteration against Picard's $2.11$ s, approximately $60\times$. The headline Picard budget is $(K_{\mathrm P},K_{\mathrm{in}})=(30,50)$, while JP uses $K=12$ ADMM steps. These are implementation timings under the recorded parallel workloads, not a complexity ratio at matched certified accuracy. Table~\ref{tab:jpcr-budget} reports separate RTX 5000 Ada budget-sweep timings and certificates at audited warm inputs; those pointwise certificates do not certify every step of the plotted production runs.

\subsection{Imaging-scale comparison}
\label{app:jp-imaging}

We extend the matched-budget benchmark of §\ref{sec:experiments} to JP-IMLA on all three phantoms used in Table~\ref{tab:cross-phantom} (geometric, cameraman, Shepp--Logan), at $N=2{,}000$ on RTX PRO 6000 Blackwell, with seed offsets matching the canonical TRUE-Picard runs.

\begin{figure}[h]
\centering
\includegraphics[width=\linewidth]{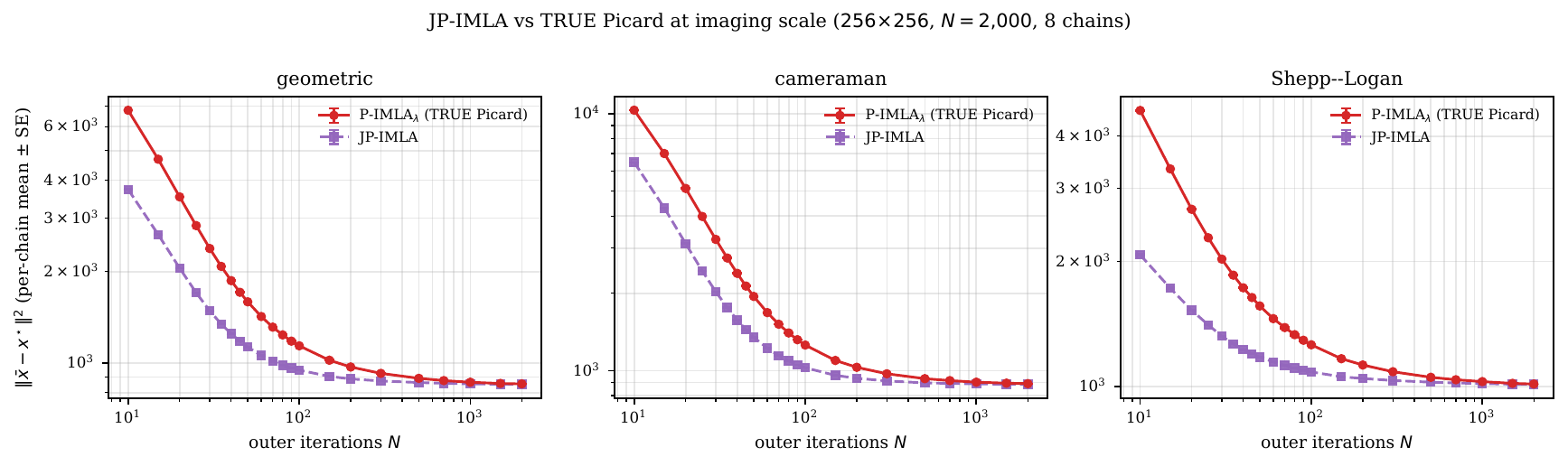}
\caption{\textbf{JP-IMLA tracks TRUE Picard at imaging scale across three phantom morphologies.} Per-chain SSE means $\pm$ SE; both samplers run the same $N = 2{,}000$-iteration matched-budget protocol. The TRUE/JP traces are visually indistinguishable for $N \geq 200$ on all three phantoms. JP uses the original $K=12$ budget; Table~\ref{tab:jpcr-budget} separately reports budget sensitivity and pointwise certificates.}
\label{fig:app-b-jp-imaging}
\end{figure}

\begin{table}[h]
\centering
\small
\caption{\textbf{Cross-phantom JP-IMLA vs TRUE Picard at $N = 2{,}000$.} Per-chain SSE means $\pm$ SE (JP: eight chains; Picard: eight geometric and four per other phantom). The kernels and finite inner solvers differ. Relative SSE is a reconstruction statistic, not a distributional error bound.}
\label{tab:jp-vs-true}
\setlength{\tabcolsep}{6pt}
\begin{tabular}{@{}lccc@{}}
\toprule
Phantom & TRUE Picard SSE & JP-IMLA SSE & Drift (JP/TRUE $- 1$) \\
\midrule
geometric    & $855.22 \pm 0.66$  & $852.72 \pm 0.65$  & $-0.29\%$ \\
cameraman    & $890.24 \pm 1.19$  & $884.52 \pm 0.62$  & $-0.64\%$ \\
Shepp--Logan & $1{,}013.99 \pm 0.81$ & $1{,}012.27 \pm 0.63$ & $-0.17\%$ \\
\bottomrule
\end{tabular}
\end{table}

At these finite budgets, the reported JP and Picard SSEs differ by at most $0.7\%$. This agreement does not imply equality of invariant laws or uniform solver accuracy; the model and finite-ADMM terms are quantified separately in Appendix~\ref{app:jp-theory}.

\section{Auxiliary experimental results}
\label{app:experiments}

This appendix collects (a) validation experiments for Theorems~\ref{thm:gaussian-exact}--\ref{thm:bea-bias} that the main-text §\ref{sec:experiments} summarises in one sentence, and (b) intermediate-checkpoint reconstructions for the matched-budget benchmark of Figure~\ref{fig:imaging-panel}.

\paragraph{Geometric- and Shepp--Logan-phantom imaging panels.}

\begin{figure}[h]
\centering
\includegraphics[width=0.85\linewidth]{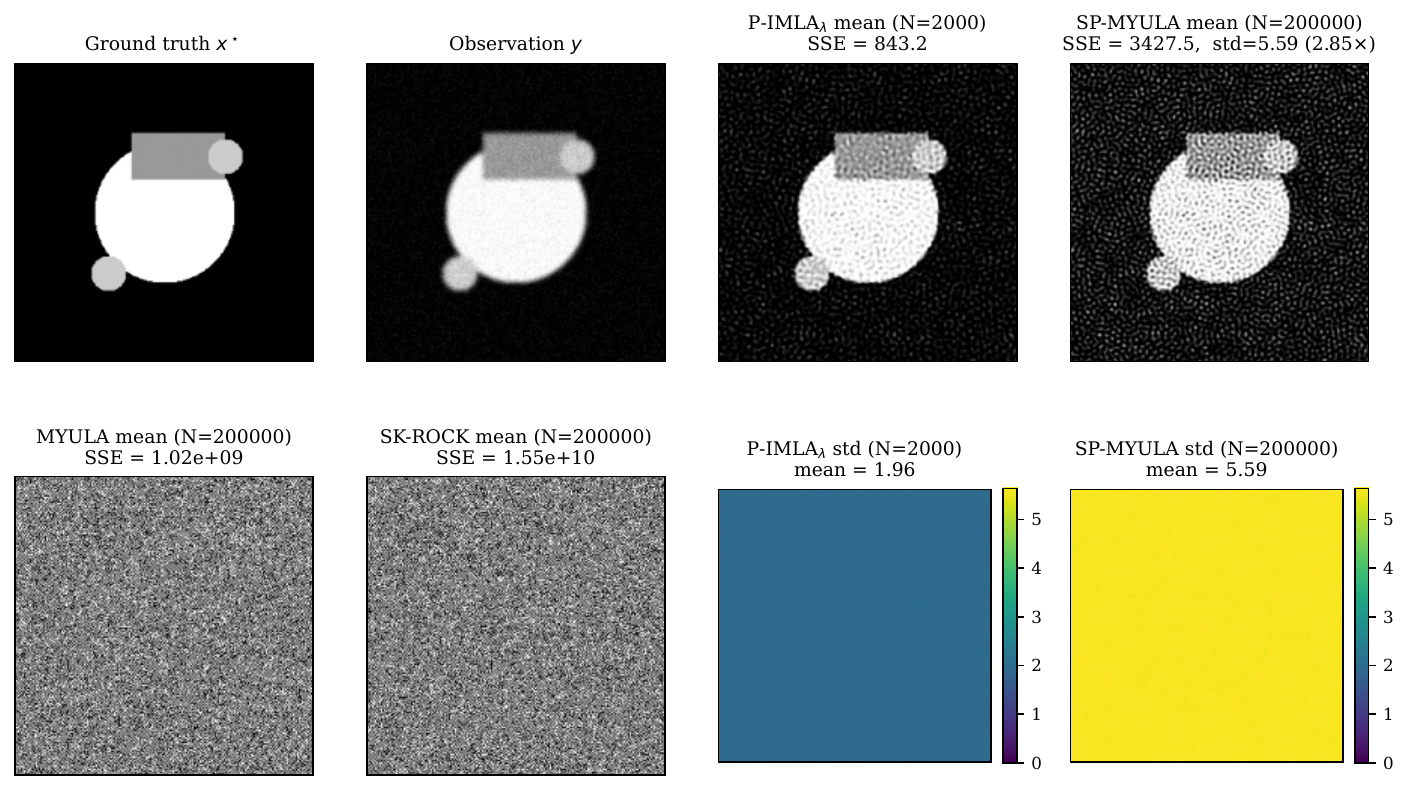}
\caption{\textbf{Geometric phantom counterpart to main-text Figure~\ref{fig:imaging-panel}}, matched-budget reconstructions: ours at $N = 2{,}000$ (top right; ensemble SSE $843.1$, std mean $1.95$); SP-MYULA / MYULA / SK-ROCK at $N = 2\!\times\!10^{5}$. Same structural conclusions as the cameraman panel apply: ours recovers the four phantom regions cleanly, SP-MYULA's mean is similar but std uniformly inflated $\sim$$2.86\!\times$, MYULA and SK-ROCK mean fields are unrecognisable.}
\label{fig:imaging-panel-geometric}
\end{figure}

\paragraph{P-IMLA$_\lambda$ recovery at intermediate $N$ (geometric).}
Figure~\ref{fig:imaging-panel-geometric} shows the headline pair $N_{\mathrm{ours}} = 2{,}000$ vs $N_{\mathrm{base}} = 2\!\times\!10^{5}$. Identically rendered panels at $N_{\mathrm{ours}} \in \{10, 50, 100, 200, 500, 1000\}$ (each paired with the corresponding $100\!\times$ baseline budget) are provided in the supplementary material. Visual quality stabilises by $N \approx 200$ (where $\hat{R} = 1.0004$); $N = 10$ and $N = 50$ show partial mixing artefacts. The contraction timescale $N\approx22$ does not predict a threshold for $\hat R$ or visual quality.

Uncertainty conventions are experiment-specific. For averages over $K$ independent seeds, $\mathrm{SE}$ denotes the sample standard deviation divided by $\sqrt{K}$; single-chain batch-means errors and deterministic calculations are identified separately below.

\paragraph{Sampler naming conventions.} Throughout this appendix we use \emph{SP-IMLA} for the smooth-target P-IMLA of \eqref{eq:pimla-smooth} with $M = H$; \emph{SP-ULA} for the corresponding explicit-Euler scheme $X_{n+1} = X_n - \delta M^{-1}\nabla f(X_n) + \sqrt{2\delta}\, M^{-1/2}\xi_n$ at $M = H$; and \emph{SP-MYULA} for the explicit-Euler counterpart of P-IMLA$_\lambda$ on the same $M$-norm-Moreau-smoothed potential $U_\lambda = f + g^{\lambda, M}$:
\begin{equation}
X_{n+1} \;=\; X_n - \delta M^{-1}\nabla U_\lambda(X_n) + \sqrt{2\delta}\,M^{-1/2}\xi_n,\qquad M = H,
\label{eq:sp-myula}
\end{equation}
identical to MYULA \citep{durmus2018efficient} except (i) the Moreau envelope is taken in the $M$-norm matching the preconditioner geometry (matching the body's convention), and (ii) the drift and noise are preconditioned by $M^{-1}$ and $M^{-1/2}$ respectively. SP-MYULA isolates the effect of midpoint vs.\ Euler integration with preconditioner choice, Moreau parameterisation, and noise injection all matched to P-IMLA$_\lambda$, the observed SSE gap also depends on finite-solver accuracy and initial transients.

\subsection{Variance-budget identity (Theorem~\ref{thm:gaussian-exact} and Proposition~\ref{prop:p-ula-failure})}
\label{app:exp-variance-budget}

We test the variance-budget identity~\eqref{eq:budget-midpoint} and the Euler overfill~\eqref{eq:budget-euler} on two settings: \textbf{(a)}~a 1-D Gaussian $f(x) = x^2/2$ at $M = 1$, swept across eight step sizes $\delta \in [0.1, 1.8]$; and \textbf{(b)}~a $d = 100$ anisotropic Gaussian $H = \operatorname{diag}(1, \ldots, 100)$, $M = H$, $\kappa_{\mathrm{eff}} = 1$. Each sampler runs $N = 5\!\times\!10^6$ post-burn-in samples. Setup~A uses batch-means SEs (batch size $10^4$). Setup~B computes per-mode variance ratios $\mathrm{VR}_j$ from one vector chain in whitened coordinates; the SE of their average is the sample standard deviation across the $100$ independent Gaussian modes divided by $\sqrt{100}$.

\begin{table}[h]
\centering
\small
\caption{\textbf{Variance-budget verification (1-D Gaussian, Setup~A).} SP-IMLA's $\mathrm{VR}$ lies in $[0.9978,\, 1.0009]$ across all eight $\delta$ ($|z|$-scores $\leq 1.34$); Setup~B per-mode average is $0.999937 \pm 0.000087$. SP-ULA tracks $1/(1-\delta/2)$ to within $0.12\%$.}
\label{tab:variance-budget}
\setlength{\tabcolsep}{9pt}
\renewcommand{\arraystretch}{1.05}
\begin{tabular}{@{}cccc@{}}
\toprule
$\delta$ & SP-IMLA $\mathrm{VR}$ & SP-ULA $\mathrm{VR}$ & Theory $1/(1-\delta/2)$ \\
\midrule
$0.1$ & $0.99885$ & $1.0539$ & $1.0526$ \\
$0.5$ & $1.00033$ & $1.3324$ & $1.3333$ \\
$1.0$ & $0.99949$ & $2.0015$ & $2.0000$ \\
$1.5$ & $0.99953$ & $3.9986$ & $4.0000$ \\
$1.8$ & $0.99941$ & $9.9976$ & $10.000$ \\
\bottomrule
\end{tabular}
\end{table}

SP-IMLA's $\mathrm{VR}$ is statistically indistinguishable from $1$ across an $18$-fold step-size range, confirming the Cayley algebra cancellation \eqref{eq:budget-midpoint} to four decimals even at $\delta = 1.8$, where SP-ULA's variance has inflated ten-fold. Theorem~\ref{thm:gaussian-exact} makes no step-size assumption; explicit Euler breaks down precisely in the large-$\delta$ regime that preconditioning calls for.

\subsection{Quantitative $2\!\times\!2$ landscape on a Gaussian target}
\label{app:exp-2x2}

Figure~\ref{fig:2x2} reports a fixed-budget statistic on the Gaussian setup of \ref{app:exp-contraction}. Explicit Euler is either slow at stable steps or unstable beyond its spectral boundary. Both midpoint schemes have the exact Gaussian invariant law with exact proximal solves. For unpreconditioned IMLA, low-curvature modes retain their initial-mean error over the reported horizon: $98.7\%$ of the displayed VR $=48.6$ is transient, while the post-burn-in stationary ratio is $1.000$. The quantity is a time-averaged normalised mean-square error, not stationary variance inflation. Preconditioning reduces the relevant condition number and shortens that transient. Appendix~\ref{app:r5-vr} gives the exact AR(1) decomposition and its stationary-start check.

\subsection{Contraction rate and \texorpdfstring{$\kappa_f$}{kappa-f}-independence (Theorem~\ref{thm:contraction})}
\label{app:exp-contraction}

\begin{figure}[h]
\centering
\includegraphics[width=0.95\linewidth]{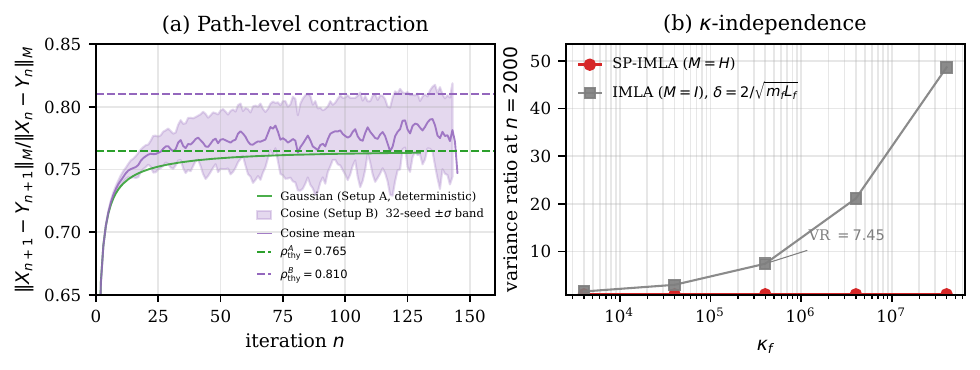}
\caption{\textbf{(a)}~Path-level contraction ratio $\|X_{n+1} - Y_{n+1}\|_M / \|X_n - Y_n\|_M$ for two chains driven by shared noise. Setup~A (Gaussian, deterministic) attains mean ratio $0.7544$, $98.65\%$ of the bound $\rho_{\mathrm{thy}} = 0.7647$. Setup~B (32 independent seeds, $d = 50$ cosine-perturbed target $f + 0.3\sum_j(\cos x_j - 1)$) gives seed-mean max ratio $0.8074 \pm 0.0015$ vs the strict bound $\rho_B = 0.8100$ ($1.73$~SE below). Shading shows one standard deviation across seeds; quoted uncertainties are SEs. \textbf{(b)}~Cold-start time-averaged normalized MSE over $n=2000$ iterations as $\kappa_f$ varies over four orders of magnitude (Exp~04, 32 independent seeds, $M = H$). SP-IMLA's $\mathrm{VR}^{\mathrm{mode}} = 1.0246 \pm 0.0001$ across $\kappa_f \in [4\!\times\!10^{3},\, 4\!\times\!10^{7}]$; IMLA grows monotonically from $\mathrm{VR}=7.449$ to $48.6$. These values include the initial-mean transient; both exact midpoint kernels preserve Gaussian stationary variance (Appendix~\ref{app:r5-vr}).}
\label{fig:contraction-kappa}
\end{figure}

The path-level panel verifies the deterministic contraction inequality of Theorem~\ref{thm:contraction}: at every iteration the $M$-norm separation contracts to within float precision, with Gaussians approaching the contraction bound more tightly than the cosine target ($98.65\%$ vs $93.13\%$)---quantitative evidence that Gaussian targets attain the bound (Theorem~\ref{thm:contraction}). The $\kappa$-independence panel verifies the resulting $\sqrt{\kappa_{\mathrm{eff}}}$ rate at the mixing-rate optimum (Remark~\ref{rem:precision-protocol}): the $\sqrt{\kappa_f}$ dependence the IMLA curve exhibits is replaced under preconditioning by a constant.

\subsection{Joint smoothing and step-size sweep at imaging scale}
\label{app:lambda-sweep}

A $\lambda$-sweep at the same $256\!\times\!256$ setting over a $200$-fold range $\lambda \in [10^{-3},\, 0.2]$ (Exp~10) exhibits two robust empirical features: (i)~$\kappa_{\mathrm{eff}}^\lambda = 1 + 1/\lambda$ matches the measured spectrum at every $\lambda$ to four significant figures, while MYULA's $\kappa_{\mathrm{eff}}$ remains four orders of magnitude larger; (ii)~$\mathrm{ESS}/\text{iter}$ saturates at $1.00$ for $\lambda \geq 5\!\times\!10^{-3}$. The sweep traces a joint $(\delta_\lambda^\star, \rho_a^\star)$ schedule rather than independently varying $\lambda$ in the inner solver; Term~III's $\sqrt{\lambda}$ scaling is verified directly in 1-D in §\ref{app:moreau-bias} (Remark~\ref{rem:duality}).

\subsection{Step-size robustness and the antithetic regime}
\label{app:exp-antithetic}

\begin{figure}[h]
\centering
\includegraphics[width=0.7\linewidth]{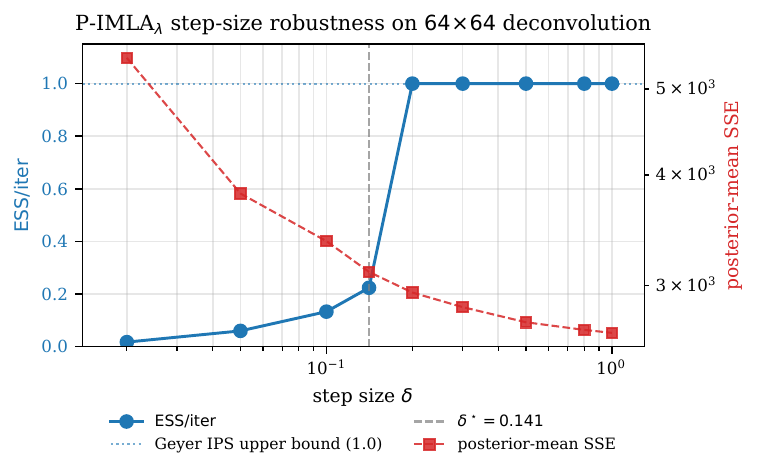}
\caption{\textbf{Step-size robustness of P-IMLA$_\lambda$ on $64\!\times\!64$ deconvolution.} $\mathrm{ESS}/\mathrm{iter}$ rises through the worst-case $\delta^\star = 0.141$ and \emph{saturates at $1.00$} (implementation cap) across $\delta \in [0.2,\, 1.0]$. Posterior-mean SSE remains bounded across the same range. The negative-Cayley antithetic regime at $\delta > \delta^\star$ produces alternating-sign autocorrelations that the reflected-proximal contraction bound of Theorem~\ref{thm:contraction} (which uses $|r(L_M)|$) does not capture; a signed analysis closing this gap is an open problem.}
\label{fig:app-c-antithetic}
\end{figure}

Sweeping $\delta$ on a $64\!\times\!64$ non-smooth benchmark (same $M = \hat{H}$, $\lambda = 5\!\times\!10^{-3}$, $\delta^\star = 2/\sqrt{m_M L_\lambda} \approx 0.141$), we observe a regime not predicted by Theorem~\ref{thm:contraction}'s worst-case bound. The Cayley factor $r(\ell)$ is negative for eigenvalues $\ell>2/\delta$, allowing alternating-sign correlations. At $\delta>\delta^\star$, the reported effective-sample-size per iteration (Geyer initial-positive-sequence estimator, capped at one in our implementation) plateaus at $\mathrm{ESS}/\text{iter} = 1.00$ for $\delta \in [0.2,\, 1.0]$ (Figure~\ref{fig:app-c-antithetic}), an order-of-magnitude robustness range beyond $\delta^\star$. The Wasserstein contraction of Theorem~\ref{thm:contraction} does not forbid this---$|r(L_\lambda)| \to 1$ as $\delta \to \infty$---but at finite $\delta$ alternating-sign autocorrelations can reduce variance without improving the worst-case path-level bound. A signed analysis of this regime is an open problem.

\subsection{Beyond Fourier-diagonal forward operators (random fan-beam)}
\label{app:exp-radial}

To verify that the framework is not specific to circulant forward operators, we re-run the matched-budget protocol of §\ref{sec:experiments} with a non-Fourier $A$. \textbf{Setup}: $64\!\times\!64$ image ($d = 4{,}096$), $m = 4{,}096$ random fan-beam projection lines through the image domain (sparse $A \in \mathbb{R}^{m\times d}$ with $\sim\!2.51\!\times\!10^{5}$ nonzero entries, $\sim$$64$ pixels per ray); Gaussian noise $\sigma_{\mathrm{noise}} = 0.05$; TV regulariser $\rho_{\mathrm{TV}} = 0.3$, $\lambda = 5\!\times\!10^{-3}$, $\delta = 0.141$ as in the headline benchmark. The preconditioner is a randomised Nystr\"om \citep{frangella2023randomized} approximation $M = U\Lambda U^\top + \beta I$ of $A^\top A / \sigma_{\mathrm{noise}}^{2}$ at rank $r = 128$; the inner Picard $u$-update~\eqref{eq:picard-fixed-point} is no longer Fourier-diagonal, so we replace it by $K_{\mathrm{cg}} = 12$ steps of conjugate gradient on the SPD operator $L = (1/\lambda) M + \rho_a D^\top D$ with relative tolerance $10^{-6}$. This auxiliary run uses $(K_{\mathrm P},K_{\mathrm{in}})=(8,15)$ and ADMM warm starts, a separate inner-solver budget from the headline $(30,50)$ protocol.

\textbf{Matched-budget result.} P-IMLA$_\lambda$ runs $K = 8$ chains for $N = 300$ outer iterations; SP-MYULA runs $K = 4$ chains for $N = 30{,}000$ ($100\!\times$ iteration ratio). P-IMLA$_\lambda$ first crosses $\hat{R} < 1.05$ at the second checkpoint $N = 5$ ($\hat{R} = 1.0485$) and reaches $\hat{R} = 0.9990$ at $N = 300$, with per-chain SSE $36.99 \pm 0.23$; SP-MYULA crosses $\hat{R} < 1.05$ at $N = 10$ ($\hat{R} = 1.0190$) and reaches $\hat{R} = 1.0000$ at $N = 30{,}000$ with per-chain SSE $38.94 \pm 0.14$. Wall-clocks are $10.1$--$20.8$\,min/chain for ours and $137.4$\,min/chain for SP-MYULA, a $6.5\!-\!13.7\!\times$ difference under the stated iteration allocation. The SP-MYULA/P-IMLA SSE ratio is $1.05$ ($38.94/36.99$). These finite-budget results demonstrate operation beyond Fourier-diagonal models; they do not measure stationary bias or establish a $100\!\times$ mixing-rate improvement. The TV envelope does not satisfy Theorem~\ref{thm:bea-bias}'s smoothness assumptions. At rank $r = 128 \ll d = 4{,}096$, the Nystr\"om $M$ leaves the orthogonal-complement modes essentially un-preconditioned. Image-scale CT/MRI validation with rank matched to the spectrum remains open (Appendix~\ref{app:limitations}).

\subsection{Formal drift-only coefficient: quartic stress tests}
\label{app:exp-bea-1d}

We illustrate the formal drift-only coefficient on the 1-D target $\pi \propto e^{-f(x)}$ with $f(x) = x^2/2 + \varepsilon\, x^4/24$, choosing the observable $\varphi(x) = x^2$. Then $h(x) = f''(x) = 1 + (\varepsilon/2)\,x^2$ and $\mathrm{Cov}_\pi(h, \varphi) = (\varepsilon/2)\,\mathrm{Var}_\pi(x^2)$, so the predicted leading bias coefficient is $c_1 = (\varepsilon/8)\,\mathrm{Var}_\pi(x^2)$ with $\mathrm{Var}_\pi(x^2)$ computed by quadrature. Each plotted point uses one P-IMLA chain with $N = 5\!\times\!10^6$ post-burn-in iterations; the figure displays point estimates without error bars. The implementation computes batch-means SEs with batch size $5{,}000$; its separate Richardson coefficient check reports SEs across $32$ independent seeds, without bootstrap resampling.

\begin{figure}[h]
\centering
\includegraphics[width=0.95\linewidth]{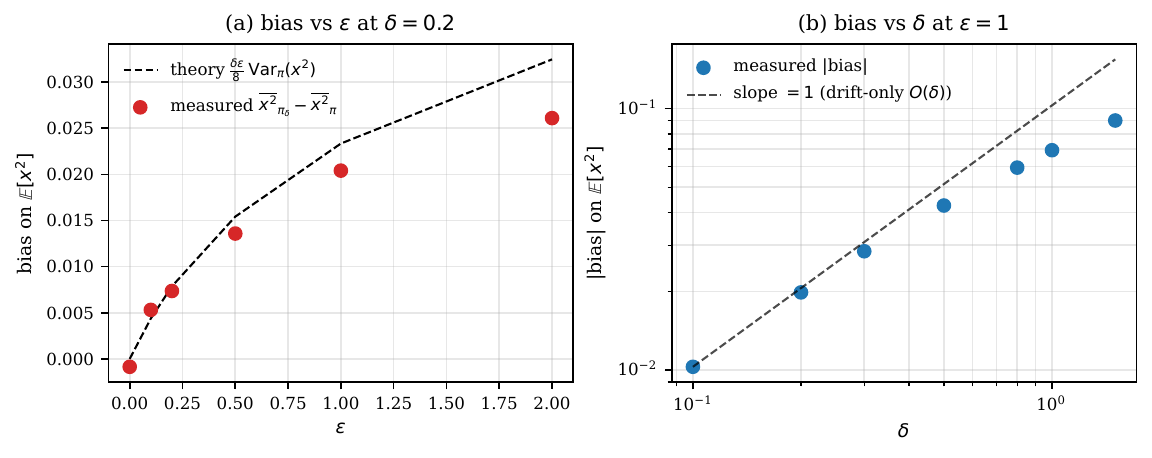}
\caption{\textbf{Formal drift-only coefficient on a 1-D quartic test target.} \textbf{(a)}~At fixed $\delta$, measured bias estimates on $\mathbb{E}[x^2]$ are compared with the leading-order prediction $\delta\varepsilon/8\cdot\mathrm{Var}_\pi(x^2)$; no uncertainty bars are displayed. \textbf{(b)}~At fixed $\varepsilon$, the bias scales as $\delta^1$ on log--log axes (reference dashed line), consistent with a leading $O(\delta)$ weak bias. Drift-only refers to the differential order of the defect; P-ULA also has a leading $O(\delta)$ weak bias (Appendix~\ref{app:pula-bea}).}
\label{fig:app-c-drift-only}
\end{figure}

This quartic target is not globally $L$-smooth, so the experiment illustrates the coefficient outside the stated theorem assumptions; the globally smooth $d=256$ test is in App.~\ref{app:rebuttal-experiments}. The empirical $\delta$-slope of $1.00$ (panel b) and the linear $\varepsilon$-fit (panel a) jointly confirm the leading-order coefficient predicted by the BEA expansion of \S\ref{app:thm3}.

\paragraph{2-D formal-coefficient stress test.}
The 1-D test illustrates the coefficient when $h(x) = f''(x)$ is non-constant. To examine the formula beyond one dimension, we repeat the test on $f(x_1, x_2) = (x_1^2 + x_2^2)/2 + \varepsilon\, x_1^2 x_2^2 / 4$, where $h(x) = \operatorname{tr}(\nabla^2 f) = 2 + \varepsilon(x_1^2 + x_2^2)/2$ and $\operatorname{Cov}_\pi(h, \varphi)$ is computed by 2-D quadrature for $\varphi(x) = x_1^2 + x_2^2$. P-IMLA samples are run for $N = 5\!\times\!10^{6}$ post-burn-in iterations.

\begin{figure}[h]
\centering
\includegraphics[width=0.95\linewidth]{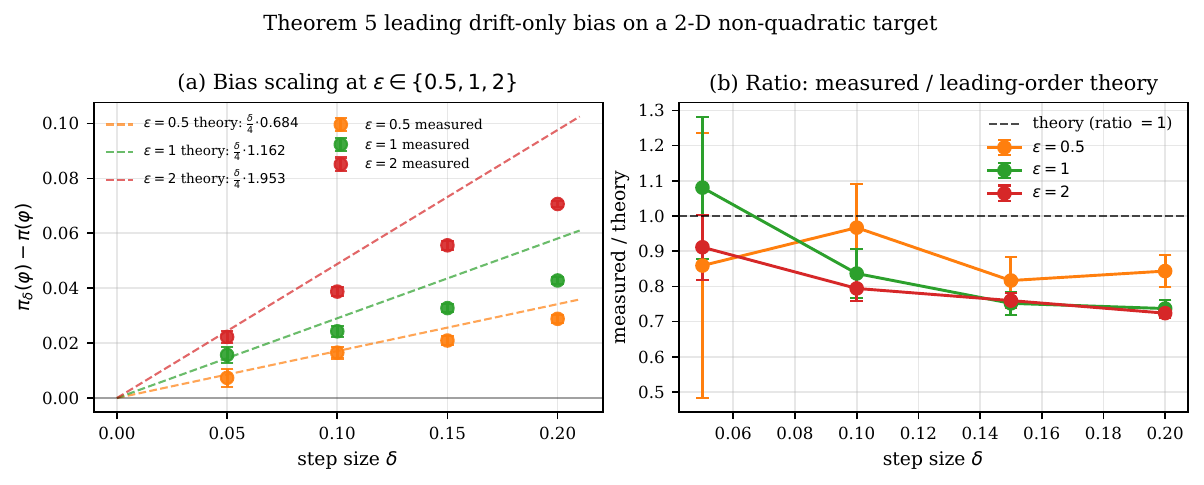}
\caption{\textbf{2-D formal-coefficient check.} \textbf{(a)}~Measured bias on $\mathbb{E}[\varphi]$ scales linearly with $\delta$ at each $\varepsilon$, with theory line $\delta\,\operatorname{Cov}_\pi(h,\varphi)/4$ from Theorem~\ref{thm:bea-bias} (dashed). \textbf{(b)}~Measured/theory ratio across all $(\varepsilon, \delta)$ tested, clustering near $1$; the $\sim\!0.72\!-\!0.84$ values at the largest $\delta$ are consistent with a negative $O(\delta^2)$ next-order BEA correction.}
\label{fig:app-c-bea-2d}
\end{figure}

The control case $\varepsilon = 0$ yields measured bias indistinguishable from zero (consistent with $h \equiv 2$ giving $\operatorname{Cov}_\pi(h, \varphi) = 0$), and across $\varepsilon \in \{0.5, 1, 2\}$ the measured bias matches the leading-order prediction within ratio $0.72\!-\!1.08$ across all $\delta \in [0.05, 0.20]$. The polynomial interaction is not globally smooth or globally convex, so this is a formal-coefficient check rather than a verification under Theorem~\ref{thm:bea-bias}\textquotesingle s assumptions.

\subsection{SK-ROCK $\delta$-stability sweep}
\label{app:exp-skrock-delta}

We run SK-ROCK \citep[$s = 10$, $\eta = 0.05$,][]{pereyra2020accelerating} on the same $64\!\times\!64$ Gaussian-deconvolution benchmark used in \S\ref{app:exp-antithetic} ($\sigma_{\mathrm{blur}} = 2.0$, $\sigma_{\mathrm{noise}} = 0.05$, $\rho_{\mathrm{TV}} = 0.02$, $\lambda = 5\!\times\!10^{-3}$, $M = I$), and sweep $\delta$ over an order-of-magnitude range straddling the theoretical Chebyshev stability boundary $\delta_{\max}^{\mathrm{thy}} = 2 s^2 (1 + \eta/s^2)/L_M = 0.9955$ (with $L_M = 1 + 1/\lambda = 201$). For each $\delta$ the chain is run for $5{,}500$ iterations ($500$ burn-in, $5{,}000$ posterior); a sample is flagged as diverged if $\|X\|_\infty > 10^6$ or non-finite at any iteration.

\begin{figure}[h]
\centering
\includegraphics[width=0.7\linewidth]{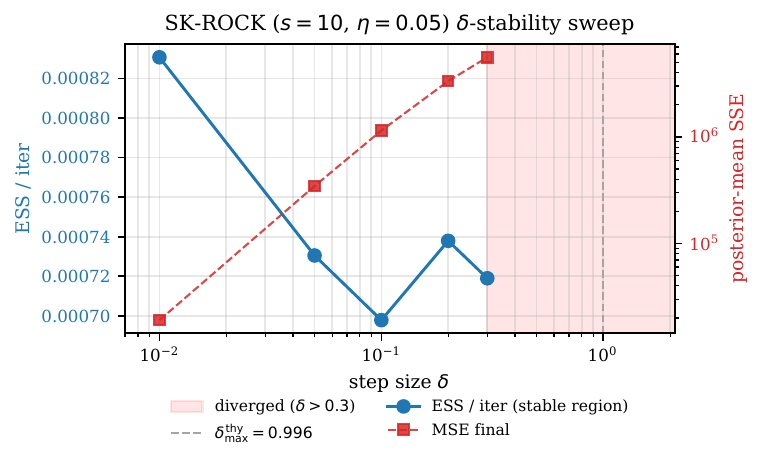}
\caption{\textbf{SK-ROCK $\delta$-stability sweep on a $64\!\times\!64$ Moreau-smoothed posterior.} Stable region (left of red shading): $\delta \leq 0.30$. Diverged region (red): $\delta \geq 0.50$. The empirical stability boundary $\delta_{\max}^{\mathrm{emp}} \approx 0.30$ is roughly $3\!\times$ tighter than the smooth-case Chebyshev prediction $\delta_{\max}^{\mathrm{thy}} = 0.9955$, indicating that SK-ROCK's published stability margin is \emph{optimistic} (over-predicts the stable region) once the Moreau-smoothed sub-gradient enters the iteration. Even within the stable region, $\mathrm{ESS}/\text{iter} \in [7\!\times\!10^{-4},\, 8\!\times\!10^{-4}]$ remains essentially flat, an order $10^3$ below P-IMLA$_\lambda$'s $\mathrm{ESS}/\text{iter} \approx 1$ at the same $\lambda$ (cf.~§\ref{app:exp-antithetic}); MSE rises monotonically with $\delta$ within the stable region as the chain has not finished a single integrated-autocorrelation time within the $5{,}000$-sample budget.}
\label{fig:app-c-skrock-delta}
\end{figure}

The takeaway is two-fold: (i)~SK-ROCK's stability margin contracts substantially under Moreau-smoothed proximal handling, restricting the practical $\delta$ range to roughly $[10^{-2},\, 0.3]$ at this $\lambda$; (ii)~within that range, $\mathrm{ESS}/\text{iter}$ is bounded by $\sim 10^{-3}$, so the iteration-budget gap to P-IMLA$_\lambda$ identified in the main-text matched-budget benchmark (Table~\ref{tab:imaging}) is intrinsic to the explicit-Euler $+$ Chebyshev recursion's mixing rate, not to a sub-optimal step-size choice.

\subsection{Term~III $\sqrt{\lambda}$ scaling on polyhedral $g$}
\label{app:moreau-bias}

The Talagrand-based derivation of Term~III is generic in that it uses only the pointwise bound $(g - g^{\lambda, M})(x) \leq \tfrac{\lambda}{2}(L_g^M)^2$ together with strong convexity of~$f$ in the $M$-norm; it makes no structural assumption on the sub-level geometry of~$g$ beyond Lipschitzness. For the polyhedral regularizers that motivate P-IMLA$_\lambda$ in practice (TV, $\ell_1$), we expect the bound to be loose because the $M$-norm subgradient bound collapses across a codimension-1 switching set. We verify this directly on the 1-D target $\pi \propto e^{-f - g}$, $f(x) = x^2/2$, in two contrasting setups: a \emph{symmetric} polyhedral $g_{\mathrm{sym}}(x) = |x|$ and an \emph{asymmetric} polyhedral $g_{\mathrm{asym}}(x) = \max(2x, -x)$. The target distributions $\pi$ and $\pi_\lambda$ are approximated by deterministic quadrature on a log-concentrated grid, and $W_2^M(\pi_\lambda, \pi)$ is evaluated via the 1-D quantile identity $W_2^2(P, Q) = \int_0^1 (F_P^{-1}(u) - F_Q^{-1}(u))^2\,du$ at $17$ log-spaced values of $\lambda \in [10^{-4}, 1]$. This calculation has numerical integration and interpolation error, but no Monte Carlo sampling error bars.

\begin{figure}[ht]
\centering
\includegraphics[width=0.72\linewidth]{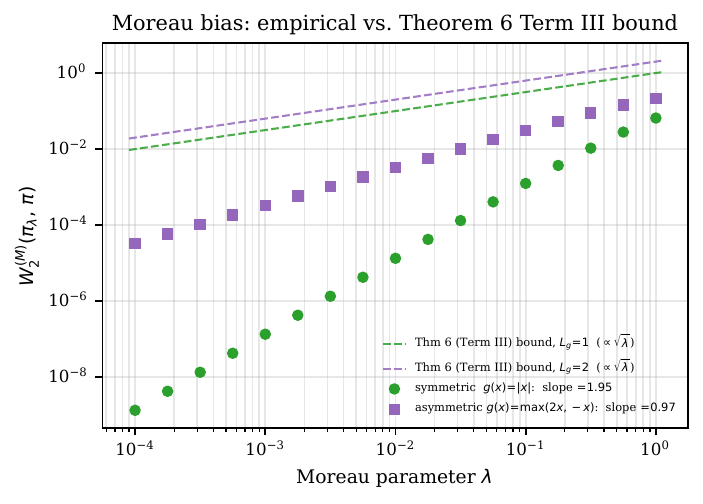}
\caption{\textbf{Empirical verification of Theorem~\ref{thm:end-to-end} Term~III.} The dashed lines show the theoretical $L_g^M\sqrt{\lambda/m_M}$ bounds ($L_g^M = 1$ for $g_{\mathrm{sym}}$, $L_g^M = 2$ for $g_{\mathrm{asym}}$). The empirical $W_2^M(\pi_\lambda, \pi)$ lies \emph{strictly below} the bound at every $\lambda$ in both setups (maximum $W_2/\text{bound}$ ratios $0.065$ and $0.106$). The log-log fitted slopes are $0.97$ for the asymmetric case and $1.95$ for the symmetric case, both tighter than the theorem's worst-case exponent of $0.5$. These two examples show that the generic bound can be conservative; they do not establish an improved rate for all polyhedral penalties.}
\label{fig:moreau-bias-verification}
\end{figure}

Both slopes exceed the generic exponent $1/2$. The difference between the two examples is consistent with cancellation in the symmetric case, but a uniform polyhedral improvement or a universal symmetry-based $O(\lambda^2)$ rate is not proved here.

\subsection{Baseline tuning sweep (fairness)}
\label{app:exp-baseline-tuning}

To pre-empt the question of whether the explicit-Euler baselines are fairly tuned, we sweep $\delta \in \{10^{-6}, 10^{-5}, 10^{-4}, 10^{-3}, 5\!\times\!10^{-3}, 10^{-2}, 5\!\times\!10^{-2}, 10^{-1}, 5\!\times\!10^{-1}\}$ for SP-MYULA and MYULA on the same $64\!\times\!64$ benchmark of \S\ref{app:exp-antithetic}. Per $\delta$ we run $5{,}500$ iterations ($500$ burn-in) and report ESS/iter at the diagnostic pixel (Geyer initial-positive sequence) and the post-burn posterior-mean SSE. The body benchmark of §\ref{sec:experiments} uses (rounded down from the largest stable for safety margin) $\delta_{\mathrm{SP\text{-}MYULA}} = 9.45\!\times\!10^{-3}$, $\delta_{\mathrm{MYULA}} = 3\!\times\!10^{-3}$, and $\delta_{\mathrm{SK\text{-}ROCK}} = 0.25$ (cf.\ §\ref{app:exp-skrock-delta}).

\begin{figure}[h]
\centering
\includegraphics[width=0.95\linewidth]{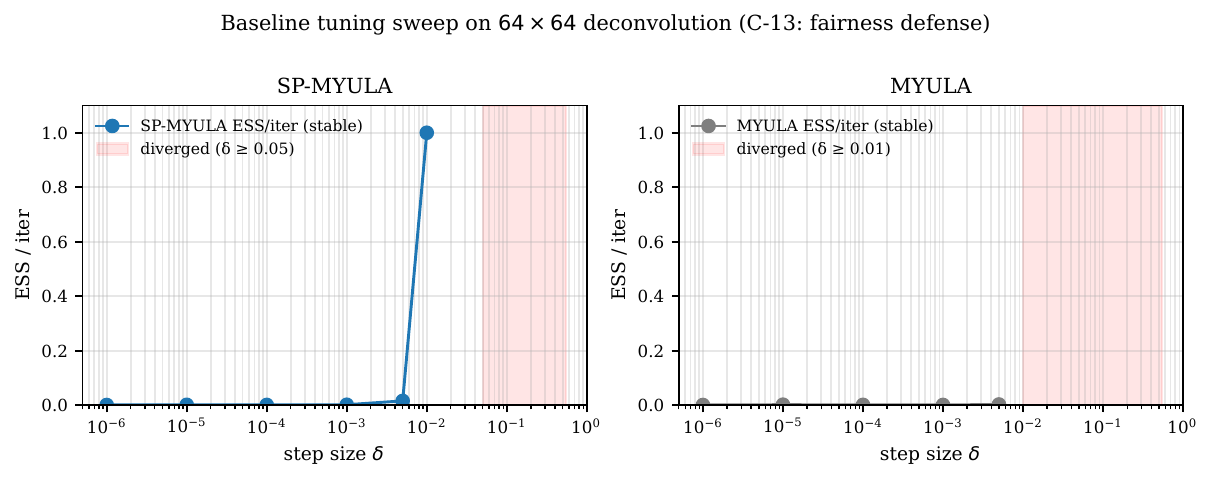}
\caption{\textbf{Baseline tuning sweep on $64\!\times\!64$ deconvolution.} \textbf{Left:} SP-MYULA stable through $\delta = 10^{-2}$ (ESS/iter peaks at $1.0$ at the largest stable $\delta$); diverges from $\delta \geq 5\!\times\!10^{-2}$. \textbf{Right:} MYULA stable only through $\delta = 5\!\times\!10^{-3}$, with best ESS/iter $= 2.6\!\times\!10^{-3}$ ($\sim\!380\!\times$ lower than SP-MYULA's best); diverges from $\delta \geq 10^{-2}$. The body benchmark of §\ref{sec:experiments} runs each baseline at its best stable step size from this sweep, controlling step-size choice within this sweep; it does not establish optimal baseline inner-solver tuning.}
\label{fig:app-c-baseline-tuning}
\end{figure}

The headline narrative implication: at imaging-scale curvature ($\kappa_f$ here is similar to the body $\kappa_f$ ratio), MYULA's stability boundary forces $\delta$ so small that even at its \emph{best} stable choice it requires $\sim\!400\!\times$ more iterations (best ESS/iter $\approx 2.6\!\times\!10^{-3}$ vs P-IMLA$_\lambda$'s ESS/iter $\approx 1$ in §\ref{app:exp-antithetic}) to extract a single effective sample.

\subsection{Scope of empirical comparison: the Langevin family}
\label{app:scope}

The matched-budget benchmarks of §\ref{sec:experiments} compare P-IMLA$_\lambda$ against three Langevin-family baselines---MYULA, SK-ROCK, SP-MYULA---under matched preconditioning, Moreau parameter, and noise injection. This isolates the effect we analyse and prove rate theorems for: the integration scheme (explicit Euler vs.\ implicit midpoint) on the same underlying Langevin SDE with non-smooth proximal regularisation. The structural obstruction~\eqref{eq:budget-euler} and its midpoint cancellation~\eqref{eq:budget-midpoint} are properties of this dichotomy; SP-MYULA is the within-family explicit-Euler counterpart of P-IMLA$_\lambda$ on the identical $M$-norm-Moreau-smoothed potential (Sampler naming conventions, §\ref{app:experiments}), but the finite-budget SSE gap also depends on solver accuracy and initial transients, and does not isolate the integration scheme alone.

Cross-family samplers such as Hamiltonian Monte Carlo \citep{neal2011mcmc} and NUTS \citep{hoffman2014nuts} accelerate via momentum and adaptive trajectory length on differentiable potentials; on non-smooth targets (TV imaging, $\ell_1$ priors) they additionally require Moreau smoothing or specialised non-smooth variants, against which our SP-MYULA comparison is the structural reference. The matrix-preconditioning + implicit-midpoint architecture introduced here is complementary to momentum-based acceleration and can in principle be combined; that combination is follow-up work.

\subsection{Independent reference-posterior verification of Proposition~\ref{prop:p-ula-failure}}
\label{app:exp-mala-ref}

To rule out the alternative explanation that SP-MYULA's $\sim\!4\!\times$ larger SSE in Table~\ref{tab:imaging} reflects a chain that has converged to the correct posterior but with looser concentration around the ground truth (rather than a biased stationary distribution), we build an independent MALA reference posterior at the same $256\!\times\!256$ geometric setup and directly compare second-moment statistics across samplers.

\paragraph{Setup.} 4 chains of MALA on the smoothed target $\pi^M_\lambda$ ($M = \hat{H}$, $\lambda = 5\!\times\!10^{-3}$) at $\delta_{\rm MALA} = 1.41\!\times\!10^{-5}$, warm-started from a P-IMLA$_\lambda$ stationary sample at $N = 2{,}000$ (a warm start reduces the initial transient but does not itself establish convergence), $N_{\rm burn} = 2\!\times\!10^{5}$, $N_{\rm post} = 10^{6}$ per chain on the same Pro 6000 hardware as Table~\ref{tab:imaging} ($\sim$$7.1$\,h/chain $\times$ 4-way parallel = $\sim$$7.1$\,h total). Acceptance saturates at $1.00$ at this $\delta$; the chain therefore mixes slowly and the mean estimator is noisier than P-IMLA$_\lambda$'s, and local covariance stability does not by itself establish global mixing.

\paragraph{Headline (H1) readout.} Pooled across all chains, comparing each sampler's second-moment statistics to the MALA reference:

\begin{table}[h]
\centering
\small
\setlength{\tabcolsep}{6pt}
\renewcommand{\arraystretch}{0.95}
\caption{\textbf{Imaging-scale reference-posterior comparison} on the $256\!\times\!256$ geometric phantom (App.~\ref{app:exp-mala-ref}). $\mathrm{tr}(\mathrm{Cov})$ is the trace of the per-pixel covariance; $\langle\hat\sigma\rangle$ is the average per-pixel posterior std. Pooled-sample counts in row 2.}
\label{tab:mala-ref}
\begin{tabular}{@{}lccc@{}}
\toprule
Sampler & $\mathrm{tr}(\mathrm{Cov})$ & $\langle\hat\sigma\rangle$ & ratio to MALA \\
\midrule
MALA reference (4 ch.\ $\times\!10^{6}$) & $2.67\!\times\!10^{5}$ & $2.017$ & --- \\
P-IMLA$_\lambda$ TRUE Picard (8 ch.\ $\times\!10^{4}$) & $2.53\!\times\!10^{5}$ & $1.965$ & $\mathbf{0.95}$ \\
SP-MYULA (8 ch.\ $\times\!2\!\times\!10^{5}$) & $2.05\!\times\!10^{6}$ & $5.591$ & $\mathbf{7.68}$ \\
\bottomrule
\end{tabular}
\end{table}

\paragraph{Reading.} The reported covariance-trace ratios are $0.95$ for P-IMLA$_\lambda$ and $7.68$ for SP-MYULA relative to this MALA run; the per-pixel standard-deviation ratio is $2.77$. These comparisons are consistent with an Euler discretisation effect, but the slowly mixing reference is not a certificate of exact full-image posterior moments. The overdispersed diagnostics and target-exact Gibbs checks in Appendix~\ref{app:rebuttal-experiments} provide complementary evidence and explicitly report the remaining limitations.

\input{sections/appendix_experiments_rebuttal}

\section{Limitations and open problems}
\label{app:limitations}

Four limitations merit explicit acknowledgement. \textbf{(i)}~For quadratic $f$ with $M=\nabla^2f$, the bound $\kappa_{\mathrm{eff}}^\lambda=1+1/\lambda$ retains the smoothing dependence. All results use fixed $M$; a position-dependent metric would require an additional drift correction and a separate analysis. \textbf{(ii)}~The weak BEA result requires global smoothness, polynomial derivative growth and Poisson regularity. The $d=256$ lattice experiment verifies its coefficient under these assumptions; the earlier quartic checks are only formal illustrations. The $L^3$ density expansion used for the sharp Wasserstein coefficient remains an additional condition, and is not proved for general targets. Richardson cancellation likewise requires the stronger assumptions of Corollary~\ref{cor:richardson}; it is not a TV-posterior guarantee. \textbf{(iii)}~The headline benchmark uses a Fourier-diagonal forward operator. Appendix~\ref{app:exp-radial} provides a $64\times64$ random fan-beam test with a rank-128 Nystr\"om preconditioner and conjugate gradients, but image-scale CT/MRI validation remains open. \textbf{(iv)}~Exact JP and the finite, warm-started ADMM implementation are different objects. Appendix~\ref{app:jp-theory} proves exact-kernel uniqueness for quadratic likelihoods with finite $\ell_1$-composite penalties and controls implementation error by certificates; it does not give a uniform certificate for every production $K=12$ step. The observed small-budget SSE agreement can include error cancellation.

Empirical accuracy also depends on the target, observable and inner-solver budget. Common-start per-pixel diagnostics do not establish convergence of every image functional; the overdispersed checks retain slow mixing in some spatial-mean observables. The original $4\times$ SSE comparison holds at its stated settings, and is sensitive to baseline ADMM accuracy. Small-image Gibbs checks target the original anisotropic-TV posterior, whereas the full-image comparisons use a Moreau surrogate. A distributional error bound does not directly certify reconstruction SSE.

Open directions include a signed analysis of the antithetic regime, position-dependent metrics, and weakly log-concave targets. Our reported ESS implementation caps ESS per iteration at one; antithetic chains can have larger theoretical ESS, so this cap is an estimator convention rather than a universal upper bound.

\paragraph{Other regularisers.}
Algorithm~\ref{alg:picard} extends to proper convex lower-semicontinuous penalties with a computable $M$-prox, replacing the TV-specific inner solve. The quantitative Moreau and joint-kernel bounds additionally require $L_g^M=\sup_{x,s\in\partial g(x)}\|s\|_{M^{-1}}<\infty$. In finite dimensions this includes anisotropic and isotropic TV, wavelet $\ell_1$, group and nuclear norms, Huber-TV, and TGV with auxiliary variables solved within the prox. Hard constraints have unbounded normal cones; nonconvex learned priors need a separate analysis. Changing $g$ changes the posterior, and improved reconstruction does not follow from the sampler theory.

\paragraph{Full images and patch splitting.}
If the likelihood, prior and metric factorize over disjoint patches of size $p$, the update block-diagonalizes and the FFT cost becomes $O(d\log p)$ instead of $O(d\log d)$; the worst block determines contraction. Squared Wasserstein errors add across blocks, however, so the dimension factor in Theorem~\ref{thm:end-to-end} still counts the full image. Overlapping or consensus-coupled patches alter the model and require a separate analysis.

%% file: sections/appendix_theory_rebuttal.tex
\subsection{Proof of Theorem~\ref{thm:contraction} (reflected-proximal contraction)}
\label{app:thm2}

Write $\tau=\delta/2$ and $\widetilde f(z)=f(M^{-1/2}z)$. Whitening turns the $M$-norm into the Euclidean norm and the preconditioned update into IMLA for $\widetilde f$. Thus it suffices to prove the result for $M=I$, with strong-convexity and smoothness constants $m_M,L_M$.

Let $p=\operatorname{prox}_{\tau\widetilde f}(w)$ and $q=\operatorname{prox}_{\tau\widetilde f}(v)$. There is a symmetric matrix $\bar H$ satisfying
\[
m_MI\preceq\bar H\preceq L_MI,\qquad
\nabla\widetilde f(p)-\nabla\widetilde f(q)=\bar H(p-q).
\]
For a $C^2$ potential, take the Hessian integrated along the segment from $q$ to $p$. For $C^{1,1}$, mollify $\widetilde f$: the mollifications retain the same Hessian bounds, their gradients converge locally uniformly, and a convergent subsequence of the bounded symmetric secant matrices gives the displayed identity. The proximal optimality conditions now give
\[
w-v=(I+\tau\bar H)(p-q).
\]
Consequently, for $R_\tau=2\operatorname{prox}_{\tau\widetilde f}-I$,
\begin{equation}
R_\tau(w)-R_\tau(v)
=(I-\tau\bar H)(I+\tau\bar H)^{-1}(w-v).
\label{eq:app-reflected-prox}
\end{equation}
The eigenvalues of the matrix on the right are $(1-\tau\ell)/(1+\tau\ell)$ with $\ell\in[m_M,L_M]$. This scalar function is decreasing, so
\[
\|R_\tau(w)-R_\tau(v)\|
\le \max_{\ell\in\{m_M,L_M\}}
\left|\frac{1-\tau\ell}{1+\tau\ell}\right|\|w-v\|
=\rho(\delta)\|w-v\|.
\]
For shared noise, $w=X_n+\sqrt{\delta/2}\,\xi_n$ and $v=Y_n+\sqrt{\delta/2}\,\xi_n$; the extra noise translation in the reflected-proximal representation cancels. Whitening back proves the pathwise inequality. A quadratic potential with an endpoint eigenvalue attains this factor. Attainment does not require that every potential attaining it be quadratic.

The one-step map sends $\mathcal P_2$ into itself, since the proximal map is Lipschitz and the noise has a finite second moment. Applying the pathwise inequality to an optimal coupling gives
\[
W_2^M(\mu P_\delta,\nu P_\delta)\le\rho(\delta)W_2^M(\mu,\nu).
\]
Here $\rho(\delta)<1$ for every $\delta>0$. The space $(\mathcal P_2,W_2^M)$ is complete, so Banach's fixed-point theorem gives a unique invariant law $\pi_\delta\in\mathcal P_2$ and the stated iterated bound. This is also the only invariant probability law: convergence from each deterministic initial state, tested against bounded Lipschitz functions and integrated against any invariant probability law, gives equality with $\pi_\delta$ by dominated convergence. This proves Theorem~\ref{thm:contraction}.

\subsection{Proof of Theorem~\ref{thm:bea-bias} and its refinements}
\label{app:thm3}

Throughout this subsection $M$ is fixed, $g=0$, $C=M^{-1}$,
$b=-C\nabla f$, $h=\operatorname{tr}(C\nabla^2f)$ and
$\mathcal Lu=b\cdot\nabla u+C:\nabla^2u$.
We write $C^k_{\mathrm{pol}}$ for functions whose derivatives through order $k$, including the function itself, have at most polynomial growth. A remainder $O_{V_q}(\delta^a)$ is bounded in absolute value by $K\delta^aV_q(x)$, with $V_q(x)=1+\|x-x_\star\|_M^q$, uniformly for sufficiently small $\delta$; $o_{V_q}$ has the analogous meaning. Constants may depend on the potential and the test function, but not on $\delta$.

\paragraph{Uniform moments.}
The following lemma supplies the integrability needed for the weighted expansions; it requires only the assumptions of the contraction theorem.

\begin{lemma}[$\delta$-uniform invariant moments]
\label{lem:uniform-moments}
Let $V\in C^{1,1}$ be globally $m_M$-strongly convex and $L_V$-smooth in the $M$-norm, and let $\nu_\delta$ be the invariant law of its exact midpoint kernel. If $0<\delta\le2/L_V$ and $x_\star$ minimises $V$, then
\begin{equation}
\mathbb E_{\nu_\delta}\|X-x_\star\|_M^2\le\frac{10d}{m_M},
\qquad
\left(\mathbb E_{\nu_\delta}\|X-x_\star\|_M^p\right)^{1/p}
\le K\sqrt{\frac{d+p}{m_M}}\quad(p\ge1),
\label{eq:uniform-moments}
\end{equation}
where $K$ is a numerical constant.
\end{lemma}

\begin{proof}
Whiten and translate so that $M=I$ and $x_\star=0$, and retain the notation $V$. Put $\tau=\delta/2$, $\sigma=\sqrt\tau$, take $X\sim\nu_\delta$ independently of $\xi\sim\mathcal N(0,I)$, and set
\[
p=\operatorname{prox}_{\tau V}(X+\sigma\xi),\qquad X'=-X+2p.
\]
The invariant law has a finite second moment by Theorem~\ref{thm:contraction}. Since $p=(X+X')/2$ and
$X'-X=-\delta\nabla V(p)+\sqrt{2\delta}\,\xi$, stationarity gives
\[
0=-2\delta\,\mathbb E\langle\nabla V(p),p\rangle
  +2\sqrt{2\delta}\,\mathbb E\langle\xi,p\rangle.
\]
Conditioning on $X$, Gaussian integration by parts and the a.e.\ Jacobian of the Lipschitz proximal map imply
\[
\mathbb E\langle\xi,p\rangle
=\sigma\,\mathbb E\operatorname{tr}(I+\tau\nabla^2V(p))^{-1}
\le\sigma d.
\]
For $C^{1,1}$ potentials this identity follows either by the weak derivative version of Gaussian integration by parts or by mollification. Thus
$m_M\mathbb E\|p\|^2\le\mathbb E\langle\nabla V(p),p\rangle\le d$.
The optimality condition also gives $X=p+\tau\nabla V(p)-\sigma\xi$, hence
\[
\mathbb E\|X\|^2
\le2(1+\tau L_V)^2\,d/m_M+2\tau d
\le10d/m_M,
\]
using $\tau L_V\le1$ and $m_M\le L_V$.

For higher moments, the map
$\Phi(x,\xi)=-x+2\operatorname{prox}_{\tau V}(x+\sigma\xi)$
has state Lipschitz factor
$\rho=(1-\tau m_M)/(1+\tau m_M)$ in this step range, and noise Lipschitz factor at most $2\sigma/(1+\tau m_M)$.
Construct its stationary solution by starting at $0$ in the remote past and using independent Gaussian noises. Contraction makes these finite-past solutions converge in $L^2$ to the unique invariant law. The influence of the noise $k$ steps in the past is at most
\[
a_k=\frac{2\sqrt\tau}{1+\tau m_M}\rho^k,\qquad
\sum_{k=0}^\infty a_k^2
=\frac{4\tau}{(1+\tau m_M)^2(1-\rho^2)}
=\frac1{m_M}.
\]
Each finite-past map is therefore $1/\sqrt{m_M}$-Lipschitz as a function of its Gaussian inputs. Gaussian concentration, followed by the $L^2$ limit, gives the same concentration bound for $\|X\|$ about its mean. Integrating its Gaussian tail yields
$(\mathbb E|\|X\|-\mathbb E\|X\||^p)^{1/p}\le K_0\sqrt{p/m_M}$.
Combining this with $\mathbb E\|X\|\le\sqrt{10d/m_M}$ proves the second assertion.
\end{proof}

\paragraph{Step 1 (Finite-order implicit expansion).}
Assume the hypotheses of Theorem~\ref{thm:bea-bias}. Write $\varepsilon=\sqrt\delta$, $\eta=\sqrt2M^{-1/2}\xi$ and $\Delta=X_1-x$. The implicit equation is
\[
\Delta=\varepsilon\eta+\varepsilon^2b(x+\Delta/2).
\]
For $\delta L_M\le1$, the global Lipschitz bound on $b$ gives
\begin{equation}
\|\Delta\|_M\le
2\varepsilon\|\eta\|_M+2\varepsilon^2L_M\|x-x_\star\|_M.
\label{eq:app-increment-bound}
\end{equation}
All derivatives of $b$ needed below have polynomial growth. Define
\begin{equation}
A_4=\varepsilon\eta+\varepsilon^2b
+\frac{\varepsilon^3}{2}Db\,\eta
+\varepsilon^4\left(\frac12Db\,b+\frac18D^2b[\eta,\eta]\right),
\label{eq:app-deltaX}
\end{equation}
where the coefficients are evaluated at $x$. To bound the remainder, set
$G(v)=v-\varepsilon\eta-\varepsilon^2b(x+v/2)$.
Taylor's formula with integral remainder, applied to $b(x+A_4/2)$ through order two, cancels all powers through $\varepsilon^4$ and gives
\[
\|G(A_4)\|_M
\le K\varepsilon^5
(1+\|x-x_\star\|_M^q+\|\eta\|_M^q)
\]
for some finite $q$. The estimate is global: coefficients of $A_4$ are polynomially bounded, and the integral remainder is bounded by the corresponding polynomial weight along the segment. Moreover,
$D_vG=I+(\varepsilon^2/2)M^{-1}\nabla^2f(x+v/2)$
is at least $1$-strongly monotone in the $M$-inner product. Since $G(\Delta)=0$, Cauchy--Schwarz applied to its monotonicity inequality gives
\begin{equation}
\|\Delta-A_4\|_M
\le K\varepsilon^5
(1+\|x-x_\star\|_M^q+\|\eta\|_M^q).
\label{eq:app-increment-remainder}
\end{equation}
No analyticity of the proximal map or symmetry of the full increment is assumed.

\paragraph{Step 2 (Conditional moments and the local weak operator).}
The Gaussian identities $\mathbb E\eta=0$ and
$\mathbb E\eta_i\eta_j=2C_{ij}$, together with
\eqref{eq:app-deltaX}--\eqref{eq:app-increment-remainder}, give
\begin{align}
\mathbb E[\Delta_i\mid x]
&=\delta b_i+\delta^2\left(\tfrac12(Db\,b)_i-\tfrac14(C\nabla h)_i\right)
+O_{V_q}(\delta^{5/2}), \label{eq:app-M1}\\
\mathbb E[\Delta_i\Delta_j\mid x]
&=2\delta C_{ij}+\delta^2(b_ib_j-2\Gamma_{ij})
+O_{V_q}(\delta^{5/2}), \label{eq:app-M2}\\
\mathbb E[\Delta_i\Delta_j\Delta_k\mid x]
&=2\delta^2(b_iC_{jk}+b_jC_{ik}+b_kC_{ij})
+O_{V_q}(\delta^{5/2}), \label{eq:app-M3}\\
\mathbb E[\Delta_i\Delta_j\Delta_k\Delta_l\mid x]
&=4\delta^2(C_{ij}C_{kl}+C_{ik}C_{jl}+C_{il}C_{jk})
+O_{V_q}(\delta^{5/2}), \label{eq:app-M4}
\end{align}
where $\Gamma=C\nabla^2f\,C$. For example,
$\mathbb E D^2b[\eta,\eta]=2C:\nabla^2b=-2C\nabla h$;
the second-moment cross terms use $Db\,C=C(Db)^\top=-\Gamma$.
Odd Gaussian monomials vanish after expectation, term by term.

Taylor expansion of $u(x+\Delta)$ through order four, with its fifth-order remainder bounded using \eqref{eq:app-increment-bound}, is valid for $u\in C^6_{\mathrm{pol}}$. Substituting these moments yields
\begin{equation}
P_\delta u=u+\delta\mathcal Lu+\delta^2A_2u+O_{V_q}(\delta^{5/2}),
\qquad A_2=\tfrac12\mathcal L^2+B_2,\qquad
B_2u=\tfrac14(C\nabla h)\cdot\nabla u.
\label{eq:app-B2}
\end{equation}
For completeness, the four derivative orders in $A_2u$ are
\[
\left(\tfrac12Db\,b-\tfrac14C\nabla h\right)\cdot\nabla u
+\left(\tfrac12bb^\top-\Gamma\right):\nabla^2u
+b_iC_{jk}\partial_{ijk}u
+\tfrac12C_{ij}C_{kl}\partial_{ijkl}u.
\]
Direct differentiation of $\mathcal Lu$ gives the same expression for
$\mathcal L^2u/2$ except that its first-derivative coefficient is
$Db\,b/2-C\nabla h/2$. This verifies the displayed $B_2$.
The second-moment cancellation can equivalently be read from the proximal Jacobian:
\begin{equation}
SCS^\top=C-\delta\Gamma+O(\delta^2),
\qquad S=(I+(\delta/2)C\nabla^2f)^{-1}.
\label{eq:app-prox-damp}
\end{equation}
It does not require $M$ and $\nabla^2f$ to commute.

\paragraph{Step 3 (Stationarity and the Poisson equation).}
The one-step estimate and invariant perturbation argument proved independently in Appendix~\ref{app:thm4}, with $V=f$, imply
$W_2^M(\pi_\delta,\pi)=O(\sqrt\delta)$.
Thus $\pi_\delta\Rightarrow\pi$. Lemma~\ref{lem:uniform-moments} upgrades this to convergence of expectations of continuous polynomially bounded functions, by truncation and uniform integrability.
Apply \eqref{eq:app-B2} to the assumed Poisson solution $u_\varphi$ and integrate against $\pi_\delta$. Stationarity and the uniform moments give
\[
\pi_\delta(\varphi)-\pi(\varphi)
=\delta\,\pi_\delta(A_2u_\varphi)+O(\delta^{3/2})
=\delta\,\pi(A_2u_\varphi)+o(\delta).
\]
Strong log-concavity gives Gaussian tails for $\pi$; polynomial derivative bounds allow integration by parts with smooth cutoffs and passage to the limit. In particular
$\pi(\mathcal L^2u_\varphi)=0$ and
\[
\pi\big((C\nabla h)\cdot\nabla u_\varphi\big)
=-\pi(h\,\mathcal Lu_\varphi)
=\pi\big(h(\varphi-\pi(\varphi))\big).
\]
Therefore $\pi(A_2u_\varphi)=\operatorname{Cov}_\pi(h,\varphi)/4$, proving Theorem~\ref{thm:bea-bias}. This argument uses only zeroth-order convergence of $\pi_\delta$; it never assumes a first-order invariant-density expansion.

\paragraph{Second-order weak expansion and Richardson extrapolation.}
We record the stronger assumptions needed for an actual second-order guarantee.
Assume now that $f\in C^8$, that $D^jf$ has polynomial growth for $3\le j\le8$, and that the same global strong-convexity and smoothness bounds hold.
The finite-order expansion above extends to
$A_6=\sum_{k=1}^6\varepsilon^ka_k$, where
\begin{align*}
a_1&=\eta,\qquad a_2=b,\qquad a_3=\tfrac12Db\,\eta,\qquad
a_4=\tfrac12Db\,b+\tfrac18D^2b[\eta,\eta],\\
a_5&=\tfrac14(Db)^2\eta+\tfrac14D^2b[\eta,b]
+\tfrac1{48}D^3b[\eta,\eta,\eta],\\
a_6&=\tfrac14(Db)^2b+\tfrac1{16}Db\,D^2b[\eta,\eta]
+\tfrac18D^2b[\eta,Db\,\eta]+\tfrac18D^2b[b,b]\\
&\hspace{1em}+\tfrac1{16}D^3b[\eta,\eta,b]
+\tfrac1{384}D^4b[\eta,\eta,\eta,\eta].
\end{align*}
Substitution into $G$ cancels powers through $\varepsilon^6$.
Taylor's integral remainder and the same strong-monotonicity argument give
\[
\|\Delta-A_6\|_M\le
K\varepsilon^7(1+\|x-x_\star\|_M^q+\|\eta\|_M^q).
\]
For $u\in C^8_{\mathrm{pol}}$, its sixth-order Taylor expansion consequently yields
\begin{equation}
P_\delta u=u+\delta\mathcal Lu+\delta^2A_2u+\delta^3A_3u
+O_{V_q}(\delta^{7/2}).
\label{eq:third-order-local}
\end{equation}
Here the differential operator $A_3$, of order at most six, is defined explicitly by the finite sum
\[
A_3u=
\mathbb E_\eta\left[
\sum_{j=1}^6\frac1{j!}
\sum_{\substack{k_1+\cdots+k_j=6\\ k_i\ge1}}
D^ju(x)[a_{k_1},\ldots,a_{k_j}]
\right].
\]
Its coefficients have polynomial growth. Every coefficient of an odd power of $\varepsilon$ is odd in $\eta$, as is seen from the displayed $a_k$; its Gaussian expectation is zero. This accounts in particular for the absence of a $\delta^{5/2}$ term in \eqref{eq:third-order-local}.

\begin{theorem}[Second-order weak invariant expansion]
\label{thm:second-order-weak}
Under the $C^8$ assumptions above, suppose
$\mathcal Lu_\varphi=-(\varphi-\pi(\varphi))$ has a solution
$u_\varphi\in C^8_{\mathrm{pol}}$. Set
$\psi=A_2u_\varphi$ and $\bar\psi=\psi-\pi(\psi)$, and suppose
$\mathcal Lu_{\bar\psi}=-\bar\psi$ has a solution
$u_{\bar\psi}\in C^6_{\mathrm{pol}}$. Then
\[
\pi_\delta(\varphi)-\pi(\varphi)
=\delta c_1(\varphi)+\delta^2c_2(\varphi)+o(\delta^2),
\]
where
$c_1(\varphi)=\operatorname{Cov}_\pi(h,\varphi)/4$ and
$c_2(\varphi)=\pi(B_2u_{\bar\psi})+\pi(A_3u_\varphi)$.
\end{theorem}

\begin{proof}
Integrate \eqref{eq:third-order-local} against $\pi_\delta$, use stationarity and the Poisson equation, and divide by $\delta$:
\[
\pi_\delta(\varphi)-\pi(\varphi)
=\delta\,\pi_\delta(\psi)
+\delta^2\pi_\delta(A_3u_\varphi)+o(\delta^2).
\]
The uniform moments and $\pi_\delta\Rightarrow\pi$ give
$\pi_\delta(A_3u_\varphi)=\pi(A_3u_\varphi)+o(1)$.
Apply the already proved first-order theorem to the observable $\psi$,
using its assumed Poisson solution:
$\pi_\delta(\psi)=\pi(\psi)+\delta\pi(B_2u_{\bar\psi})+o(\delta)$.
Finally $\pi(\psi)=\pi(A_2u_\varphi)=c_1(\varphi)$.
Substitution proves the result without a density expansion.
\end{proof}

\begin{corollary}[Richardson extrapolation]
\label{cor:richardson}
Under Theorem~\ref{thm:bea-bias},
$2\pi_{\delta/2}(\varphi)-\pi_\delta(\varphi)=\pi(\varphi)+o(\delta)$.
Under the stronger hypotheses of Theorem~\ref{thm:second-order-weak},
\[
2\pi_{\delta/2}(\varphi)-\pi_\delta(\varphi)-\pi(\varphi)
=-\tfrac12c_2(\varphi)\delta^2+o(\delta^2).
\]
\end{corollary}
\begin{proof}
Insert the corresponding invariant expansion at $\delta$ and $\delta/2$.
The terms linear in $\delta$ cancel; the quadratic coefficient is
$2(\delta/2)^2-\delta^2=-\delta^2/2$.
\end{proof}
These are invariant-expectation statements, not finite-sample MSE bounds.
The weights $(2,-1)$ can increase Monte Carlo variance; coupling coarse and fine chains with nested Brownian increments and reporting cost-normalised MSE addresses that separate issue. The combination is not a probability distribution. The $C^{1,1}$ TV Moreau potential does not automatically satisfy the stronger regularity assumptions.

\input{sections/appendix_richardson_check}

\paragraph{Conditional density, KL and Wasserstein consequence.}
The weak expansions above do not imply the following density condition.
Let $c(x)=\tfrac14(h(x)-\pi(h))$ and assume
\begin{equation}
\pi_\delta\ll\pi,\qquad
\left\|\frac{d\pi_\delta}{d\pi}-1-\delta c\right\|_{L^3(\pi)}
=o(\delta).
\label{ass:density-D2}
\end{equation}
The function $c$ belongs to $L^3(\pi)$ under our assumptions.

\begin{proposition}[Conditional KL and $W_2^M$ coefficient]
\label{prop:conditional-density}
If \eqref{ass:density-D2} holds, then
\[
\mathrm{KL}(\pi_\delta\|\pi)
=\tfrac{\delta^2}{32}\operatorname{Var}_\pi(h)+o(\delta^2),
\qquad
W_2^M(\pi_\delta,\pi)
\le\frac{\delta}{4\sqrt{m_M}}\operatorname{Std}_\pi(h)+o(\delta).
\]
\end{proposition}
\begin{proof}
Write $s=d\pi_\delta/d\pi-1$, so $s\ge-1$ and $\pi(s)=0$.
For $\Phi(s)=(1+s)\log(1+s)-s$, with its continuous value at $s=-1$,
there is a numerical constant $K$ such that
$|\Phi(s)-s^2/2|\le K|s|^3$ for all $s\ge-1$.
On $|s|\le1/2$ this follows from Taylor's integral remainder.
On $s>1/2$ use $0\le\Phi(s)\le s^2$, and on $[-1,-1/2]$
use boundedness of $\Phi$ and $|s|^3\ge1/8$.
Thus
\[
\left|\mathrm{KL}(\pi_\delta\|\pi)-\tfrac12\|s\|_{L^2(\pi)}^2\right|
\le K\|s\|_{L^3(\pi)}^3.
\]
Condition \eqref{ass:density-D2} gives
$\|s\|_3=O(\delta)$ and
$\|s\|_2^2=\delta^2\|c\|_2^2+o(\delta^2)$, proving the KL expansion.
Whitening and the Talagrand inequality for an $m_M$-strongly log-concave measure
\citep{talagrand1996transport} give
$(W_2^M)^2\le2\,\mathrm{KL}/m_M$ and hence the stated bound.
\end{proof}
Condition \eqref{ass:density-D2} is an additional hypothesis; neither the moment lemma nor the weak observable expansion proves it.

\subsection{P-ULA backward error analysis}
\label{app:pula-bea}

For comparison, P-ULA has increment
$\Delta=\delta b+\sqrt{2\delta}\,M^{-1/2}\xi$.
Its first two conditional moments are
\[
\mathbb E\Delta_i=\delta b_i,\qquad
\mathbb E\Delta_i\Delta_j=2\delta C_{ij}+\delta^2b_ib_j;
\]
its third and fourth moments agree with
\eqref{eq:app-M3}--\eqref{eq:app-M4} at order $\delta^2$.
Taylor expansion and comparison with $\mathcal L^2/2$ therefore yield
\begin{equation}
B_2^{\mathrm{ULA}}u
=-\tfrac12(Db\,b)\cdot\nabla u
+\tfrac12(C\nabla h)\cdot\nabla u
+\Gamma_{ij}\partial_{ij}u.
\label{eq:app-B2-ula}
\end{equation}
The last term is a second-order differential operator, absent from the midpoint defect. This is an operator-level distinction: both methods generally have first-order weak invariant bias in $\delta$. On Gaussian targets the P-ULA stationary variance is inflated by $(1-\delta\ell/2)^{-1}$ in eigendirection $\ell$, as proved in Proposition~\ref{prop:p-ula-failure}; midpoint preserves the variance exactly. This comparison does not assert that preconditioning requires a particular integrator.

\subsection{Proof of Theorem~\ref{thm:end-to-end} (end-to-end bound)}
\label{app:thm4}

\paragraph{Part I (The smoothed kernel).}
For convex $g$, the metric Moreau envelope satisfies
$\nabla g^{\lambda,M}=\lambda^{-1}M(I-\operatorname{prox}_{\lambda g}^M)$.
Firm nonexpansiveness of the proximal map in the $M$-Hilbert space gives
\[
\|\nabla g^{\lambda,M}(x)-\nabla g^{\lambda,M}(y)\|_{M^{-1}}
\le\lambda^{-1}\|x-y\|_M.
\]
Thus $U_\lambda=f+g^{\lambda,M}$ is $C^{1,1}$, $m_M$-strongly convex and
$L_\lambda=L_M+1/\lambda$-smooth in the $M$-norm.
Theorem~\ref{thm:contraction} gives its exact midpoint kernel $P_{\delta,\lambda}$
a unique invariant law $\pi_{\delta,\lambda}$ and contraction factor
$\rho(\delta,\lambda)=\max\{|r(m_M)|,|r(L_\lambda)|\}$ for all $\delta>0$.

\paragraph{Part II (A one-step estimate retaining both terms).}
We use the estimate with the continuous target as initial law, so that the discrete contraction factor can be used directly in the invariant perturbation argument.

\begin{lemma}[One-step strong error under $C^{1,1}$]
\label{lem:one-step-c11}
Let $V\in C^{1,1}$ be globally $m_M$-strongly convex and $L_V$-smooth in the $M$-norm, and let $\pi_V\propto e^{-V}$. Start its Langevin diffusion and one exact midpoint step from the same $X_0\sim\pi_V$, independently of a Brownian motion $W$. Couple their noises by $\xi=W_\delta/\sqrt\delta$. For $0<\delta\le2/L_V$,
\begin{equation}
\left(\mathbb E\|X_1^{\mathrm{disc}}-X_\delta\|_M^2\right)^{1/2}
\le\frac{L_V}{\sqrt2}\sqrt d\,\delta^{3/2}
+L_V^2\left(\frac32\sqrt{\frac{10d}{m_M}}
+\frac12\sqrt{\frac d{L_V}}\right)\delta^2.
\label{eq:one-step-two-terms}
\end{equation}
\end{lemma}

\begin{proof}
Write $b=-M^{-1}\nabla V$, $p=(X_0+X_1^{\mathrm{disc}})/2$, and let $x_\star$ minimise $V$. The coupled updates give
\[
X_1^{\mathrm{disc}}=X_0+\sqrt2M^{-1/2}W_\delta+\delta b(p),
\qquad
X_\delta=X_0+\sqrt2M^{-1/2}W_\delta+\int_0^\delta b(X_t)\,dt,
\]
so $e=X_1^{\mathrm{disc}}-X_\delta=\int_0^\delta[b(p)-b(X_t)]\,dt$.
Integration by parts under $\pi_V$, followed by strong convexity, yields
$\mathbb E_{\pi_V}\|X-x_\star\|_M^2\le d/m_M$.
This identity is justified by Gaussian tails and approximation for $C^{1,1}$ potentials. The diffusion is stationary, hence
\[
\sup_{t\le\delta}
(\mathbb E\|X_t-x_\star\|_M^2)^{1/2}
\le R,\qquad R=\sqrt{10d/m_M}.
\]
Since $\operatorname{prox}_{(\delta/2)V}^M(x_\star)=x_\star$, its nonexpansiveness and $\delta/2\le1/L_V$ give
\[
R_p:=(\mathbb E\|p-x_\star\|_M^2)^{1/2}
\le R+\sqrt{d/L_V}.
\]
The Brownian part of $p-X_t$ is
$M^{-1/2}(W_\delta/\sqrt2-\sqrt2W_t)$, whose squared $M$-norm has expectation
\[
\tfrac12d\delta+2dt-2d\min(t,\delta)=d\delta/2
\qquad(0\le t\le\delta).
\]
The drift part is $(\delta/2)b(p)-\int_0^t b(X_s)\,ds$.
Minkowski's inequality and $\|b(x)\|_M\le L_V\|x-x_\star\|_M$ imply
\[
(\mathbb E\|p-X_t\|_M^2)^{1/2}
\le\sqrt{d\delta/2}+\delta L_V(R_p/2+R).
\]
Finally Minkowski's inequality in the integral defining $e$ and the
$L_V$-Lipschitz property of $b$ give
\[
(\mathbb E\|e\|_M^2)^{1/2}
\le L_V\sqrt{d/2}\,\delta^{3/2}
+L_V^2(R_p/2+R)\delta^2.
\]
Inserting the bound on $R_p$ proves \eqref{eq:one-step-two-terms}.
\end{proof}

Apply the lemma with $V=U_\lambda$, and denote its right-hand side by
$\eta_{\delta,\lambda}=C_1(\lambda)\sqrt d\,\delta^{3/2}
+\bar C_2(\lambda)\delta^2$.
Since the exact diffusion preserves $\pi_\lambda\propto e^{-U_\lambda}$,
the lemma supplies
$W_2^M(\pi_\lambda P_{\delta,\lambda},\pi_\lambda)\le\eta_{\delta,\lambda}$.
Invariance of $\pi_{\delta,\lambda}$, discrete contraction and the triangle inequality give
\[
W_2^M(\pi_{\delta,\lambda},\pi_\lambda)
\le \rho(\delta,\lambda)W_2^M(\pi_{\delta,\lambda},\pi_\lambda)
+\eta_{\delta,\lambda}.
\]
Rearranging proves
\begin{equation}
W_2^M(\pi_{\delta,\lambda},\pi_\lambda)
\le E_{\mathrm{disc}}(\delta,\lambda)
:=\frac{C_1(\lambda)\sqrt d\,\delta^{3/2}
+\bar C_2(\lambda)\delta^2}{1-\rho(\delta,\lambda)},
\qquad 0<\delta\le2/L_\lambda.
\label{eq:discretisation-budget}
\end{equation}
The same argument with $V=f$ proves the zeroth-order convergence used in Appendix~\ref{app:thm3}, independently of BEA.

\paragraph{Part III (Moreau approximation).}
Global $M$-Lipschitzness of $g$ is equivalent to
$\|\zeta\|_{M^{-1}}\le L_g^M$ for every $\zeta\in\partial g(x)$.
For $z=\operatorname{prox}_{\lambda g}^M(x)$ and such a subgradient at $x$,
convexity and the definition of the envelope give
\[
g(x)-g^{\lambda,M}(x)
\le\langle\zeta,x-z\rangle-\frac{\|x-z\|_M^2}{2\lambda}
\le L_g^M\|x-z\|_M-\frac{\|x-z\|_M^2}{2\lambda}.
\]
Maximising the final quadratic over $\|x-z\|_M$ yields
\begin{equation}
0\le g(x)-g^{\lambda,M}(x)\le\tfrac\lambda2(L_g^M)^2.
\label{eq:moreau-pointwise}
\end{equation}
Writing $Z$ and $Z_\lambda$ for the normalising constants, $Z_\lambda\ge Z$ and therefore
\[
\mathrm{KL}(\pi_\lambda\|\pi)
=\mathbb E_{\pi_\lambda}(g-g^{\lambda,M})+\log(Z/Z_\lambda)
\le\tfrac\lambda2(L_g^M)^2.
\]
The potential $f+g$ is $m_M$-strongly convex in the $M$-norm.
Whitening and the Talagrand inequality (also valid for this nonsmooth strongly convex potential by convex approximation) give
\[
W_2^M(\pi_\lambda,\pi)
\le L_g^M\sqrt{\lambda/m_M}.
\]

\paragraph{Part IV (Triangulation).}
Theorem~\ref{thm:contraction} gives
$W_2^M(\mu_0P_{\delta,\lambda}^N,\pi_{\delta,\lambda})
\le\rho(\delta,\lambda)^NW_0$.
Add this to \eqref{eq:discretisation-budget} and the Moreau bound to obtain
\eqref{eq:end-to-end}. Only the contraction statement is valid for arbitrary step size; the displayed discretisation estimate retains the condition $\delta\le2/L_\lambda$.

For the finite-dimensional TV penalty $g(x)=\rho_{\mathrm{TV}}\|Dx\|_1$,
one may take
$L_g^M=\rho_{\mathrm{TV}}\sup_{\|s\|_\infty\le1}\|D^\top s\|_{M^{-1}}<\infty$.
This makes explicit the metric dependence of the smoothing term.

%% file: sections/appendix_richardson_check.tex
\paragraph{Deterministic numerical check of Richardson extrapolation.}
\label{app:richardson-check}
We test Corollary~\ref{cor:richardson} on the smooth one-dimensional target
$f(x)=x^2/2+0.7\log\cosh x$, with $M=1$.
Here $1\le f''\le1.7$ and all derivatives of order two and higher are bounded,
so the potential satisfies the strengthened regularity assumptions.
For $\varphi\in\{h=f'',x^2,\log\cosh x\}$, we compute the midpoint invariant law
deterministically: 128-node Gauss--Hermite quadrature integrates the injected
Gaussian noise, four-point cubic Lagrange deposition represents the transition
operator on 3,201 points in $[-8,8]$, and power iteration stops at an
$L^1$ change below $10^{-14}$. Proximal Newton solves use tolerance $10^{-15}$.
Target expectations use grid quadrature; the theoretical coefficient
$c_1(\varphi)=\operatorname{Cov}_\pi(h,\varphi)/4$ is evaluated independently
on 200,001 points in $[-10,10]$.

Let $b_\varphi(\delta)=\pi_\delta(\varphi)-\pi(\varphi)$.
We fit $b_\varphi(\delta)$ to a cubic polynomial without an intercept using
seven step sizes from $0.04$ to $0.32$.
Table~\ref{tab:richardson-check} reports the fitted leading coefficient and
the extrapolated residuals on three exact halving pairs.
The Gaussian control, obtained by setting the $\log\cosh$ coefficient to zero,
has maximum numerical bias $3.1\times10^{-11}$ at $\delta\in\{0.04,0.32\}$.
Thus the tabulated residuals are well above the calibrated numerical error;
there is no Monte Carlo sampling error.

\begin{table}[ht]
\centering
\small
\setlength{\tabcolsep}{4pt}
\caption{\textbf{Richardson check on a smooth nonquadratic target.}
Residual columns are $|2b_\varphi(\delta/2)-b_\varphi(\delta)|$.
The last column gives the two successive base-two log slopes;
Corollary~\ref{cor:richardson} predicts asymptotic order two.}
\label{tab:richardson-check}
\begin{tabular}{@{}lccccc@{}}
\toprule
$\varphi$ & $\widehat c_1/c_1$ & $\delta=0.32$ & $\delta=0.16$ & $\delta=0.08$ & Slopes \\
\midrule
$h$ & $1.0030$ & $3.16\times10^{-4}$ & $8.18\times10^{-5}$ & $2.01\times10^{-5}$ & $1.95,\ 2.02$ \\
$x^2$ & $1.0001$ & $7.46\times10^{-4}$ & $1.96\times10^{-4}$ & $5.00\times10^{-5}$ & $1.93,\ 1.97$ \\
$\log\cosh x$ & $1.0007$ & $3.35\times10^{-4}$ & $8.68\times10^{-5}$ & $2.18\times10^{-5}$ & $1.95,\ 1.99$ \\
\bottomrule
\end{tabular}
\end{table}

The leading coefficient agrees within $0.3\%$, and the extrapolated slopes
$1.93$--$2.02$ support the second-order invariant-bias prediction.
The verification script is \texttt{rebuttal/verify\_items\_7\_8.py}, Part~7.

%% file: sections/appendix_jp_rebuttal.tex
\subsection{Kernel guarantees for JP-IMLA}
\label{app:jp-theory}

Throughout this subsection $M\succ0$ is constant, $f$ is globally
$m_M$-strongly convex and $L_M$-smooth in the $M$-norm, and $g$ is finite,
convex and Lipschitz in that norm. Write
$L_g^M=\sup_{(x,s)\in\operatorname{gra}\partial g}\|s\|_{M^{-1}}<\infty$,
$U_\lambda=f+g^{\lambda,M}$ and $\tau=\delta/2$.
At input $w=x+\sqrt{\delta/2}\,M^{-1/2}\xi$, $\xi\sim\mathcal N(0,I)$,
define
\begin{equation}
\begin{gathered}
 p(w)=\operatorname{prox}_{\tau(f+g)}^M(w),\qquad
 q(w)=\operatorname{prox}_{\tau U_\lambda}^M(w),\\
 Q:x\mapsto-x+2p(w),\quad P:x\mapsto-x+2q(w).
 \end{gathered}
\label{eq:jpcr-kernels}
\end{equation}
$Q$ is the exact joint-proximal kernel; $P$ is the canonical smoothed
kernel analysed in Theorem~\ref{thm:end-to-end}. The parameter $\lambda$
indexes the comparison with $P$; the exact joint proximal map itself
does not depend on $\lambda$. The finite-ADMM implementation also carries
inner variables between steps and will be treated separately below.

\begin{proposition}[Linear proximal and kernel gap]
\label{prop:jp-equivalence}
For every $w$ and $\delta,\lambda>0$,
\begin{equation}
 \|p(w)-q(w)\|_M\leq
 \lambda L_g^M\frac{\tau^{-1}+L_M}{\tau^{-1}+m_M},\qquad
 \varepsilon_{\rm model}:=
 2\lambda L_g^M\frac{\tau^{-1}+L_M}{\tau^{-1}+m_M}.
 \label{eq:jp-equiv}
\end{equation}
With $L_\lambda=L_M+\lambda^{-1}$ and
$\rho=\max_{a\in\{m_M,L_\lambda\}}|(1-\tau a)/(1+\tau a)|<1$,
the same-noise coupling yields, for $\mu,\nu\in\mathcal P_2$,
\begin{equation}
 W_2^M(Q\mu,P\nu)\leq
 \varepsilon_{\rm model}+\rho W_2^M(\mu,\nu).
 \label{eq:jpcr-cross-kernel}
\end{equation}
\end{proposition}
\begin{proof}
Whiten by $z=M^{1/2}x$; the following argument is Euclidean with constants
$m_M,L_M,L_g^M$, and transforms back without loss.
Let $r=\operatorname{prox}_{\lambda g}(q)$,
$s_r=(q-r)/\lambda\in\partial g(r)$, and choose
$s_p\in\partial g(p)$ from the joint optimality condition. With
$d=q-p$ and $h=q-r$, subtraction of the two optimality conditions gives
\[
 A d+s_r-s_p=0,\qquad
 A d:=\tau^{-1}(q-p)+\nabla f(q)-\nabla f(p).
\]
Monotonicity between $(r,s_r)$ and $(p,s_p)$ gives
$\langle A d,d-h\rangle\leq0$. Strong monotonicity and Lipschitzness of
$\tau^{-1}I+\nabla f$ imply
\[
 (\tau^{-1}+m_M)\|d\|^2
 \leq\langle A d,d\rangle
 \leq\langle A d,h\rangle
 \leq(\tau^{-1}+L_M)\|d\|\,\lambda L_g^M.
\]
This proves \eqref{eq:jp-equiv}, including $d=0$.
Reflection doubles the proximal difference. At a common state and noise,
the $Q$ and $P$ outputs thus differ by at most $\varepsilon_{\rm model}$;
the $P$ outputs from two states differ by at most $\rho$ times the state
distance by Theorem~\ref{thm:contraction}. Apply the triangle inequality
in $L^2$ to an optimal initial coupling to obtain
\eqref{eq:jpcr-cross-kernel}.
\end{proof}

The linear bound strengthens the submitted
$L_g^M\sqrt{\tau\lambda/2}$ proximal bound; a smaller observed gap does
not contradict that earlier upper bound. At the spectrally balanced
$\tau_*=(m_ML_\lambda)^{-1/2}$, direct substitution gives
\begin{equation}
 \frac{\varepsilon_{\rm model}}{1-\rho_*}
 =L_g^M\sqrt{\frac{\lambda}{m_M}}\,C_{\rm JP}(\lambda),\quad
 C_{\rm JP}(\lambda)=
 \frac{\sqrt{m_ML_\lambda}+L_M}{\sqrt{m_ML_\lambda}+m_M}
 \left(\sqrt{\lambda L_\lambda}+\sqrt{\lambda m_M}\right)
 \longrightarrow1.
 \label{eq:jpcr-balanced}
\end{equation}
Thus the accumulated joint-model gap has the same order as the Moreau
term. This identity concerns the kernel comparison, which holds for every
$\delta>0$; it does not enlarge the step range of the discretisation
estimate in Theorem~\ref{thm:end-to-end}. For TV,
$L_g^M=O(\rho_{\rm TV}\sqrt d)$ with a constant depending on $D$ and $M$;
the dimension dependence is not removed by the comparison.

\paragraph{A certificate for the finite ADMM solve.}
Suppose $g=\phi\circ D$ and set
$F_w(x)=f(x)+\|x-w\|_M^2/(2\tau)$,
$J_w(x)=F_w(x)+\phi(Dx)$ and
$\mu_{\rm prox}=m_M+\tau^{-1}$.
Let $(x_K,z_K,y_K)$ be the primal, split and unscaled dual variables after
a complete ADMM iterate with an exact $z$ update. Then
$y_K\in\partial\phi(z_K)$: this follows directly from the $z$ optimality
condition and the subsequent dual update. Define
\begin{equation}
 \begin{split}
 a_K&=\nabla F_w(x_K)+D^\top y_K,\\
 c_K&=\phi(Dx_K)-\phi(z_K)-\langle y_K,Dx_K-z_K\rangle,\\
 U_K&=c_K+\frac{\|a_K\|_{M^{-1}}^2}{2\mu_{\rm prox}}.
 \end{split}
 \label{eq:jpcr-certificate}
\end{equation}
\begin{proposition}[Primal--dual inner-error certificate]
\label{prop:jpcr-certificate}
Under the preceding conditions,
\begin{equation}
 0\leq J_w(x_K)-J_w(p)\leq U_K,\qquad
 \|x_K-p\|_M\leq\sqrt{2U_K/\mu_{\rm prox}}.
 \label{eq:jpcr-certified-distance}
\end{equation}
\end{proposition}
\begin{proof}
Fenchel weak duality gives
$-F_w^*(-D^\top y_K)-\phi^*(y_K)\leq J_w(p)$.
Put $s=-D^\top y_K$. The primal--dual gap is the sum of
\[
 F_w(x_K)+F_w^*(s)-\langle s,x_K\rangle
 \quad\text{and}\quad
 \phi(Dx_K)+\phi^*(y_K)-\langle y_K,Dx_K\rangle.
\]
The first term is at most
$\|\nabla F_w(x_K)-s\|_{M^{-1}}^2/(2\mu_{\rm prox})$ because
$F_w^*$ is $1/\mu_{\rm prox}$-smooth in the dual norm. Fenchel equality
at $(z_K,y_K)$ makes the second term exactly $c_K\geq0$.
Finally, $\mu_{\rm prox}$-strong convexity of $J_w$ gives
$J_w(x_K)-J_w(p)\geq\mu_{\rm prox}\|x_K-p\|_M^2/2$.
\end{proof}

For anisotropic TV, $\phi(z)=\rho_{\rm TV}\|z\|_1$ and the exact
soft-threshold update ensures $\|y_K\|_\infty\leq\rho_{\rm TV}$.
The equivalent box-feasible expression
$c_K=\rho_{\rm TV}\|Dx_K\|_1-\langle y_K,Dx_K\rangle$ can be used
after projecting a floating-point dual onto this box; it only requires
dual feasibility. A split residual alone omits this complementarity
term and is not the certificate in \eqref{eq:jpcr-certificate}.

\paragraph{Implemented process and end-to-end bound.}
Let $\widehat X_j$ be the image iterate and $S_j$ all warm-started ADMM
variables. At each step fresh Gaussian noise is independent of the past;
the inner budget $K$ can be fixed or chosen from the history. Apply the
certificate at that step's actual input and let
\[
 e_{K,j}=\left(\frac{8\mathbb E U_{K,j}}{\mu_{\rm prox}}\right)^{1/2},
 \qquad D_0=W_2^M(\mathcal L(\widehat X_0),\pi_{\delta,\lambda}),
\]
where $\pi_{\delta,\lambda}$ is the invariant law of $P$.
Whenever the displayed expectations are finite,
\begin{equation}
 \begin{split}
 W_2^M(\mathcal L(\widehat X_N),\pi)
 \leq{}&\rho^N D_0+
 \sum_{j=0}^{N-1}\rho^{N-1-j}
       (\varepsilon_{\rm model}+e_{K,j})\\
 &+\frac{C_1(\lambda)\sqrt d\,\delta^{3/2}
                +\overline C_2(\lambda)\delta^2}{1-\rho}
       +L_g^M\sqrt{\lambda/m_M}.
 \end{split}
 \label{eq:jpcr-end-to-end}
\end{equation}
Here the constants and step range are those of
Theorem~\ref{thm:end-to-end}; in particular its displayed small-step
estimate applies for $0<\delta\leq2/L_\lambda$.
The kernel and solver terms themselves have no such step restriction.

To prove the marginal recursion, at each step optimally couple
$\widehat X_j$ with $Y_j\sim\pi_{\delta,\lambda}$, lift this coupling to
$(\widehat X_j,S_j,Y_j)$ by conditioning on $\widehat X_j$, and draw
fresh shared noise independently. Evolve $Y_j$ with $P$.
By \eqref{eq:jpcr-certified-distance}, reflection gives
$\|\widehat X_{j+1}-Q(\widehat X_j,\xi_j)\|_M
\leq2\sqrt{2U_{K,j}/\mu_{\rm prox}}$.
Writing $D_j=W_2^M(\mathcal L(\widehat X_j),\pi_{\delta,\lambda})$,
Minkowski's inequality and \eqref{eq:jpcr-cross-kernel} yield
$D_{j+1}\leq\rho D_j+\varepsilon_{\rm model}+e_{K,j}$.
Iteration and the stationary discretisation and smoothing terms of
Theorem~\ref{thm:end-to-end} prove \eqref{eq:jpcr-end-to-end}.
This bounds the image marginal of the augmented process
$(\widehat X_j,S_j)$; no Markov assumption on $\widehat X_j$ alone is used.

\paragraph{Existence and uniqueness for the exact joint kernel.}
\begin{proposition}[Existence and stationary comparison]
\label{prop:jpcr-existence}
For every fixed $\delta>0$, $Q$ has an invariant law in $\mathcal P_2$.
Every such law $\nu_Q$ satisfies
$W_2^M(\nu_Q,\pi_{\delta,\lambda})
\leq\varepsilon_{\rm model}/(1-\rho)$.
\end{proposition}
\begin{proof}
Iterate \eqref{eq:jpcr-cross-kernel} against the stationary $P$ chain.
For any $\mu_0\in\mathcal P_2$, the distances
$W_2^M(Q^n\mu_0,\pi_{\delta,\lambda})$ are bounded uniformly in $n$.
The triangle inequality in $L^2$ therefore bounds the second moments of
$Q^n\mu_0$ uniformly. The kernel is Feller because the proximal map is
Lipschitz; dominated convergence applies to bounded continuous test
functions. Tightness of the Cesaro averages and the
Krylov--Bogoliubov argument give an invariant law, with finite second
moment by lower semicontinuity. Apply
\eqref{eq:jpcr-cross-kernel} to any invariant $\nu_Q$ and the invariant
$P$ law and rearrange to obtain the stated distance bound.
\end{proof}

For clarity, reflected proximal maps need not be strict contractions:
for $f(x)=x^2/2$, $g(x)=a|x|$, the proximal map is zero on
$[-\tau a,\tau a]$ and its reflection is $x\mapsto-x$ there.
A synchronous coupling nonetheless satisfies the following useful bound.
Write $D_n=\|X_n-Y_n\|_M$ for two exact $Q$ chains.
\begin{equation}
 D_{n+1}^2\leq D_n^2-
 4\tau m_M\left(\frac{(D_n-2\tau L_g^M)_+}{1+\tau L_M}\right)^2.
 \label{eq:jpcr-synchronous}
\end{equation}
Indeed, in whitened coordinates put $a=X_n-Y_n$ and
$b=p(w_X)-p(w_Y)$. Optimality gives
$a=b+\tau(u_X-u_Y)$, with $u_i\in\partial(f+g)(p(w_i))$.
Thus $\|2b-a\|^2\leq\|a\|^2-4\tau m_M\|b\|^2$, while
$\|a\|\leq(1+\tau L_M)\|b\|+2\tau L_g^M$.
This proves \eqref{eq:jpcr-synchronous}. In particular $D_n$ decreases
and $\limsup_nD_n\leq\delta L_g^M$. For invariant initial marginals,
dominated convergence gives the unconditional comparison
$W_2^M(\nu_1,\nu_2)\leq\delta L_g^M$ between any two invariant laws.

\begin{theorem}[Unique invariant law in the imaging setting]
\label{thm:jpcr-unique}
If $f$ is strongly convex quadratic and
$g(x)=\sum_{e=1}^E\rho_e|d_e^\top x|$, with finitely many nonzero
$d_e$ and $\rho_e>0$, then $Q$ has a unique invariant law $\nu_Q$ in
$\mathcal P_2$. For every $\mu_0\in\mathcal P_2$,
$W_2^M(Q^n\mu_0,\nu_Q)\to0$. This statement gives no convergence rate.
\end{theorem}
\begin{proof}
Whiten and write $f(x)=\tfrac12(x-\mu)^\top H(x-\mu)+\mathrm{const}$,
with $m_MI\preceq H\preceq L_MI$; the transformed $g$ is still a finite
$\ell_1$ composite. Set $A=I+\tau H$. Choose a unit vector $v$ outside
the finite union of hyperplanes $d_e^\top A^{-1}v=0$, and put
$\sigma_e=\operatorname{sign}(d_e^\top A^{-1}v)$.
For this fixed sign vector define the affine branch
\[
 p_\sigma(w)=A^{-1}
 \left(w+\tau H\mu-\tau\sum_e\rho_e\sigma_e d_e\right).
\]
For any compact set of pairs $(x,y)$, a sufficiently large fixed $t$
makes every $\sigma_e d_e^\top p_\sigma(w)$ positive for all
$w=x+tv+z$, $w=y+tv+z$, $\|z\|\leq1$, and their connecting segments.
This follows because the coefficient of $t$ is the strictly positive
$|d_e^\top A^{-1}v|$, and the remaining terms are uniformly bounded on
the compact set. At every such $w$, the branch's sign pattern verifies
the joint-proximal KKT condition, so strong convexity gives
$p(w)=p_\sigma(w)$. On this region the reflected map has linear part
$2A^{-1}-I$ and norm
\[
 \rho_f=\max_{a\in\{m_M,L_M\}}|(1-\tau a)/(1+\tau a)|<1.
\]
Consequently the Gaussian event
$\|\sqrt{\delta/2}\,\xi-tv\|\leq1$ has a strictly positive probability
$p_C$, uniform over pairs in the chosen compact set $C$, and on this
event $D_{n+1}\leq\rho_fD_n$. This argument uses a fixed affine branch
and checks KKT at its endpoints; it makes no monotonicity assumption
about a true prox path as it crosses active faces.

For two invariant initial marginals, or one arbitrary $\mu_0\in\mathcal
P_2$ and one invariant marginal, use synchronous noises. The reflected
resolvent is nonexpansive, so $D_n\downarrow D_\infty$ almost surely and
$D_n\leq D_0\in L^2$. The second-moment bound proved above makes the pair
of marginals uniformly tight. If
$q=\mathbb P(D_\infty\geq a)>0$ for some $a>0$, choose compact $C$ with
$\sup_n\mathbb P((X_n,Y_n)\notin C)<q/2$. The strict-contraction event
then gives, at every $n$,
\[
 \mathbb E(D_n^2-D_{n+1}^2)
 \geq(1-\rho_f^2)p_C\,
       \mathbb E[D_n^2\mathbf1_{\{(X_n,Y_n)\in C\}}]
 \geq(1-\rho_f^2)p_Ca^2q/2>0.
\]
Summing contradicts $\mathbb E D_0^2<\infty$. Hence $D_\infty=0$;
dominated convergence gives $\mathbb E D_n^2\to0$, proving uniqueness
and the stated $W_2$ convergence.
\end{proof}

\begin{remark}[Scope of the JP guarantees]
\label{rem:jp-vs-true}
The exact joint kernel has the model-gap and stationary guarantees above,
and the finite-ADMM process has \eqref{eq:jpcr-end-to-end}.
Uniqueness of the implemented warm-start process is not asserted.
The uniqueness theorem covers quadratic likelihoods with anisotropic TV
or a finite wavelet-$\ell_1$ penalty; it does not assert uniqueness for
every convex nonsmooth prior or a quantitative mixing rate for $Q$.
The measured per-iteration speedup reported below retains its original
solver-budget and hardware conditions; it is not a consequence of a
Wasserstein bound or a guarantee of equal accuracy at equal cost.
\end{remark}

\subsection{Kernel and finite-solver audits}
\label{app:jpcr-audit}

\begin{figure}[t]
 \centering
 \includegraphics[width=.94\linewidth]{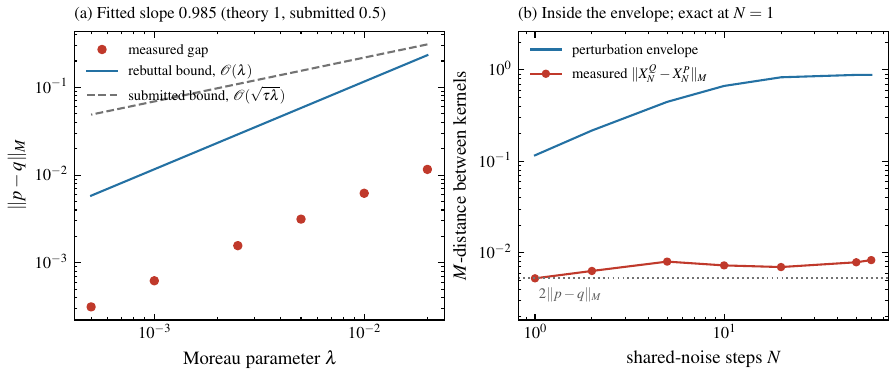}
 \caption{\textbf{Joint versus smoothed proximal kernels.}
 Left: measured proximal gap and the linear and earlier square-root
 upper bounds. The title's ``submitted $0.5$'' denotes the exponent of
 the earlier upper bound, not a predicted measured slope.
 Right: the original single shared-noise realization and the model-gap
 envelope. The corrected eight-chain RMS values are in
 Table~\ref{tab:jpcr-paired}; neither panel is a bound on SSE.}
 \label{fig:jpcr-gap}
\end{figure}

\paragraph{Exact-kernel comparison.}
The $64\times64$ audit solves the two proximal problems to a relative
movement tolerance $10^{-8}$ across six values of $\lambda$ spanning a
factor of $40$. The maximum gap over three probe points has log--log
slope $0.985$, consistent with \eqref{eq:jp-equiv}.
Figure~\ref{fig:jpcr-gap} reports this audit; its right
panel is one shared-noise realization. The independent eight-chain
aggregation is reported in Table~\ref{tab:jpcr-paired} so that a single
realization is not confused with a multi-chain RMS.
\begin{table}[t]
 \centering\small
 \caption{Eight-chain kernel comparison at $64^2$,
 $\lambda=0.005$, $\delta=0.141$. The gate predicts an initial RMS
 separation $0.00412567$; its measured/predicted ratio is $1.0000226$.}
 \label{tab:jpcr-paired}
 \begin{tabular}{rrrr}
  \toprule
  $N$ & RMS $M$-distance & Across-chain min--max & Model envelope\\
  \midrule
  1  & 0.004126 & 0.002429--0.005619 & 0.11647\\
  10 & 0.007550 & 0.006160--0.008652 & 0.66887\\
  60 & 0.007132 & 0.005937--0.008315 & 0.88406\\
  \bottomrule
 \end{tabular}
\end{table}

\paragraph{Production-budget certificate.}
At $256^2$, $\delta=0.141$, $\rho_{\rm TV}=0.3$, and $\rho_a=3$,
the three-probe reference model-gap scale is $7.16792$ in the reflected
$M$-distance. At an audited warm-start input after 200 preliminary
updates, the production budget $K=12$ gives a measured step error
$2.92717$, or $0.408$ times that scale. Its converged reference solve
is recorded in the supplied output. The computable certificate is
$56.87016$, or $7.934$ times the same scale: the observed $0.4$ ratio
is a reference-solve comparison, not a certified upper bound.
For $K=100$ the certificate is $2.84233$ ($0.397$ times that scale).
These are pointwise solver audits, not estimates of the expectation
$\mathbb E U_{K,j}$ along every step of a production chain.
The latter $K=100$ reference solve did not satisfy its requested stopping
tolerance, so we use its certificate but do not report a ``true error''
from that reference. The audit establishes the usefulness of the
certificate without declaring every finite reference solve converged.

\paragraph{Inner-budget sensitivity and timings.}
Table~\ref{tab:jpcr-budget} uses the saved, paired geometric-phantom
runs with $8$ chains, $N=2000$, $X_0=0$, and time averages accumulated
from the first step. The TRUE Picard reference uses its submitted
$(K_{\rm Picard},K_{\rm inner})=(30,50)$; JP varies only its ADMM budget.
These supplementary timings were measured on an RTX 5000 Ada and are
kept separate from the original Blackwell timings below.
\begin{table}[t]
 \centering\small
 \caption{JP inner-budget audit at $256^2$. SSE is a reconstruction
 statistic (mean $\pm$ SE over eight chains), not a kernel-distance
 estimate. ``Certificate'' is the separate warm-input upper bound
 in reflected $M$-distance; it does not use a converged reference solve.}
 \label{tab:jpcr-budget}
 \begin{tabular}{lrrr}
  \toprule
  Method / ADMM budget & Per-chain SSE & Time (min) & Certificate\\
  \midrule
  JP, $K=12$  & $851.80\pm0.77$ & 0.198 & 56.870\\
  JP, $K=25$  & $885.46\pm0.80$ & 0.386 & --\\
  JP, $K=50$  & $899.52\pm0.82$ & 0.715 & --\\
  JP, $K=100$ & $904.23\pm0.83$ & 1.348 & 2.842\\
  JP, $K=200$ & $905.40\pm0.83$ & 2.904 & --\\
  TRUE Picard & $854.49\pm0.76$ & 25.240 & --\\
  \bottomrule
 \end{tabular}
\end{table}
The near agreement of JP at $K=12$ with TRUE Picard changes as the inner
budget increases: the SSE offset is $-0.32\%$ at $K=12$ and $+5.96\%$
at $K=200$. This is consistent with partial cancellation between solver
and model errors at the production budget. It does not establish
equality of the invariant laws, and the $M$-distance bound in
Proposition~\ref{prop:jp-equivalence} is not a relative-SSE bound.
The original production setting is retained; the larger budgets serve
to expose this sensitivity and to report a tighter certificate.

\paragraph{Reproducibility.}
The eight-chain output is
\path{rebuttal/R4/exp_r4_jp_kernel_K8_20260727.json}.
The certificate and budget-sweep scripts and JSON outputs are in
\texttt{rebuttal/R4/};
\path{rebuttal/R4/README_camera_ready.md} gives the commands, fixed
configurations, and reference-convergence limitations. The audit
scripts preserve raw outputs and write new checks to separate paths.

%% file: sections/appendix_experiments_rebuttal.tex
\section{Additional bias, convergence and posterior-model checks}
\label{app:rebuttal-experiments}

\subsection{High-dimensional nonquadratic bias coefficient}
\label{app:highdim-coefficient}

The quartic examples in Appendix~\ref{app:experiments} are stress tests outside the global assumptions of Theorem~\ref{thm:bea-bias}: the one-dimensional quartic has unbounded curvature, and the coupled two-dimensional quartic is not globally convex. Here we use a smooth, globally strongly convex target with bounded Hessian on the periodic $16\times16$ lattice ($d=256$):
\begin{equation}
 \begin{gathered}
f(z)=\tfrac12\|z\|^2+\alpha\sum_{e}\log\cosh((Bz)_e),\\
 \qquad \alpha=0.7,\qquad M=I+\gamma B^\top B,
 \quad \gamma\in\{0,0.15,0.35\},
 \end{gathered}
\label{eq:lattice-coefficient-target}
\end{equation}
where $B$ is the edge-incidence matrix of the periodic square grid. Since $\|B\|^2=8$, $I\preceq\nabla^2f\preceq6.6I$; all derivatives of order two and higher are bounded. The target remains fixed when $\gamma$ changes. By translation and lattice symmetry,
\begin{equation}
 h_M(z)=\operatorname{tr}(M^{-1})+\alpha w(\gamma)\sum_e\operatorname{sech}^2((Bz)_e),
 \qquad w(\gamma)=b_e^\top M^{-1}b_e,
 \label{eq:lattice-trace}
\end{equation}
with a common leverage $w(\gamma)$ for every row $b_e^\top$ of $B$. For $\varphi=h_M$, the predicted coefficient $\tfrac14\operatorname{Var}_{\pi}(h_M)$ therefore scales as $w(\gamma)^2$.

We use five step sizes $\delta\in\{0.02,0.032,0.05,0.08,0.128\}$, $64$ chains per step size, $3{,}000$ burn-in steps and $120{,}000$ retained steps per chain. The inner Newton solve uses tolerance $10^{-11}$. A Metropolis-corrected $M$-MALA reference uses $16$ chains, $50{,}000$ burn-in steps and $900{,}000$ retained steps each. Reference rank-normalized split-$\hat R$ rounds to $1.0000$; bulk ESS for $h_M$ is about $10^6$. The observables are $h_M$, $d^{-1}\|z\|^2$, and the normalized edge energy $\|Bz\|^2/(2d)$.

\begin{figure}[htbp]
\centering
\includegraphics[width=0.98\linewidth]{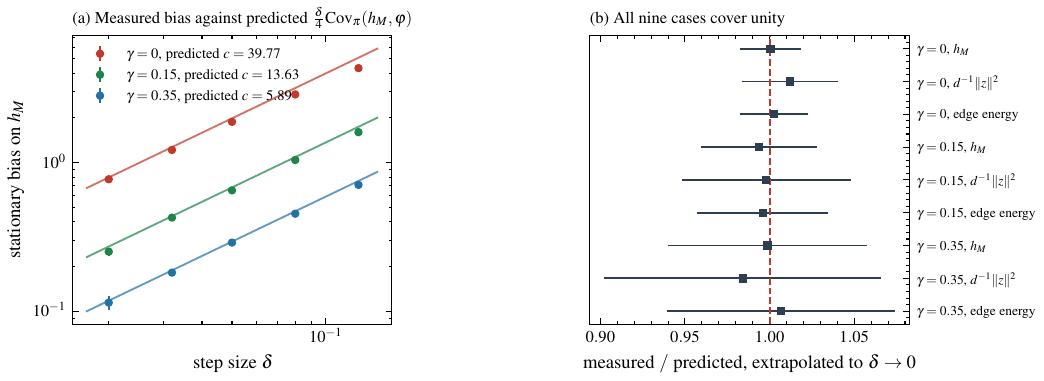}
\caption{\textbf{Bias coefficient on a fixed $d=256$ target.} Left: measured $h_M$ bias (bars: one MCSE) and the predicted leading term for three preconditioners. Right: extrapolated measured/predicted ratios with nominal $95\%$ weighted-fit intervals. The target is held fixed throughout.}
\label{fig:rebuttal-coefficient}
\end{figure}

For each observable we fit the measured-to-predicted ratio of bias divided by $\delta$ as $c_0+c_1\delta$, using inverse squared reported Monte Carlo standard errors as weights. Table~\ref{tab:lattice-coefficient} and Figure~\ref{fig:rebuttal-coefficient} report nominal $95\%$ weighted-fit intervals, $c_0\pm1.96\,\mathrm{SE}(c_0)$; these are not one-standard-error intervals. All nine contain one. The $h_M$ coefficient changes from $39.7727$ to $13.6324$ to $5.8883$, a factor of $6.75$, following the predicted leverage-squared law to within $0.2\%$. Thus the leading invariant defect depends quantitatively on $M$ even for one fixed target.

\begin{table}[htbp]
\centering\small
\caption{High-dimensional coefficient check. The slope fits $|\pi_\delta(\varphi)-\pi(\varphi)|$ against $\delta$ on logarithmic axes. The last column is the extrapolated measured/predicted ratio with a nominal $95\%$ interval, computed with the same diagonal MCSE weights as Figure~\ref{fig:rebuttal-coefficient}.}
\label{tab:lattice-coefficient}
\begin{tabular}{@{}clcc@{}}
\toprule
$\gamma$ & Observable & Slope & Ratio $c_0\pm1.96\,\mathrm{SE}$\\
\midrule
0 & $h_M$ & 0.930 & $1.0005\pm0.0179$\\
0 & $d^{-1}\|z\|^2$ & 0.928 & $1.0120\pm0.0286$\\
0 & Edge energy & 0.948 & $1.0024\pm0.0202$\\
0.15 & $h_M$ & 0.993 & $0.9938\pm0.0341$\\
0.15 & $d^{-1}\|z\|^2$ & 1.018 & $0.9980\pm0.0500$\\
0.15 & Edge energy & 1.007 & $0.9959\pm0.0388$\\
0.35 & $h_M$ & 0.986 & $0.9989\pm0.0590$\\
0.35 & $d^{-1}\|z\|^2$ & 1.052 & $0.9841\pm0.0817$\\
0.35 & Edge energy & 0.985 & $1.0066\pm0.0674$\\
\bottomrule
\end{tabular}
\end{table}

\subsection{Overdispersed chains and observable-specific diagnostics}
\label{app:overdispersed-diagnostics}

The finite-budget benchmark in Section~\ref{sec:experiments} uses a common zero start and no burn-in. We additionally run $13$ geometric-phantom configurations and five cameraman configurations at $256^2$, with eight starts: zero, the observation, the minimizer of $f$, two noise-perturbed minimizers, two constant fields and a checkerboard. Ordinary runs have $5{,}000$ iterations, discarding $500$; the three refined geometric runs have $10{,}000$, discarding $1{,}000$. The refined cameraman run has $5{,}000$, discarding $500$. These experiments use RTX~4090 hardware and a baseline $(K_{\mathrm P},K_{\rm in})=(8,15)$, not the $(30,50)$ solver budget of the headline experiment. Hence their absolute SSEs and timings are not replacements for Table~\ref{tab:imaging}.

We compute rank-normalized split-$\hat R$, bulk/tail effective sample sizes (ESS), and Monte Carlo standard errors \citep{vehtari2021rank}. The diagnostic implementation is checked on independent draws, AR(1) processes and chains with shifted means. Table~\ref{tab:rebuttal-diagnostics} reports sampling diagnostics alone; reconstruction SSE and PSNR are reported separately in Table~\ref{tab:roughness-reconstruction}.

\begin{table}[htbp]
\centering\small
\caption{Diagnostics by observable and experimental group. Energy/TV/high-frequency ranges pool only those three scalar observables; they do not bound every observable. The spatial mean retains appreciable autocorrelation, especially in the refined cameraman run. Values are rounded from the archived summaries.}
\label{tab:rebuttal-diagnostics}
\setlength{\tabcolsep}{4pt}
\begin{tabular}{@{}llccc@{}}
\toprule
Group & Observables & Maximum $\hat R$ & Bulk ESS range & Tail ESS range\\
\midrule
Geometric (13 configurations) & $f+g$, TV, high-frequency & 1.0076 & $768$--$18{,}938$ & $855$--$33{,}086$\\
Cameraman (5 configurations) & $f+g$, TV, high-frequency & 1.0141 & $432$--$19{,}516$ & $146$--$28{,}584$\\
Geometric, $\lambda=10^{-3}$ & Spatial mean & 1.0750 & $71$ & $104$\\
Refined cameraman & Spatial mean & 1.1244 & $42$ & $36$\\
\bottomrule
\end{tabular}
\end{table}

Across these runs classical per-pixel $\hat R$ is at most $1.002$, but this does not remove the slow spatial-mean mode exposed by the split diagnostics. We therefore treat the refined small-step runs as sensitivity checks, not exact-posterior ground truth or a certificate of convergence for all functionals. In particular, convergence diagnostics measure mixing of the implemented chain; they do not bound its discretization, smoothing or inner-solve bias.

\subsection{Separating draw roughness from posterior-mean texture}
\label{app:roughness-decomposition}

We vary $\delta$, $\lambda$, $\rho_{\mathrm{TV}}$ and the inner solve. The baseline is $\delta=0.141$, $\lambda=0.005$, $\rho_{\mathrm{TV}}=0.3$, $\rho_a=200$ and $(K_{\mathrm P},K_{\rm in})=(8,15)$. The inner-solve arms use $\rho_a=0.5$ with budgets $(4,8)$, $(8,15)$ and $(16,30)$. The refined runs use $\delta=0.141/8$, $\lambda=0.001$, $\rho_a=0.5$, and $(12,20)$ on geometric or $(8,15)$ on cameraman. The Moreau/Picard inner solver and the joint-proximal solver need different ADMM penalties; these settings do not prescribe changing the joint solver's penalty.

Define high-frequency draw energy by summing $|\widehat x(k)|^2/d$ over $|k|>n/4$. Reducing $\lambda$ from $0.005$ to $0.001$ lowers its chain-averaged value by $79.6\%$ on both phantoms; dividing $\delta$ by four changes it by only $+0.42\%$ and $+0.41\%$. The geometric inner-solve arms change it by $1.5$--$3.3\%$. The large $\lambda$ effect is accompanied by a covariance-trace decrease from approximately $2.45\times10^5$ to $4.99\times10^4$. The high-frequency modes occupy $0.8039$ of the Fourier grid; across all 18 configurations, chain-averaged high-frequency energy divided by $0.8039\operatorname{tr}(\operatorname{Cov}X)$ lies between $1.185$ and $1.242$. Its nearly constant proportionality links draw roughness to posterior width, with some excess high-frequency covariance. These are controlled changes of the implemented smoothed chain, not a claim that either value equals the variance of the nonsmoothed posterior.

\begin{figure}[htbp]
\centering
\includegraphics[width=0.98\linewidth]{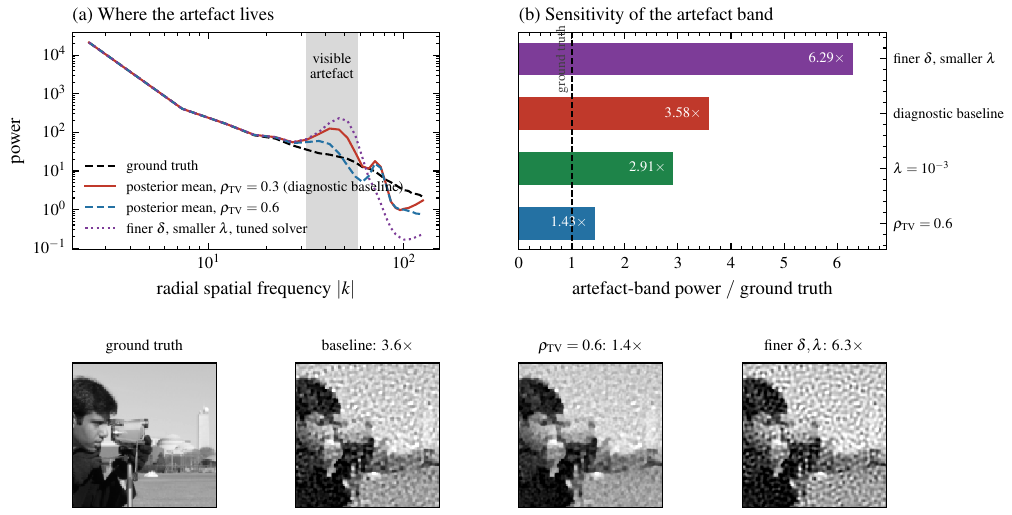}
\caption{\textbf{Cameraman posterior-mean texture and regularization sensitivity.} The shaded spectral band is $32\leq|k|<58$. Bottom panels use identical crops and grayscale limits $[0,1]$. The diagnostic baseline uses $(K_{\mathrm P},K_{\rm in})=(8,15)$ and is distinct from the headline reconstruction protocol. The refined run is a sensitivity check with the residual slow mode reported in Table~\ref{tab:rebuttal-diagnostics}.}
\label{fig:rebuttal-roughness}
\end{figure}

The visible texture in the posterior \emph{mean} lies in a different band, $32\leq|k|<58$ at $n=256$ (approximately $5.6$ pixels in wavelength). Let $B$ denote mean-image power in this band divided by ground-truth power in the same band. Doubling $\rho_{\mathrm{TV}}$ reduces $B$ from $5.47$ to $2.09$ on geometric and $3.58$ to $1.43$ on cameraman, improving PSNR by $3.46$ and $2.70$\,dB, respectively. The cameraman change is visible in Figure~\ref{fig:rebuttal-roughness}. Reducing $\lambda$ alone also reduces this mean-image band power ($3.58$ to $2.91$); the two controls therefore do not act exclusively on disjoint effects. Decreasing $\delta$ alone increases it to $5.43$, and the refined cameraman run has $B=6.29$. Thus a less textured output need not be a less biased sampler. Changing $\rho_{\mathrm{TV}}$ changes the posterior model itself; the independent target check below distinguishes this effect from a P-IMLA implementation artifact.

\begin{table}[htbp]
\centering\small
\caption{\textbf{Reconstruction quality, separate from convergence diagnostics.} These are pooled posterior-mean SSE, PSNR using the ground-truth intensity range, and mean-image band-power ratio $B$ for the overdispersed $(8,15)$ baseline protocol. All other parameters are fixed when $\rho_{\mathrm{TV}}$ is doubled.}
\label{tab:roughness-reconstruction}
\begin{tabular}{@{}lcccc@{}}
\toprule
Phantom & $\rho_{\mathrm{TV}}$ & SSE & PSNR (dB) & $B$\\
\midrule
Geometric & 0.3 & 544.94 & 20.80 & 5.47\\
Geometric & 0.6 & 245.40 & 24.27 & 2.09\\
Cameraman & 0.3 & 615.73 & 20.19 & 3.58\\
Cameraman & 0.6 & 330.71 & 22.89 & 1.43\\
\bottomrule
\end{tabular}
\end{table}

\begin{figure}[htbp]
\centering
\includegraphics[width=0.70\linewidth]{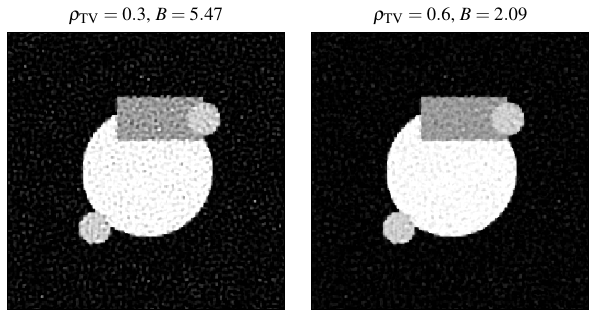}
\caption{\textbf{Geometric posterior means with stronger TV regularization.} The two archived $256^2$ reconstructions differ only in $\rho_{\mathrm{TV}}$: eight chains, $5{,}000$ iterations with $500$ discarded, $(K_{\mathrm P},K_{\rm in})=(8,15)$, and common grayscale limits $[0,1]$. Stronger regularization reduces band-power ratio $B$ from $5.47$ to $2.09$ and improves PSNR from $20.80$ to $24.27$\,dB (Table~\ref{tab:roughness-reconstruction}). This controlled diagnostic comparison uses a different inner budget and averaging protocol from the original headline reconstructions.}
\label{fig:geometric-regularization}
\end{figure}

\subsection{Independent exact-target reference at $32^2$ and $48^2$}
\label{app:exact-tv-reference}

To test the nonsmoothed posterior independently of P-IMLA, we use data-augmentation Gibbs sampling for
\begin{equation}
 \pi(x)\propto\exp\!\left[-\frac{\|Ax-y\|^2}{2\sigma^2}
 -\frac{\eta}{2}\|x\|^2-\rho_{\mathrm{TV}}\|Dx\|_1\right],
 \qquad \sigma=0.05,\quad\eta=10^{-3}.
 \label{eq:exact-tv-reference}
\end{equation}
Each Laplace factor is a Gaussian scale mixture: $z_e\mid\tau_e\sim N(0,\tau_e)$, $\tau_e\sim\operatorname{Exp}(\rho_{\mathrm{TV}}^2/2)$. The conditional precision for $x$ is $C+D^\top\operatorname{diag}(1/\tau_e)D$, where $C=A^\top A/\sigma^2+\eta I$; its conditional mean solves this precision times the mean equal to $A^\top y/\sigma^2$. We sample this Gaussian by dense Cholesky and update $1/\tau_e$ from its inverse-Gaussian conditional. These transitions use neither Moreau smoothing nor time discretization nor P-IMLA's inner solver. ``Exact-target'' describes the invariant target and conditional draws; finite Monte Carlo error remains.

The blur is fixed at $2$ pixels at both resolutions, and the artifact band is fixed in cycles per pixel, $|k|/n\in[32/256,58/256)$. These are smaller members of the same model family, not continuum-coarsened copies of the $256^2$ problem; their band-power ratios need not equal those at $256^2$. All $32^2$ runs and the $48^2$, $\rho_{\mathrm{TV}}=0.6$ run use four chains of $4{,}000$ steps, discarding $1{,}000$. The other two $48^2$ runs use eight chains of $8{,}000$, discarding $2{,}000$.

\begin{table}[htbp]
\centering\small
\caption{Exact-target Gibbs reference. $B$ is posterior-mean band power divided by ground-truth band power at the same resolution. The MC floor estimates band power due to finite-chain mean uncertainty from between-chain variation, in the same normalized units.}
\label{tab:exact-tv-reference}
\begin{tabular}{@{}cccccc@{}}
\toprule
Resolution & $\rho_{\mathrm{TV}}$ & $B$ & MC floor & SSE & PSNR (dB)\\
\midrule
$32^2$ & 0.15 & 12.394 & 0.00195 & 253.93 & 6.06\\
$32^2$ & 0.30 & 6.856 & 0.00123 & 89.66 & 10.58\\
$32^2$ & 0.60 & 3.028 & 0.00061 & 37.95 & 14.31\\
$48^2$ & 0.15 & 15.749 & 0.00040 & 627.30 & 5.65\\
$48^2$ & 0.30 & 9.403 & 0.00029 & 205.19 & 10.50\\
$48^2$ & 0.60 & 4.257 & 0.00060 & 76.35 & 14.80\\
\bottomrule
\end{tabular}
\end{table}

\begin{figure}[htbp]
\centering
\includegraphics[width=0.94\linewidth]{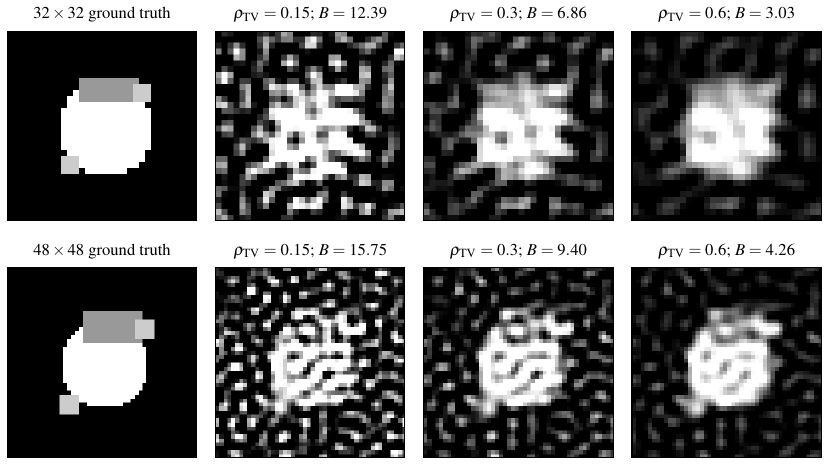}
\caption{\textbf{Posterior means of the nonsmoothed TV model from independent Gibbs sampling.} Top: $32^2$; bottom: $48^2$. All panels use grayscale limits $[0,1]$; numerical metrics use the unclipped arrays. The mean-image band-power ratio $B$ decreases as $\rho_{\mathrm{TV}}$ increases. This figure reuses the archived reference samples and introduces no new sampling run.}
\label{fig:exact-tv-reference}
\end{figure}

The mean-image texture persists under this independently implemented exact-target sampler and decreases with stronger TV regularization (Figure~\ref{fig:exact-tv-reference}, Table~\ref{tab:exact-tv-reference}). We validate the Gaussian draw against the closed-form mean marginal variance (relative error $6.66\times10^{-4}$ over $20{,}000$ draws), and cross-check a matrix-free perturb-and-MAP/CG implementation: at $32^2$, the band ratios are $6.8662$ and $6.8588$, with mean pixel standard deviations differing by $0.064\%$. Independent Metropolis sampling of the exact potential at $48^2$, $\rho_{\mathrm{TV}}=0.3$, gives $B=9.364$ versus Gibbs $9.403$ and mean pixel standard deviation $1.3338$ versus $1.3401$. These checks concern distributional summaries, not pointwise identity of finite-run posterior means. They support a model-side source of the texture without asserting that P-IMLA has zero bias relative to this reference.

\subsection{What the finite-budget Gaussian statistic measures}
\label{app:vr-transient-decomposition}
\label{app:r5-vr}

\begin{figure}[htbp]
\centering
\includegraphics[width=0.98\linewidth]{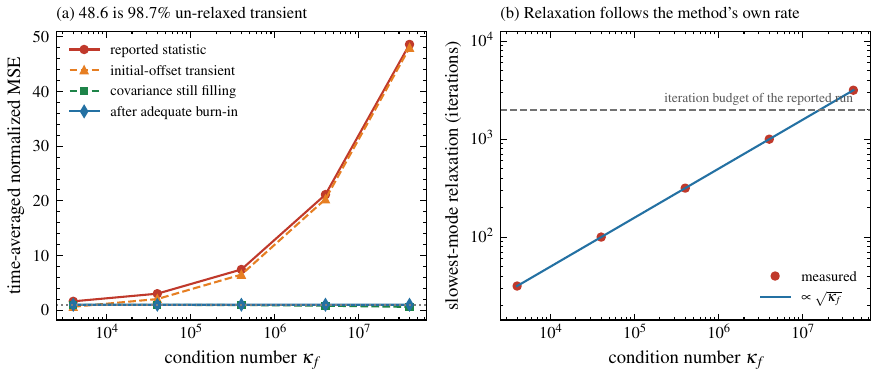}
\caption{\textbf{Initial transients in the Gaussian comparison.} Left: exact decomposition of the cold-start time-averaged normalized MSE into mean-offset and covariance terms. Right: the slowest-mode relaxation time follows $\sqrt{\kappa_f}$. Adequate burn-in recovers a ratio close to one; the invariant ratio is exactly one.}
\label{fig:vr-decomposition}
\end{figure}

For the Gaussian experiment the reported statistic is
\[
 V_N=\frac1{Nd}\sum_{n=1}^N(X_n-\mu)^\top H(X_n-\mu),
 \qquad X_0=0.
\]
It combines covariance filling and the initial mean offset. In orthonormal Hessian coordinates with eigenvalues $h_i$ and Cayley factors $r_i=(1-\delta h_i/2)/(1+\delta h_i/2)$,
\[
 \mathbb E V_N=\frac1d\sum_i\left[1-a_{i,N}+h_i\mu_i^2a_{i,N}\right],
 \qquad a_{i,N}=\frac1N\sum_{n=1}^Nr_i^{2n}.
\]
For unpreconditioned IMLA at $\kappa_f\simeq4\times10^7$ and $N=2{,}000$, this gives $48.6329=0.6104+48.0225$: $98.7\%$ is the unrelaxed initial-mean contribution. Direct simulation gives $48.5974$. The stationary variance ratio is exactly one, as required by Gaussian exactness even without preconditioning. The slowest-mode relaxation time is $3{,}162$ iterations, consistent with the method's $\sqrt{\kappa_f}$ dependence. Figure~\ref{fig:vr-decomposition} therefore explains a finite-budget mixing advantage, not failure of IMLA's Gaussian invariant covariance.

